\documentclass{article}

\usepackage{microtype}
\usepackage{graphicx}
\usepackage{subcaption}
\usepackage{stfloats}  
\usepackage{booktabs} % for professional tables
\newcommand{\GR}{\textsc{GRouter}}

\newcommand{\RETURN}{\STATE \textbf{return} }
\usepackage[accepted]{icml2026}

\usepackage{amsmath}
\usepackage{amssymb}
\usepackage{mathtools}
\usepackage{amsthm}
\usepackage{hyperref}

\usepackage[capitalize,noabbrev]{cleveref}

\theoremstyle{plain}

\theoremstyle{definition}

\theoremstyle{remark}

\usepackage[textsize=tiny]{todonotes}

\icmltitlerunning{Grouter: Decoupling Routing from Representation for Accelerated MoE Training}

\begin{document}

\twocolumn[
  \icmltitle{Grouter: Decoupling Routing from Representation for \\Accelerated MoE Training}



  \icmlsetsymbol{equal}{*}
  \icmlsetsymbol{intern}{$\dagger$}
  \begin{icmlauthorlist}
    \icmlauthor{Yuqi Xu}{intern,pekingmath,zhejiang}
    \icmlauthor{Rizhen Hu}{pekingmath}
    \icmlauthor{Zihan Liu}{pekingyuanpei}
    \icmlauthor{Mou Sun}{zhejiang}
    \icmlauthor{Kun Yuan}{pekingdata}
  \end{icmlauthorlist}

  \icmlaffiliation{pekingmath}{School of Mathematical Sciences, Peking University, Beijing, China}
  \icmlaffiliation{pekingdata}{Center for Machine Learning Research, Peking University, Beijing, China}
  \icmlaffiliation{pekingyuanpei}{Yuanpei College, Peking University, Beijing, China}
  \icmlaffiliation{zhejiang}{Zhejiang Lab, Hangzhou, China}
  %\icmlaffiliation{aisi}{AI for Science Institute,Beijing,China}

  \icmlcorrespondingauthor{Kun Yuan}{kunyuan@pku.edu.cn}
  \icmlcorrespondingauthor{Mou Sun}{123sssmmm@gmail.com}

  \icmlkeywords{Mixture-of-Experts, MoE, Preemptive Routing, Grouter, GROUTER, router, distillate}

  \vskip 0.3in
]



\printAffiliationsAndNotice{
$\dagger$: The research was finished while the author was an intern at Zhejang Lab.
}  % no special notice (required even if empty)

\begin{abstract}
\vspace{-1mm}
Traditional Mixture-of-Experts (MoE) training typically proceeds without any structural priors, effectively requiring the model to simultaneously train expert weights while searching for an optimal routing policy within a vast combinatorial space. This entanglement often leads to sluggish convergence and training instabilities. This paper introduces \GR, a preemptive routing method that by distilling high-quality structures from fully-trained MoE models and serving as a fixed router for target models. By decoupling structural optimization from weight updates, \GR\ significantly accelerates both the speed and quality of model convergence. To ensure the framework's versatility, we also introduce expert folding to adapt \GR\ across varying model configurations and expert tuning to rebalance workloads across different data distributions. Furthermore, by leveraging the structural priors provided by preemptive routing, we can implement targeted optimizations to further enhance training throughput. Experiments demonstrate that \GR\ achieves superior performance and efficiency which boosts pre-training data utilization by \textbf{4.28$\times$} and achieves up to \textbf{33.5$\%$} throughput acceleration, establishing preemptive routing as a fundamental paradigm for scalable MoE training. We publicly release our code and pretrained \GR\ checkpoints \footnotemark.
\end{abstract}

\vspace{-8mm}

\section{Introduction}\label{introduction}
\vspace{-1mm}
Large Language Models have rapidly become a foundation of modern AI, exhibiting strong capabilities in understanding and reasoning~\cite{bahdanau2014neural,vaswani2017attention,liu2019roberta,brown2020language,liu2024deepseek}. A central driver of these gains is scaling: larger models, trained on more data, tend to deliver better performance~\cite{kaplan2020scaling}. Yet naively scaling dense Transformers to the trillion-parameter regime is prohibitively expensive. Mixture-of-Experts~\cite{shazeer2017outrageously} architectures address this tension by replacing dense feed-forward layers with a pool of experts and a router that activates only a small subset per token. This sparse activation allows parameter counts to grow without a proportional increase in per-step FLOPs, enabling training and inference at practical cost~\cite{du2022glam,wei2026team}. \footnotetext{\url{https://github.com/JimmyAwoe/Grouter}}

\begin{figure*}[tp]
    \centering 
    \begin{subfigure}{0.33\textwidth}
        \centering
       \includegraphics[width=\textwidth]{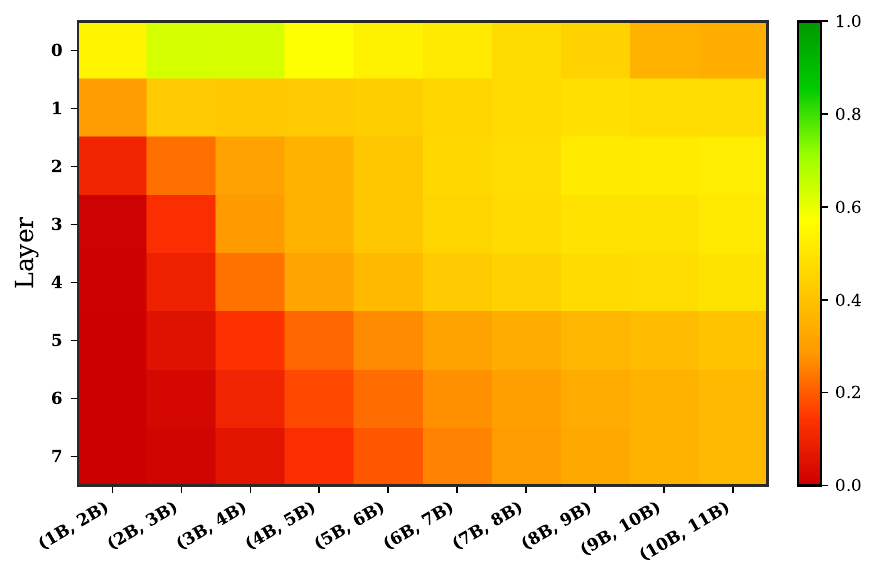} 
        \caption{} 
        \label{fig:exact_matching}
    \end{subfigure}
    \hfill  
    \begin{subfigure}{0.33\textwidth}
        \centering
       \includegraphics[width=0.95\textwidth]{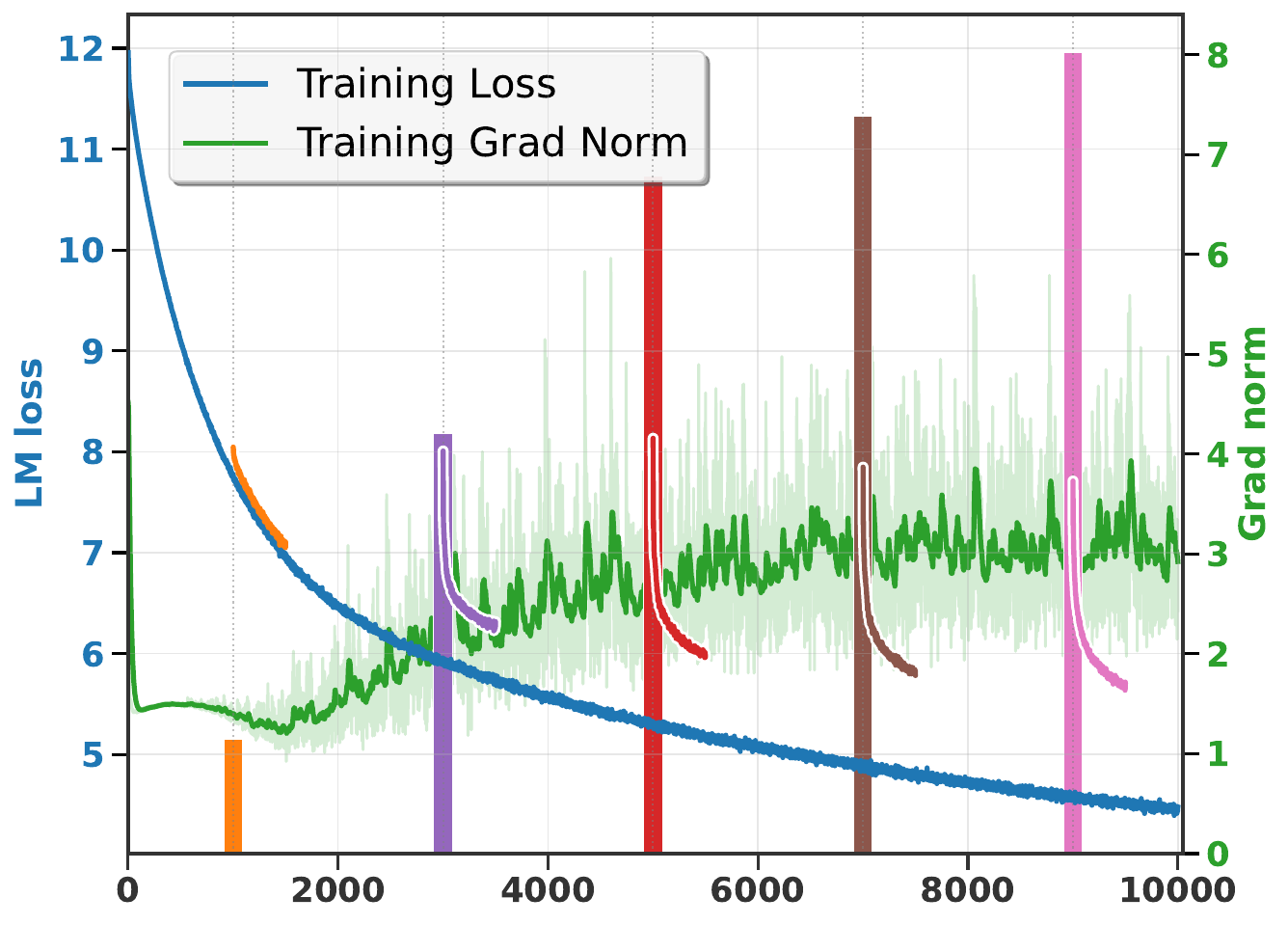} 
        \caption{} 
        \label{fig:random_routing}
    \end{subfigure}
    \hfill  
    \begin{subfigure}{0.33\textwidth}
        \centering
        \includegraphics[width=1\linewidth]{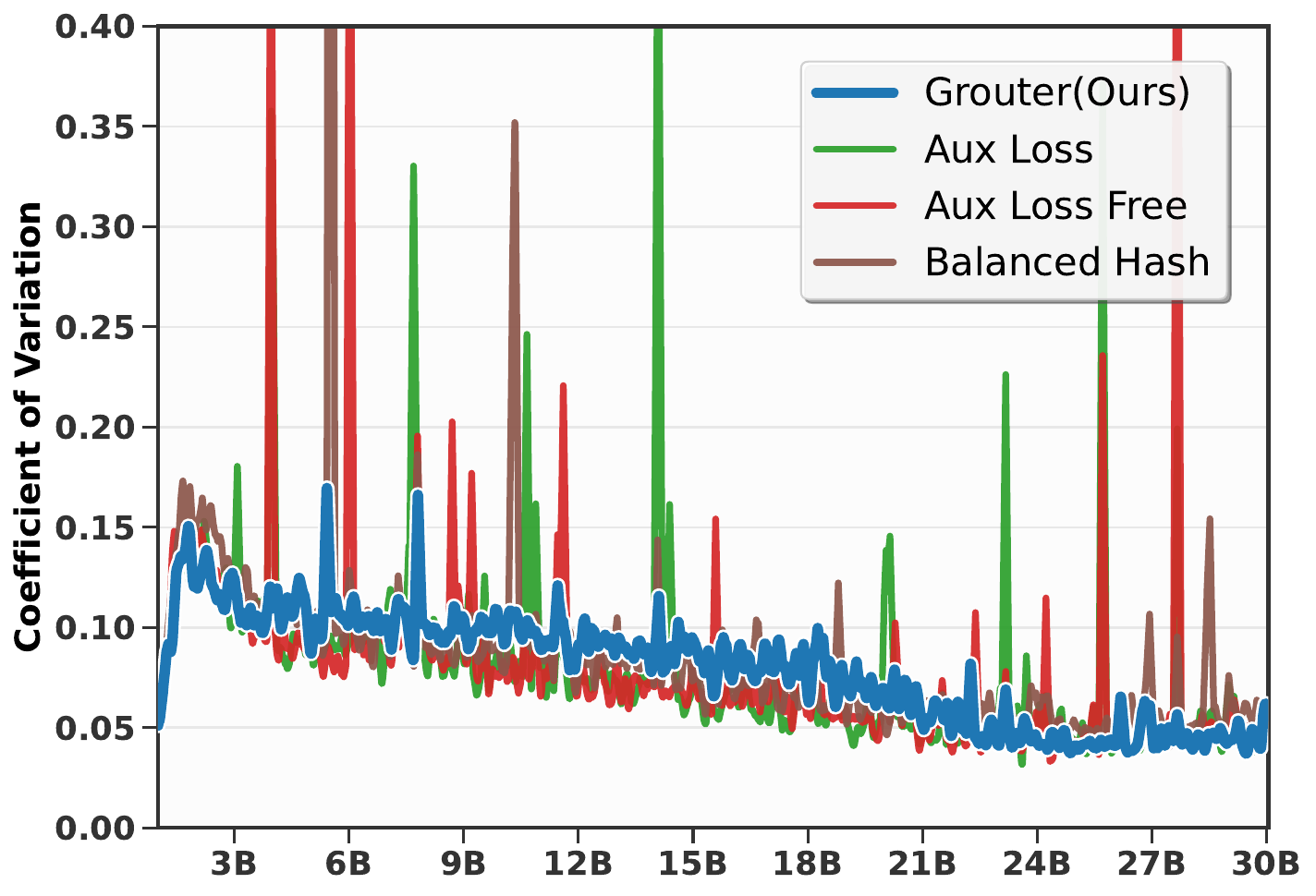}
        \caption{}  
        \label{fig:grad_norm_cv}
    \end{subfigure}
    \vspace{-9mm}
    \caption{\small (a) The percentage of tokens that maintain an exactly identical set of $\mathbf{k}$ activated experts for the same input across adjacent checkpoints. We present the more detailed descriptions and figures in \autoref{Detailed Desctiption of Routing Fluctuation Heatmap}. (b) Sensitivity analysis of expert specialization via random routing perturbations. Perturbations were applied at specific intervals with a fixed learning rate of $10^{-5}$. The resulting loss trajectories (lines) and average gradient norms (bars) reveal a clear trend: while early-stage training (1k steps) is resilient to routing noise due to a lack of specialization, later stages exhibit severe instability and loss spikes under perturbation. This demonstrates that experts gradually develop deep specialization as the router stabilizes, making the model increasingly sensitive to routing errors. (c) The coefficient of variation of the gradient norm within a sliding window of length 100. The results demonstrate that the Grouter method achieves a significant improvement in stability, whereas both the state-of-the-art load balancing approach and pure static routing assignment lead to substantial gradient fluctuations.}
    \label{fig:insight_curve}
    \vspace{-5mm}
\end{figure*}

Despite their theoretical efficiency, MoE models are notoriously difficult to train~\cite{lepikhin2020gshard}. A fundamental cause of this difficulty is the tight coupling between routing structure learning and representation learning. In standard MoE training, the router and the experts are optimized simultaneously. The router must learn to partition the input space into balanced expert assignments, while the experts must simultaneously adapt their parameters to the evolving token distributions assigned to them. As illustrated in \autoref{fig:exact_matching}, our empirical observations reveal that the routing structure remains highly unstable even after extensive training, with expert assignments for identical inputs fluctuating significantly. This concurrent optimization of the router and representations forces experts to chase a "moving target" of shifting data distributions, preventing them from achieving deep specialization. Consequently, this joint training process—rather than the MoE architecture itself—leads to insufficient convergence and poor learning efficiency. We further formalize this destructive interference within a mathematical framework in \autoref{ap: Structure-Performance Interference}, where a gradient-based analysis elucidates how such concurrent updates hinder the development of specialized representations.% 我们观察了MoE的Router在训练的不同阶段对于相同的输入是否能够保持相同的分配，结果显示在Figure 1。可以观察到即使经过较长的训练过程，MoE仍然无法维持稳定的路由分配。而基于不稳定的路由学习导致专家无法充分特化，进而导致了不充分的收敛和低下的学习效率。我们在附录1通过对梯度计算进行分析，构建了一个数学框架来解释这个干扰。

Several recent efforts have attempted to refine this routing process. Methods such as Auxiliary Loss Free~\cite{wang2024auxiliary}, differentiable routers like ReMoE~\cite{wang2024remoe} or Lory~\cite{zhong2024lory} aim to optimize the router more effectively, while TC-MoE~\cite{yan2025tc}, MoE++~\cite{jin2024moe++} and Elastic MoE~\cite{gu2026elasticmoeunlockinginferencetime}, introduce dynamic expert selection to increase structural flexibility. However, these approaches still require the model to construct its load-balanced structure on-the-fly during training. By performing structural search and representation learning within the same optimization loop, they fail to resolve the underlying instability: the router is forced to partition a feature space that is itself shifting, while experts must specialize on tokens that they may not consistently receive in subsequent iterations. While some preliminary attempts have been made to decouple the router from the MoE training process, significant challenges remain; we provide a detailed discussion on the limitations of these specific approaches in \autoref{Limitations in Existing Approaches}.

%Even methods that specifically target decoupling, like StableMoE~\cite{dai2022stablemoe, roller2021hash}, struggle because they attempt to learn the routing structure during the most volatile early phases of training; the "teacher" structure they distill from is itself a byproduct of this early-stage volatility.

%This reveals a fundamental Catch-22 in MoE training: To learn an optimal routing structure, the router requires stable expert representations as a reference; yet to develop stable experts, the model requires a consistent and high-quality routing structure to ensure specialized data distribution. Attempting to bootstrap both simultaneously from scratch creates a "moving target" problem that inevitably leads to sub-optimal convergence and limits training efficiency.

% \begin{figure}[htbp]
%     \centering
%     \includegraphics[width=0.48\textwidth]{figures/routing_exact_match_rates_heatmap.pdf}  
%     \vspace{-5mm}
%     \caption{\small 
%     % Quantifying Routing Stability by Token-to-Expert Consistency.Token-level routing consistency is measured by the fraction of tokens selecting the same expert set for an identical input across different training steps. The observed high volatility in early training confirms the intrinsic structural instability that exacerbates $\text{SPI}$.
%     The percentage of tokens that maintain an exactly identical set of $\mathbf{k}$ activated experts for the same input across adjacent checkpoints. }  
%     \label{fig:exact_matching}
%     \vspace{-5mm}
% \end{figure}

%(b)这张图展示了训练过程中对路由结果进行随机扰动的Grad norm和loss的变化情况。我们分别在训练了1000，3000，5000，7000，9000步时对router输出进行随机扰动，记录之后500步的loss情况并统计100步的grad norm平均值,学习率始终保持1e-5以消除可能的影响，分别以折线和柱状图的形式展示于图中。我们同时还展示了训练过程的损失值和grad norm情况。可以看到在训练至1000步时随机扰动几乎没有对训练造成干扰，这显示在训练前期由于路由波动，专家特化情况极其微弱，而随着训练的进行，随机扰动几乎导致训练彻底崩溃--loss大幅上涨，grad norm显著偏离正常水平。这显示专家在训练中随着router稳定逐步开始特化。

%  However, these approaches still require the model to simultaneously construct its load-balanced structure during training, fundamentally failing to resolve the instability stemming from the coupling of routing and representation. To address this, some researchers have explored decoupling strategies, yet they face significant hurdles:

% \textbf{Challenge 1: Manual designs lack expressiveness.} HashMoE~\cite{roller2021hash} attempted to use fixed, manually assigned token-to-expert mappings. While this ensures a stable structure, such rigid mappings cannot capture contextual nuances—the same token in different contexts should not be routed identically. This implies that a high-quality structure must be learned, not manually engineered.

% \textbf{Challenge 2: Early-stage training is too volatile for structure learning.} StableMoE~\cite{dai2022stablemoe} attempted to learn the structure in an initial phase and then freeze it. However, because the MoE structure is inherently dynamic and unstable during early training, the distilled auxiliary network fails to acquire a high-quality, converged routing logic.

To overcome these challenges, this paper introduces the \textbf{General Router}, abbreviated as \GR. It learns a high-quality, well-optimized structure by distilling it from a fully trained MoE model (e.g. Qwen3-30B-A3B). \GR\ avoids the pitfalls of manual design and circumvents the instability of learning from a dynamically evolving model by extracting structure from a stable, converged state. Once distilled, \GR\ serves as a plug-and-play replacement for the trainable router and offers three key advantages. \textbf{First}, \GR\ inherits high-quality structural priors from converged models, surpassing those learned from scratch. \textbf{Second}, by fixing the routing pathways, it decouples routing topology from representation, thereby eliminating optimization interference and focusing training resources on task performance. \textbf{Third}, this decoupled structure serves as a stable foundation that facilitates the injection of further routing-based optimizations.

%\GR\ acquires structures by distilling knowledge from a fully trained MoE model. To construct a structure tailored to a specific training task, \GR\ enables selective structure extraction by specifying the dataset and data ratio used during distillation. Since the extracted structure is tailored to a specific data distribution, it may exhibit load imbalance; therefore, we introduce expert tuning to perform task reallocation and promote load balance. To allow \GR\ to adapt to different MoE configurations, we further design an output head folding method. This method enables a single distilled \GR\ instance to be applied across MoE models with varying configurations. Consequently, \GR\ requires only a single distillation step before it can be flexibly transferred to diverse scenarios.

\GR\ distills structural knowledge from a fully trained MoE model. To ensure versatility, we incorporate two key mechanisms: Expert Tuning, which rebalances workloads to adapt to specific data distributions, and Expert Folding, which enables the router to fit varying MoE configurations. This design allows a single distilled \GR\ instance to be efficiently transferred across diverse training scenarios.

We summarize our key contributions as follows:
\vspace{-2mm}
\begin{enumerate}
\item \textbf{Analyzing the Necessity of Decoupling in MoE Training.} We empirically demonstrate that the entanglement between routing structure and representation learning restricts MoE scaling. We show that decoupling these processes is critical for achieving optimal convergence speed and training stability.

\item \textbf{Introducing Grouter for Preemptive Structure Construction.} We propose \GR, which distills optimal routing structures from converged models. By establishing a fixed routing topology prior to training, \GR\ fundamentally eliminates the interference between structure learning and representation updates.

%We propose \GR, a novel structure construction methodology that learns the load-balanced routing structure of a fully trained MoE model via distillation. This preemptive routing strategy completely decouples structure construction from performance optimization, thereby eliminating Structure–Performance Interference and enabling stable and effective MoE training.

\item \textbf{Expanding the Optimization Space via Structural Priors.} Leveraging the fixed priors provided by \GR, we shift data optimization from runtime to a pre-processing stage. This decoupling bypasses the limitations of dynamic routing, enabling the application of sophisticated offline algorithms to significantly expand the optimization space.
%Leveraging the routing prior information provided by \GR, we can perform data optimization in advance. Crucially, because this optimization is decoupled from the model training process, \GR\ offers a significantly larger optimization space compared to previous optimization algorithms that relied on real-time router output. This decoupling enables the application of more sophisticated and complex algorithms for data optimization.

\end{enumerate}

% 我们将Grouter适配到Megatron-LM中，并进行了充分的实验。我们的实验结果显示，在550M的模型上的预训练实验显示，相对于Baseline，Grouter可以仅使用23.3%的数据实现相同的验证集损失，即4.28倍的数据利用率。在不同架构，不同规模的实验上，Grouter同样表现出巨大优势,可以在相同的数据量下使350m模型打败1B模型。我们还基于Grouter进行了吞吐量的实验，结果显示我们可以在合适的专家并行设置下至多可以达到33.5%的吞吐量提升。这展示了Grouter全方位的对MoE预训练上的帮助和提升。我们将会开源我们的代码和\GR\的checkpoint，以帮助社区更好的适配和改进Grouter。

We implemented \GR\ within the Megatron-LM~\cite{shoeybi2019megatron} and conducted comprehensive experiments utilizing clusters of NVIDIA H100 and A100. Our pre-training results on a $550\text{M}$ parameter $\text{MoE}$ model demonstrate remarkable data efficiency: \GR\ achieved the same validation set loss as the baseline model using only \textbf{23.3\%} of the data, corresponding to a \textbf{4.28$\times$} improvement in data utilization efficiency. Furthermore, \GR\ consistently exhibited significant advantages across experiments with different architectures~\cite{openai2025gptoss120bgptoss20bmodel,deepseekv2,qwen3technicalreport} and scales. We also performed throughput experiments, showing that \GR\ can achieve up to a \textbf{33.5\%} throughput increase under appropriate Expert Parallelism settings. Collectively, these results demonstrate the comprehensive benefits and advancements that \GR\ brings to $\text{MoE}$. We will open-source our code and \GR\ checkpoints to facilitate community adoption and further improvements. 

\section{Background and Motivation}
\subsection{Mixture-of-Experts}\label{Mixture-of-Experts}
The MoE architecture has emerged as a critical paradigm for scaling Transformer models~\cite{shazeer2017outrageously}. Its primary distinction from dense Transformers lies in the Feed-Forward Network (FFN) layer, which in MoE comprises a router and a set of $E$ experts. During the forward pass, the router assigns $k$ experts to each input token, achieving sparse parameter activation. The FFN component of a single MoE layer can be formally described as:
\vspace{-2mm}
\begin{align}
    \mathbf{y} &= \sum_{i=1}^E \mathbf{r}(\mathbf{x})_i \mathbf{f}_i(\mathbf{x}), \label{eq:moe_output} \\
 \mathbf{r}(\mathbf{x}) &= \mathbf{g}\left(\text{TopK}(\mathbf{s}(\mathbf{x}), k)\right). \label{eq:routing_func}
\end{align}
Here, $\mathbf{x} \in \mathbb{R}^d$ denotes the input to the MoE layer, and $\mathbf{s}(\mathbf{x}) \in \mathbb{R}^E$ represents the raw scores assigned by the router to the $E$ experts. The $\text{TopK}(\cdot, k)$ operator retains only the scores of the $k$ highest-scoring experts. Subsequently, $\mathbf{g}(\cdot)$ is a normalization function, typically $\text{Softmax}$ or $\text{Sigmoid}$, applied to these selected scores to yield the final coefficient vector $\mathbf{r}(\mathbf{x}) \in \mathbb{R}^E$. Specifically, $\mathbf{r}(\mathbf{x})_i$ denotes the weight applied to the output of expert $\mathbf{f}_i(\mathbf{x})$. 
The MoE output $\mathbf{y}$ is then passed through subsequent layers, eventually contributing to the final task loss $\mathcal{L}$ (e.g., cross-entropy for language modeling). 
It is noteworthy that although the notation here only explicitly shows the Router network, the overall structure is implicitly influenced by all preceding network parameters.

\subsection{Inefficiency of Joint Router-Expert Optimization}\label{Inefficiency of Joint Router-Expert Optimization}
% 我们对MoE训练不同阶段时专家的特化水平进行了实验。我们在训练的不同阶段将router的输出随机化，观察专家是否能够基于随机分配的token进行正常训练，结果展示在图2.可以看到，训练前期专家特化程度较浅，router的突然随机化并不会造成训练崩溃。而训练一段时间后专家特化程度会逐渐加深。这于我们在图1观察到的现象相吻合。这展现出在MoE的训练过程中，MoE将从router-expert共同学习逐渐过渡至固定router，提升expert能力的阶段。正如我们在introduction部分的推导，router-expert共同学习的阶段带来的moving target问题导致了训练效率的下降。因此，我们希望能够为MoE引入结构先验，解耦routing structure与representation学习过程，进而促进专家特化，提升训练效率。

% We conducted an empirical study on the expert specialization levels across different training stages. Specifically, we introduced stochastic perturbations by randomizing the router's outputs at various intervals and %观察模型的反应。如果grad norm和loss均出现激增，这说明专家无法适应随机分配的token，进而说明专家特化程度更深。 

 We conducted an empirical study on the expert specialization~\cite{hu2026synergisticintracrosslayerregularization} levels across different training stages. Specifically, we introduced stochastic perturbations by randomizing the router's outputs at various intervals and analyzed the subsequent fluctuations. A sharp spike in gradient norms and loss suggests that the experts are unable to adapt to arbitrary token assignments. Consequently, this heightened sensitivity to routing noise reflects a deeper degree of expert specialization.

%As illustrated in \autoref{fig:random_routing}, a clear trend emerges: in the early stages of training, the experts exhibit a relatively low degree of specialization; thus, abrupt randomization of the routing logic does not trigger training collapse, as the experts have not yet formed distinct niche representations. However, as training progresses, the level of specialization significantly deepens. % 结合\autoref{fig:exact_matching}，我们将这一现象归因于前期router的不稳定，因为这会导致专家频繁切换学习任务，从而无法特化。而随着路由稳定，专家特化程度逐渐加深。

As illustrated in \autoref{fig:random_routing}, a clear trend emerges: in the early stages of training, the experts exhibit a relatively low degree of specialization; thus, abrupt randomization of the routing logic does not trigger training collapse, as the experts have not yet formed distinct niche representations. However, as training progresses, the level of specialization significantly deepens. In conjunction with \autoref{fig:exact_matching}, we identify routing instability as the primary cause of this early-stage behavior: the fluctuating routing policy subjects experts to constantly shifting objectives, preventing them from specializing. Once the routing stabilizes, experts are able to commit to specific domains, leading to the observed increase in specialization. 

%我们进一步观察训练过程中Grad norm的变异系数于\autoref{fig:grad_norm_cv},发现在训练的中后期(从3B开始)，模型逐渐出现grad norm的波动，我们认为这是因为专家随着训练逐步特化，然后router偶然会出现分配失误，导致特化的专家产生极高的梯度。

We further investigate the coefficient of variation (CV) of the gradient norm throughout the training process, as shown in \autoref{fig:grad_norm_cv}. We observe that fluctuations in the gradient norm gradually emerge during the mid-to-late stages. We ascribe this phenomenon to the progressive deepening of expert specialization: as experts become highly specialized, occasional routing errors expose them to incompatible tokens, thereby triggering spikes in gradient magnitude. 

%This observation is consistent with the structural fluctuations observed in \autoref{fig:exact_matching}, confirming that MoE models naturally transition from a phase of joint router-expert exploration to a phase of expert refinement once the routing structure stabilizes. This hypothesis is further corroborated by the visualizations provided in \autoref{Detailed Desctiption of Routing Fluctuation Heatmap}.

%This observation highlights a potential improvement: the initial period of joint learning forces experts to adapt to a constantly shifting data distribution, which may delay the onset of deep specialization. 

These findings suggest that the learning process can be optimized by introducing a structural prior from the beginning, which provides a consistent data distribution for each expert.  This decoupling allows the model to bypass the volatile exploration phase and focus directly on expert specialization during the subsequent training process, thereby significantly enhancing overall training efficiency. 

We provide a more comprehensive discussion on other advantages of this decoupling strategy in \autoref{ap:Benefits of Preemptive Structures}.

%by decoupling the routing structure from representation learning.

% \vspace*{-\topmargin}
% \vspace*{-\headheight}
% \vspace*{-\headsep}
% \vspace*{-\topskip}

\begin{figure*}[tp]
    \centering 
    \begin{subfigure}{0.48\textwidth}
        \centering
        \includegraphics[width=0.9\linewidth]{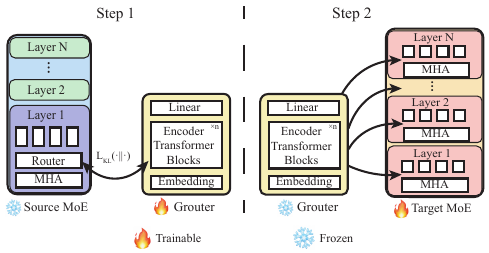}
        \caption{} 
        \label{fig:dis_pipe}
    \end{subfigure}
    \hfill  
    \begin{subfigure}{0.48\textwidth}
        \centering
        \includegraphics[width=\linewidth]{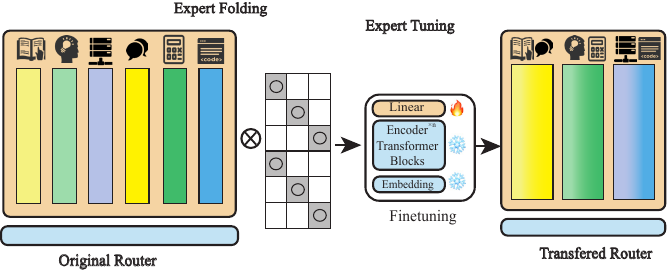}
        \caption{}  
        \label{fig:exp_fold}
    \end{subfigure}
    \vspace{-5mm}
    \caption{\small (a) Overview of the \GR\ Workflow. The \GR\ first extracts a highly optimized structural prior from the Source Model, and then injects this prior into the Target Model in a frozen state. (b) Illustration of Our Expert Tuning and Expert Folding Techniques.}
    \label{fig:grouter_struc}
    \vspace{-5mm}
\end{figure*}

%  \begin{figure}[htbp]
%     \centering
%     \includegraphics[width=\linewidth]{figures/pipeline.png}
%     \caption{\small Overview of the $\text{Grouter}$ Workflow. This figure illustrates the operational flow of our $\text{Grouter}$: the $\text{Grouter}$ first extracts a highly optimized structural prior from the Source Model, and then injects this prior into the Target Model in a frozen state.}
%     \label{fig:dis_pipe}
% \end{figure}

\section{Method}

The core objective of \GR\ is to fundamentally decouple the routing structure from representation learning by proactively extracting a stable, high-quality routing structure $\mathbf{r}^*(\cdot)$ from a well-optimized source MoE model. This extraction process involves \GR\ learning to replicate the input-to-expert mappings and weight assignments produced by the source model's router. The resulting fixed structure, provided by the frozen \GR\ network, then serves as a preemptive structural prior for target MoE model training. This approach fundamentally decouples structural optimization from task performance optimization, ensuring efficient training.

% The core objective of \GR\ is to fundamentally resolve the $\text{SPI}$ by proactively extracting a stable, high-quality routing configuration $\mathbf{r}^*(\cdot)$ from a well-optimized Source $\text{MoE}$ Model. This extraction process involves \GR\ learning to map inputs to the appropriate expert selection and weight assignment decisions made by the source model's router. 

% The resulting fixed structure, provided by the frozen \GR\ network, then serves as a preemptive structural prior. This approach fundamentally decouples the structural optimization from the main performance optimization of the Target $\text{MoE}$ Model, ensuring stable and efficient training.

\subsection{Grouter Architecture and Structure Extraction}
\subsubsection{Grouter Architecture}
The \GR\ network, denoted as $\mathbf{G}$, is designed as a lightweight, standalone structure extractor. Its primary role is to learn the desired routing decisions $\mathbf{r}^*(\cdot)$ from the source model. Unlike traditional MoE routers that operate within the model backbone, \GR\ processes raw token sequences $\mathbf{X}$ directly from the tokenizer, operating entirely independently. The architecture comprises an embedding layer, $N$ Transformer encoder blocks, and a final linear projection layer. Formally, $\mathbf{G}$ can be expressed as:
% The Grouter network, denoted as $\textbf{G}$, is designed to function as a lightweight, standalone structure extractor. Its primary role is to learn the desired routing decisions $\mathbf{r}^*(\cdot)$ from the source model. Unlike a traditional $\text{MoE}$ router, Grouter operates entirely decoupled from the backbone, directly processing the raw token output $\textbf{X}$ from the tokenizer. The architecture is composed of an initial embedding layer, $\mathbf{N}$ layers of $\text{Transformer}$ encoder blocks, and a final linear layer. Specifically, the function $\textbf{G}$ can been shown as:
\begin{equation}
    \mathbf{G}(\mathbf{X}) = \mathbf{W}_s \left( \text{Enc}^N \left( \text{Emb}(\mathbf{X}) \right) \right).
\end{equation}
Here, $\text{Emb}(\cdot)$ denotes the embedding layer, $\text{Enc}^N$ represents a stack of $N$ Transformer encoder blocks, and $\mathbf{W}_s$ is the final linear projection layer. The use of Transformer encoder blocks enables $\mathbf{G}$ to capture global context, allowing it to assess the contextual relevance of inputs when determining optimal expert assignments. This lightweight architecture minimizes computational overhead during pre-training. We present experiments in \autoref{ap:Ablation} that demonstrate the advantages of this architecture.

% Here, $\text{Emb}(\cdot)$ denotes the embedding layer, $\text{Enc}^N$ denotes the stack of $\textbf{N}$ $\text{Transformer}$ encoder blocks, and $\mathbf{W}_s$ is the final linear score layer. The use of $\text{Transformer}$ encoder blocks enables $G$ to possess a "global receptive field," allowing it to better assess the contextual relevance of the input when determining the optimal expert assignment. The design employs a lightweight architecture to minimize computational overhead during the pre-training phase.

\subsubsection{Structure Extraction}

As seen in \autoref{fig:dis_pipe}, We employ knowledge distillation to extract the high-quality stable routing structure from a large, fully converged source MoE model into the lightweight \GR\ network. The distillation objective is to train \GR\ to precisely replicate the expert assignment weights of the source model's router, thereby obtaining the desired preemptive router $\mathbf{r}^*(\cdot)$.

For a given input $\mathbf{X}$, let $\mathbf{H}$ be the hidden state that $\mathbf{X}$ propagates to at the input of the selected router layer in the source model $\mathbf{S}$. The raw expert scores output by this router are denoted by $\mathbf{s}_{\text{Sou}}(\mathbf{H})$. The distillation loss $\mathcal{L}_{\text{Distill}}$ is formulated using the Kullback-Leibler divergence: 
\begin{equation*}
    \mathcal{L}_{\text{Distill}} = D_{\text{KL}} \left( \text{Softmax}(\mathbf{s}_{\text{Sou}}(\mathbf{H})) \parallel \text{Softmax}(\mathbf{G}(\mathbf{X})) \right).
\end{equation*}
Unlike typical knowledge distillation, we do not incorporate a temperature parameter in the Softmax calculation. This is crucial because \GR\ must accurately learn the true magnitudes of expert contribution weights, not merely the rank ordering or relative differences of the raw logits. 

% For a given input $\mathbf{X}$, let $\mathbf{s}_S(\mathbf{X})$ be the raw expert scores output by the chosen layer's router in the Source Model $\textbf{S}$. The distillation loss $\mathcal{L}_{\text{Distill}}$ is formulated using the Kullback-Leibler Divergence:
% \begin{equation}
%     \mathcal{L}_{\text{Distill}} = D_{\text{KL}} \left( \text{Softmax}(\mathbf{s}_S(\mathbf{X})) \parallel \text{Softmax}(\mathbf{G}(\mathbf{X})) \right).
% \end{equation}
% Unlike typical $\text{KD}$, we do not incorporate a temperature parameter in the $\text{Softmax}$ calculation. This is crucial because Grouter must accurately learn the true magnitude of the expert contribution ratios $\mathbf{r}(\mathbf{X})$, not just the rank ordering or relative differences of the raw logits.

\textbf{Shared \GR\ Implementation.} In multi-layer MoE architectures, we implement a single \GR\ to guide all MoE layers. This design is supported by empirically observed redundancy in layer-wise routing. Specifically, \citet{cai2024textit} suggest that complex, layer-specific routing decisions are unnecessary, as high correlation exists among the routing structures of different MoE layers. This inherent redundancy indicates that a single, universally applicable structural prior suffices for the entire network. Our subsequent experiments confirm that this shared \GR\ does not compromise model performance. This design choice is critical for maintaining the computational efficiency of our method.

% \textbf{Shared Grouter Implementation.} In a multi-layer $\text{MoE}$ architecture, we implement only one single Grouter to instruct all $\text{MoE}$ layers. This choice is supported by the empirically observed redundancy of layer-wise router design. Specifically, research ~\cite{cai2024textit} suggests that the complex, layer-specific routing decisions are largely unnecessary, as a high correlation exists among the routing structures of different $\text{MoE}$ layers. This inherent redundancy means that a single, universally applicable structural prior is sufficient for the entire network. Subsequent experiments confirm that adopting this single \GR\ does not compromise the model's performance. Capitalizing on this insight is critical to maintaining the high lightweightness of our method.

\textbf{Layer Selection Strategy.} To maximize the quality of the structural prior derived from a single \GR, we distill the routing output from the first source MoE layer. This selection is motivated by the sequential nature of Transformers: routing deviations originating in early MoE layers are progressively amplified in deeper layers. This accumulation results in significantly higher routing fluctuation and lower structural stability in downstream layers. By distilling from the most stable routing pattern, \GR\ acquires a robust structural prior $\mathbf{r}^*(\cdot)$ that is minimally affected by cumulative errors.

% \textbf{Layer Selection Strategy.} To maximize the quality of the structural prior derived from a single \GR, we distill the routing output from the first MoE layer. This selection is motivated by the sequential nature of Transformers: routing deviations originating in early MoE layers are progressively amplified in deeper layers. This accumulation results in significantly higher routing fluctuation and lower structural stability in downstream layers. By distilling from the most stable routing pattern, \GR\ acquires a robust structural prior $\mathbf{r}^*(\cdot)$ that is minimally affected by cumulative errors.

\subsection{Expert Folding and Expert Expanding}
A critical challenge in adopting a fixed structural prior is its transferability across MoE models with varying expert configurations. Specifically, the number of experts in the source model from which \GR\ is distilled may differ from the number in the target model that \GR\ is intended to guide. To address this limitation, we propose Expert Folding and Expert Expanding, which enables a single \GR\ instance to flexibly adapt to varying expert counts.

% A critical challenge for adopting a fixed structural prior is its transferability across $\text{MoE}$ models with divergent expert configurations. Specifically, the number of experts in the Source Model from which \GR\ is distilled may not align with the number of experts in the Target Model that \GR\ is intended to instruct. To address this constraint, we propose Expert Folding to enable a single \GR\ instance to flexibly adapt to varying expert counts.

\subsubsection{Expert Folding}

\textbf{Expert Folding Mapping.} The Expert Folding procedure leverages an affinity-based merging strategy to construct a mapping that reduces the source expert count $E_S$ to the target count $E_T$. The process begins by characterizing the functional relationships among source experts. We execute \GR\ on the training dataset to compute the Expert Co-activation Affinity Matrix $\mathbf{P} \in \mathbb{R}^{E_S \times E_S}$, where element $\mathbf{P}_{ij}$ quantifies how often source experts $i$ and $j$ are simultaneously activated by \GR\ for the same input token. Next, we determine the required merging size for each of the $E_T$ target expert. Target experts are designed to incorporate $G_{\text{avg}} = \lfloor E_S / E_T \rfloor$ source experts on average, with $N_{\text{extra}} = E_S \bmod E_T$ groups requiring an additional expert. With group sizes established, we iteratively select an unassigned source expert $e_i$ to initiate a new merging group $C_k$. To maximize the resulting composite expert's effectiveness, we then repeatedly select the unassigned expert $e_j$ that maximizes collective co-activation affinity with the current group members:
\begin{equation}
    \underset{e_j \notin \text{Assigned}}{\arg\max} \sum_{e_m \in C_k} \mathbf{P}_{e_m, e_j}. 
\end{equation}
This process continues until group $C_k$ reaches its target size. By mapping each $C_k$ to a target expert $e_k$, we ensure maximal preservation of essential structures and specialized function within the resulting composite target experts.

\textbf{Matrix Folding Implementation.} The folding procedure is implemented efficiently as a linear transformation applied to the final score layer of the distilled \GR\ as shown in \autoref{fig:exp_fold}. Let $\mathbf{W}_s \in \mathbb{R}^{d \times E_S}$ be the original weight matrix of \GR's final score layer, and let $\mathbf{M} \in \{0,1\}^{E_S \times E_T}$ be the binary mapping matrix derived from the affinity-based merging strategy, where $\mathbf{M}_{ij} = 1$ indicates that source expert $i$ is mapped to target expert $j$. The folded weight matrix for the target model, $\mathbf{\tilde{W}}_s \in \mathbb{R}^{d \times E_T}$, is computed as:
\begin{equation}
    \mathbf{\tilde{W}}_s = \mathbf{W}_s \mathbf{M}.
\end{equation}
This operation is computationally negligible and requires minimal storage. Because $\mathbf{M}$ can be precomputed and stored for various target configurations, a single distilled \GR\ instance can serve as a structural prior for diverse MoE models simply by selecting the appropriate $\mathbf{M}$. This enables exceptional transferability and configuration flexibility for \GR\ with minimal overhead.

% \textbf{Matrix Folding Implementation.} The folding procedure is implemented efficiently as a linear transformation applied to the final score layer of the distilled \GR. Let $\mathbf{W}_s \in \mathbb{R}^{\cdot \times \mathbf{E}_S}$ be the original weight matrix of \GR's final score layer, and let $\mathbf{M} \in \{\mathbf{0},\mathbf{1}\}^{\mathbf{E}_S \times \mathbf{E}_T}$ be the binary mapping matrix derived from the greedy merging strategy, where $\mathbf{M}_{ij}$ symbolize source expert $i$ whether mapped to target expert $j$. The folded weight matrix for the Target Model, $\mathbf{\tilde{W}}_G \in \mathbb{R}^{\cdot \times E_T}$, is computed as:
% \begin{equation}
%     \mathbf{\tilde{W}}_s = \mathbf{W}_s \mathbf{M}.
% \end{equation}
% This operation is computationally and storage-wise trivial. Because $\mathbf{M}$ can be pre-calculated and stored for various target configurations, a single distilled \GR\ instance can serve as a structural prior for diverse $\text{MoE}$ models simply by selecting the appropriate $\mathbf{M}$. This enables the exceptional transferability and configuration flexibility of \GR\ with minimal overhead.
%  \begin{figure}[htbp]
%     \centering
%     \includegraphics[width=\linewidth]{figures/expet_fold_tune.png}
%     \caption{\small Illustration of Our Expert Tuning and Expert Folding Techniques.}
%     \label{fig:exp_fold}
%     \vspace{-2mm}
% \end{figure}

\subsubsection{Expert Expanding}\label{chapter:expert_expanding}

\textbf{Underserved Tokens Identification.} The expansion procedure begins by identifying tokens for which the existing structural prior provides inadequate coverage. Specifically, we execute the frozen \GR\ on a representative corpus $\mathcal{D}$ and extract the hidden representations $\mathbf{h} \in \mathbb{R}^d$ at the input of the final score layer $\mathbf{W}_s$ via a forward hook. For each token, we compute the maximum expert affinity score $s_{\max} = \max_{j} (\mathbf{h}^\top \mathbf{w}_j)$, where $\mathbf{w}_j$ denotes the $j$-th row of $\mathbf{W}_s$. Tokens whose $s_{\max}$ falls below the $q$-th quantile of the score distribution are designated as \emph{underserved tokens}, forming the set $\mathcal{U}$. Intuitively, these are tokens for which no existing expert direction provides a strong routing signal, and they constitute the natural demand for new experts.

\textbf{Centroid Clustering.} Given $E_{\Delta} = E_T - E_S$ new experts to be constructed, we apply $k$-means clustering on the hidden representations of the underserved token set $\mathcal{U}$ with $k = E_{\Delta}$. The resulting centroids $\{\mathbf{c}_1, \ldots, \mathbf{c}_{E_{\Delta}}\} \subset \mathbb{R}^d$ identify $E_{\Delta}$ representative directions in the hidden-state space that characterize the distribution of tokens poorly served by the current routing structure. Each centroid provides the initial direction for one new expert.

\textbf{Orthogonal Projection and Normalization.} To ensure the new expert directions are maximally independent from the existing ones, we project the centroids onto the null space of the current score weight matrix. Let $\mathbf{A} = \mathbf{W}_s^\top \in \mathbb{R}^{d \times E_S}$. The projection is computed as:
\begin{equation}
	\mathbf{c}_i' = \mathbf{c}_i - \mathbf{A}(\mathbf{A}^\top \mathbf{A})^{-1}\mathbf{A}^\top \mathbf{c}_i, \quad i = 1, \ldots, E_{\Delta}.
\end{equation}
This removes all components aligned with existing expert directions, preventing the new experts from duplicating learned routing signals. Subsequently, we apply QR decomposition~\cite{golub2013matrix} to the projected centroid matrix to enforce mutual orthogonality among the new experts themselves. Finally, each new direction is rescaled to match the average $\ell_2$-norm of the original expert weight vectors, ensuring consistent scoring magnitudes across old and new experts.

\textbf{Weight Assembly.} The expanded weight matrix is constructed by concatenating the original and new expert directions:
\begin{equation}
	\mathbf{W}_s^{\text{exp}} = \begin{bmatrix} \mathbf{W}_s \\ \mathbf{W}_{\Delta} \end{bmatrix} \in \mathbb{R}^{E_T \times d},
\end{equation}
where $\mathbf{W}_{\Delta} \in \mathbb{R}^{E_{\Delta} \times d}$ contains the orthogonalized and normalized new expert directions. Critically, the original weights $\mathbf{W}_s$ remain unchanged, preserving the full structural prior learned during distillation.

\subsection{Expert Tuning}\label{chapter:expert_tuning}
Once the \GR\ network $\mathbf{G}$ is distilled and expert folding is applied, its weights are frozen, yielding the fixed structural prior $\mathbf{G}^*$. This $\mathbf{G}^*$ provides constant, preemptive routing decisions for the target MoE model, successfully decoupling structure from performance optimization.

% Once the \GR\ network $\mathbf{G}$ is distilled and the expert folding procedure is applied, its weights are frozen, yielding the fixed structural prior $\mathbf{G}^*$. This $\mathbf{G}^*$ provides a constant, preemptive routing decision for the Target $\text{MoE}$ Model, successfully decoupling structure from performance optimization.

However, the load balancing of $\mathbf{G}^*$ may be suboptimal. This suboptimality does not stem from failures in structure extraction; rather, it arises because the structural prior distilled from the source model is optimized for its training data distribution $\mathcal{D}_S$, which may differ from the target model's deployment distribution $\mathcal{D}_T$. Due to inherent expert specialization and task-optimal routing patterns learned on $\mathcal{D}_S$, the distilled $\mathbf{G}^*$ naturally inherits a routing bias that, when applied to $\mathcal{D}_T$, leads to imbalanced load distribution.

% However, the load balancing performance of the resulting $\mathbf{G}^*$ may still be suboptimal. This suboptimality does not stem from a failure in structure extraction; instead, it arises because the structural prior distilled from the Source Model is optimized for its training data distribution $\mathcal{D}_S$, which may differ from the Target Model's deployment distribution $\mathcal{D}_T$. Due to the inherent expert specialization and task-optimal routing learned on $\mathcal{D}_S$, the distilled $\mathbf{G}^*$ naturally inherits a routing tendency that, when applied to $\mathcal{D}_T$, leads to an imbalanced load distribution.

 %我们在使用Grouter加速训练前，再基于$\mathcal{D}_T$进行一次微调。我们选择Gshard中使用的负载均衡损失$\mathcal{L}_{aux}$作为损失函数来促使Grouter适应新的分布。为了尽可能的保持Grouter学习到的结构，我们冻结除最后的线性映射层以外的参数，仅对这一几乎可忽略不计的参数量进行微调，我们后续的实验证明，仅需极少量的数据即可达到完全可使用的负载均衡水平。这使得Grouter可以方便的，轻量化的迁徙到不同的训练任务中。

To mitigate this imbalance without compromising the stable structure, we perform a lightweight fine-tuning of the \GR\ based on the Target Model's training distribution $\mathcal{D}_T$ before the main training phase as seen in \autoref{fig:exp_fold}. We adopt the $\mathcal{L}_{\text{aux}}$ load balancing loss, similar to that used in \cite{lepikhin2020gshard}, as the sole optimization objective to encourage the \GR\ to adapt to the new distribution.

Crucially, to maximally preserve the structure learned during distillation, we freeze all \GR\ parameters except for the final linear projection layer. This fine-tuning is applied only to this negligibly small parameter count. Our subsequent empirical results demonstrate that this adjustment requires only a minimal amount of data to achieve a perfectly usable level of load balance. This feature allows the \GR\ to be conveniently transferred with minimal overhead to target training tasks.

\begin{figure*}[tp]
    \centering
    \includegraphics[width=1\linewidth]{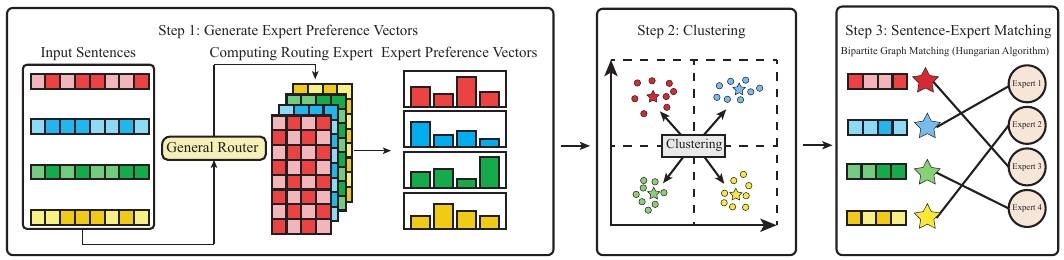}
    \caption{The workflow of our EP communication optimization strategy. Step 1: Sequences pass through \GR\ to generate token-level assignments, which are aggregated into sequence-level routing affinity vectors. Step 2: Sequences are clustered based on affinity. Using cluster centroids as preference weights, we solve an optimization problem to assign experts to EP devices, maximizing the alignment between experts and sequence clusters. Step 3: With expert locations fixed, sequences are assigned to the EP device that minimizes the resulting communication volume.}
    \label{fig:pipeline_for_ep_save}
    \vspace{-4mm}
\end{figure*}

%这张图展示了我们EP通信优化的全流程。Step1: 序列首先通过Grouter体现获取专家指派信息，接下来统计每条序列所有token的指派结果获取整条序列的 routing affinity vectors。Step2: 基于routing affinity vectors进行聚类，将序列按照EP数划分为多个簇，每一簇的簇心代表这一簇对于不同的专家的偏好程度。我们以每一簇对不同专家的偏好程度为权，以最大化整体专家偏好程度为目标，求解优化问题将专家分配至不同簇，相对应这些专家将被放置于不同的EP节点 Step3: 固定每个EP节点的专家组后，我们可以针对每个序列求解放置至不同节点的通信量，我们将每个序列分配至通信量最小的节点。

\subsection{Enhancing Training Efficiency via Preemptive Routing}

The fixed and decoupled structural prior $\mathbf{r}^*(\mathbf{X})$ provided by the frozen \GR\ effectively decouple the routing structure from representation learning. This stable routing decision simultaneously presents a unique opportunity to significantly enhance training efficiency. Unlike traditional methods~\cite{he2022fastermoe,nie2023flexmoe,liu2025netmoe,zhang2025comet} that must operate on dynamic, real-time routing decisions, \GR\ offers a preemptive, known routing map throughout the entire training process, enabling offline optimization and eliminating runtime overhead.

\subsubsection{Decoupling Preemptive Routing}
% 由于Grouter完全与模型解耦，因此我们不再进行训练前向传播时的路由，而是将路由任务放到数据预处理部分。具体来说，我们会提前将数据用Grouter网络处理，之后存储为数据文件，在模型前传时传入到各个MoE网络指导路由。由于对于每个token需存储激活专家数量个 uint8 类型和 torch.bfloat16数据，且可以处理一次后多次使用，因此Grouter提供了一种利用存储空间提升运行速度的途径。

{Given that the \GR\ is completely decoupled and fixed, we shift the routing task from the model's forward pass to the data preprocessing pipeline. Specifically, the input data is processed by the \GR\ network ahead of time, and the resulting routing decisions are then cached and stored as part of the processed dataset. During the model's forward pass, these stored decisions are loaded and passed directly to guide the $\text{MoE}$ layers.}

{This mechanism offers a trade-off between storage and computation. For each token, only the active expert indices  and their corresponding gating weights need to be stored. Critically, since the routing is computed only once but can be reused across multiple training epochs, the \GR\ provides an efficient pathway to leverage storage capacity for runtime acceleration.}

\subsubsection{Decoupling Communication Optimization}

Training large $\text{MoE}$ models often relies on Expert Parallelism ($\text{EP}$), which distributes experts across devices. This requires all-to-all communication during the token dispatch phase, where the dynamic nature of routing necessitates synchronous communication optimization.

Another key innovation of the \GR\ framework is to decouple this communication optimization from the synchronous training path as illustrated in \autoref{fig:pipeline_for_ep_save}. We define the routing affinity vector $\boldsymbol{\phi}(\mathbf{X}) \in \mathbb{R}^{\mathbf{E}_T}$ of a sample $\mathbf{X}$ as the average expert selection frequency across all tokens $t \in \mathbf{X}$:
\begin{equation}
    \boldsymbol{\phi}(\mathbf{X})_i = \frac{1}{|\mathbf{X}|} \sum_{t \in \mathbf{X}} \mathbb{I}\left[ e_i \in \text{TopK}(\mathbf{G}^*(\mathbf{x}_t), k) \right]
\end{equation}
where $\mathbb{I}[\cdot]$ is the indicator function, $\mathbf{e}_i$ is expert $i$ and $\mathbf{G}^*$ is the frozen \GR.

Leveraging $\mathbf{G}^*$, we collect the routing affinity vectors for all training samples to construct the full affinity set 
\begin{equation}
    \Phi= \{\boldsymbol{\phi}(\mathbf{X}_i) \mid \mathbf{X}_i \in \mathcal{D}_{\text{train}}\}
\end{equation}
 This set serves as the basis for the subsequent structural optimization.

\textbf{Expert Grouping.} The first step for optimization is to establish fixed expert groups that are co-located on physical devices. Each sample's affinity vector $\boldsymbol{\phi}(\mathbf{X})$ indicates its overall preference for the $E_T$ experts. We apply a clustering algorithm to the set of affinity vectors $\Phi$. The number of clusters is explicitly set to $\mathbf{N}_p$, which is the total number of Expert Parallel partitions. Subsequently, experts are assigned to these clusters with the objective of maximizing the aggregated affinity, thereby constructing $\mathbf{N}_p$ distinct expert groups:

\vspace{-4mm}
\begin{equation}
    \{\mathcal{E}_p\}_{p=1}^{N_p} = \operatorname{Clustering}(\Phi, N_p) \quad \mathcal{E}_p \subset \{1, \dots, E_T\}
\end{equation}
where $ \{1, \dots, E_T\}$ is the experts of Target MoE Model. Each cluster $\mathcal{E}_p \subseteq \{1, \dots, E_T\}$ represents an Expert Group, a subset of experts that are frequently co-activated by a specific type of input sample, which are subsequently mapped to the $\mathbf{N}_p$ physical $\text{EP}$ partitions.

\textbf{Sample Placement Optimization.} Once the Expert Groups are fixed, we determine the optimal placement for each input sample $\mathbf{X}_i$ by minimizing the communication cost. Specifically, the optimal partition ID is determined by:
\begin{equation}
  \text{PartitionID}(\mathbf{X}_i) = \underset{p \in \{1,\dots,N_p\}}{\arg\min} \left( \text{Cost}(\mathbf{X}_i, \mathcal{E}_p) \right)
\end{equation}
where $\text{Cost}(\mathbf{X}_i, \mathcal{E}_p)$ represents the communication cost incurred by assigning sample $\mathbf{X}_i$ to the partition containing Expert Group $\mathcal{E}_p$. This optimization yields $\mathbf{N}_p$ fixed Sample Groups, where each corresponds to one $\text{EP}$ partition. During training, samples are statically assigned to their corresponding $\text{EP}$ group based on $\text{PartitionID}(\mathbf{X}_i)$.

This optimization process fundamentally transforms the challenge of dynamic, runtime communication into a manageable, pre-computed resource allocation problem. By utilizing the fixed structural prior of \GR\ to define expert groups and sample groups, we effectively eliminate the high latency and overhead of synchronous communication optimization, leading to significant acceleration and stability in $\text{MoE}$ training.

\begin{figure*}[tp]
    \centering 
    \begin{subfigure}{0.48\textwidth}
        \caption{}
        \label{fig:pretrain_loss}
        \centering
        \includegraphics[width=0.9\linewidth]{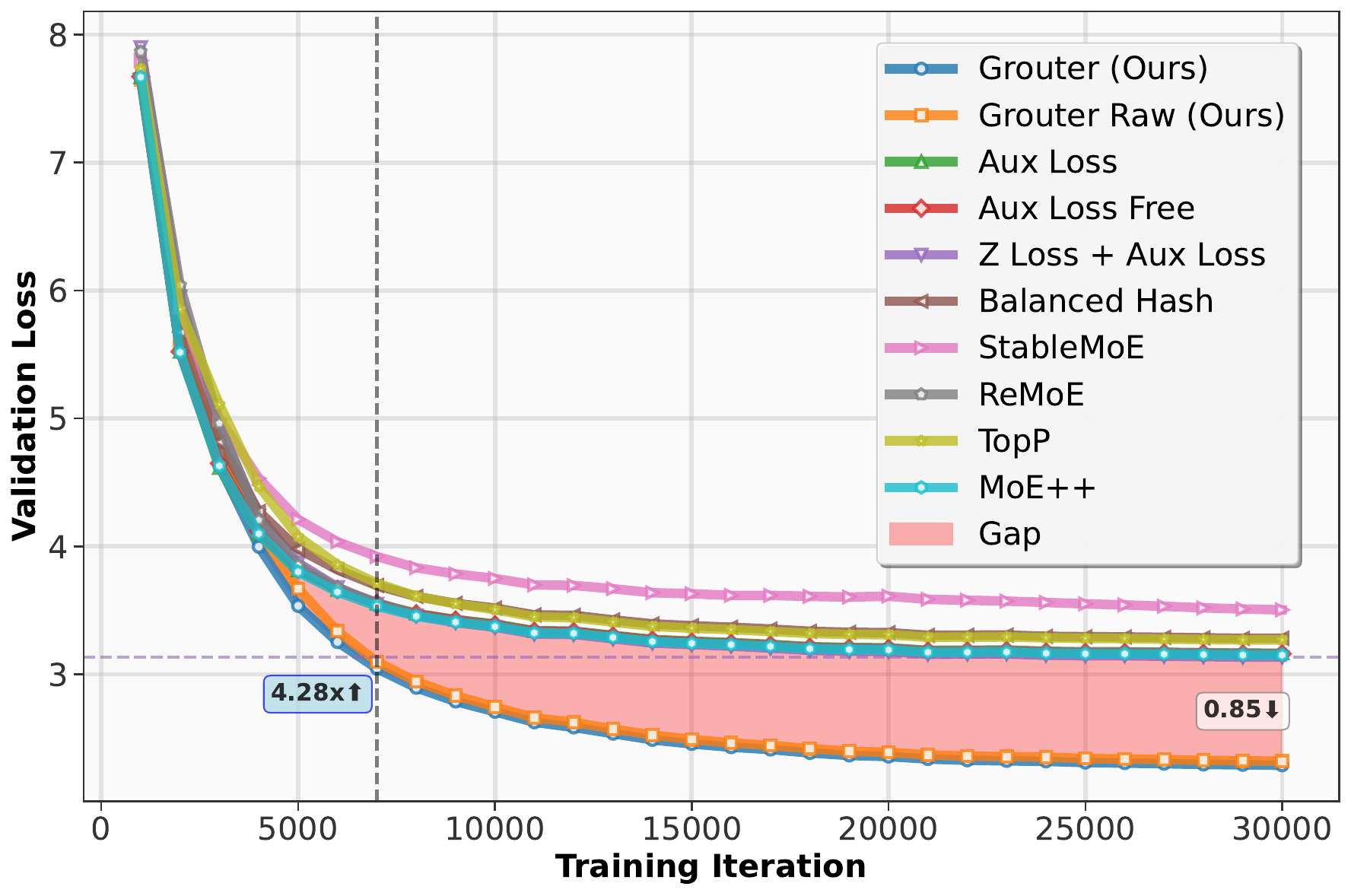}
    \end{subfigure}
    \hfill  
    \begin{subfigure}{0.48\textwidth}
        \caption{}
        \label{fig:load_balance}
        \centering
        \includegraphics[width=0.9\linewidth]{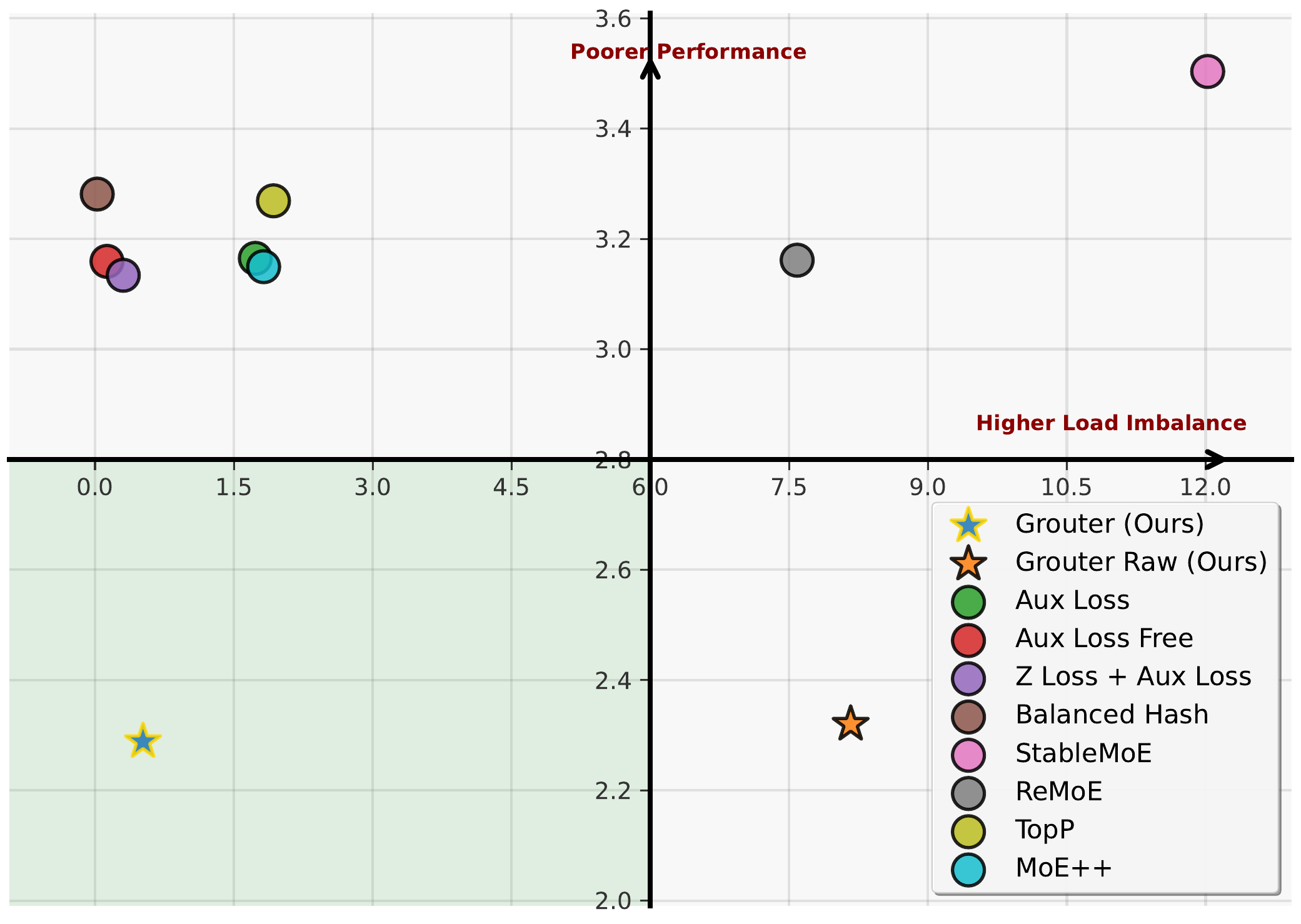}
    \end{subfigure}
    \caption{\small (a)  Pre-training Validation Loss Curves Across 30B Tokens. $\text{Grouter Raw}$ denotes the distilled $\text{Grouter}$ used without the subsequent Expert-Tuning. The shaded Gap region illustrates the loss difference between the $\text{Grouter}$ curve and the best baseline. Overall, $\text{Grouter}$ achieves a $4.28\times$ acceleration or a maximum loss reduction of $0.85$ at the same training data volume. (b) Comparison of Performance and Load Balance Trade-off. The third quadrant is highlighted in green to emphasize that models located within this region achieve an optimal balance between low load violation and superior model performance}
    \vspace{-4mm}
\end{figure*}

\section{Experiment}

\subsection{Experiment Setup}
\textbf{Infrastructure} Our experiments were conducted on a cluster comprising eight $\text{NVIDIA A}100$ GPUs and eight $\text{NVIDIA H}100$ GPUs, interconnected via $\text{NVLink}$. The software environment utilized $\text{PyTorch 2.8.0}$ with $\text{CUDA 12.9}$ and the $\text{Megatron-LM}$ framework (commit hash $\text{e7c55de9}$).

\textbf{Grouter Distillation Setup} We distill our \GR\ from the Qwen3-30B-A3B model~\cite{qwen3technicalreport}. The distillation process is performed on the C4 dataset\cite{raffel2020exploring} for $2.6$ Billion tokens. We construct the \GR\ using a three-layer Transformer Encoder, resulting in a total parameter count of $60$M. Notably, $50$M of these parameters reside in the computationally inexpensive Embedding layer, which makes the overall \GR\ architecture highly lightweight. The consistently decreasing loss curve, as shown in \autoref{apd:distill_loss}, demonstrates the successful extraction of the structural prior. Unless otherwise specified, all subsequent experiments are based on this single distilled \GR\ instance.

\textbf{Model Configurations} We define five distinct MoE model variants for our experiments: Tiny-Qwen3, Mini-Qwen3, Samll-Qwen3, Mini-DS-V2-Lite, and Mini-GPT-OSS. These models are structurally based on the architectures of Qwen3-30B-A3B~\cite{qwen3technicalreport}, DeepSeek-V2-Lite~\cite{deepseekv2}, and GPT-OSS-20B~\cite{openai2025gptoss120bgptoss20bmodel}, respectively. Their detailed configurations are presented in the \autoref{ap:configuration}.

\vspace{-4mm}

% \textbf{Training Setting} 我们在C4数据集上训练我们的模型，我们使用AdamW优化器，ZeRO1零冗余优化器，梯度检查点等技术也被使用以高效利用显存。

\begin{figure*}[tp]
    \centering 
    \begin{subfigure}{0.48\textwidth}
        \centering
        \includegraphics[width=0.65\linewidth]{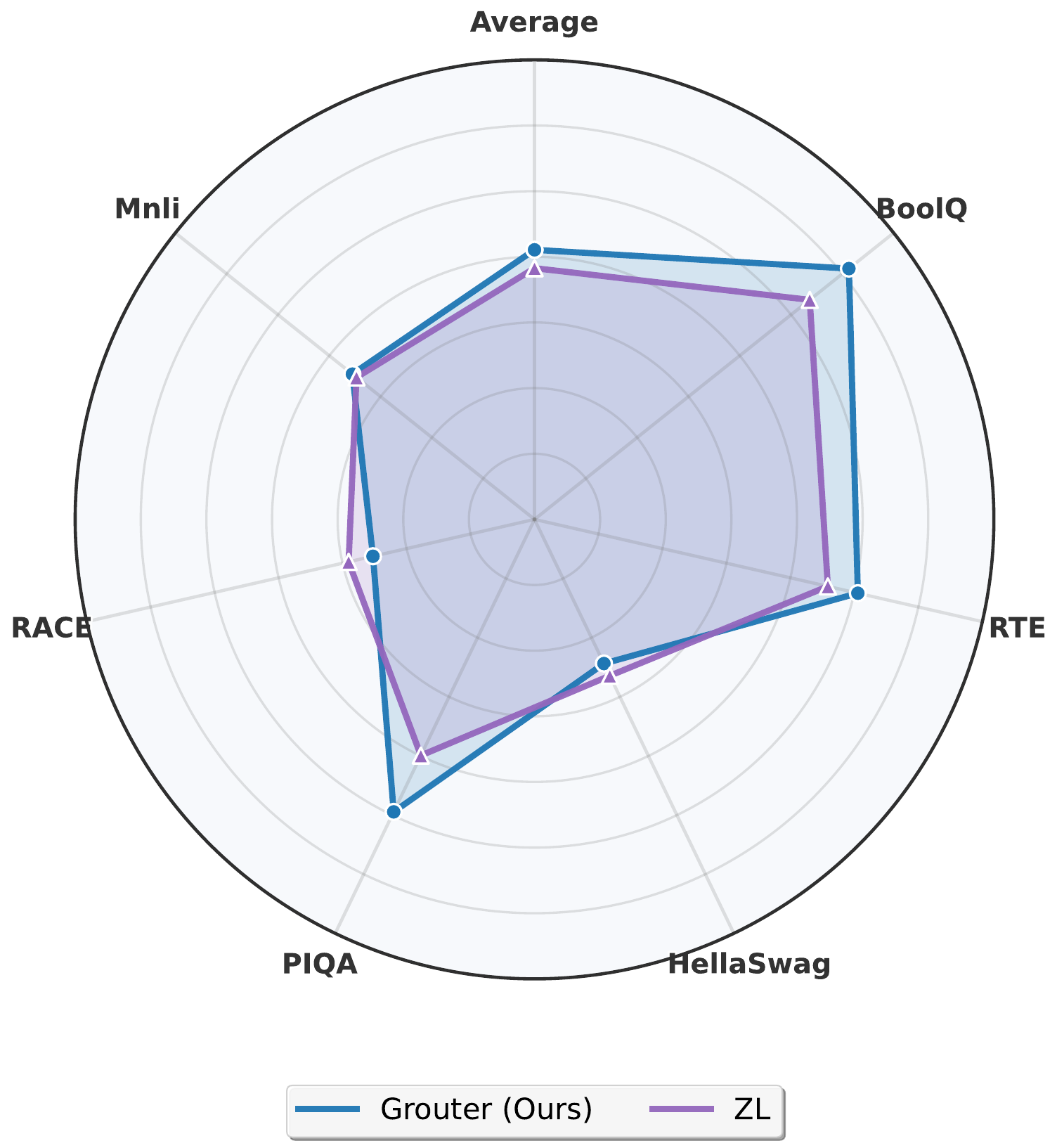}
        \caption{}
        \label{fig:downstream}
    \end{subfigure}
    \hfill  
    \begin{subfigure}{0.48\textwidth}
           \centering
        \includegraphics[width=1\linewidth]{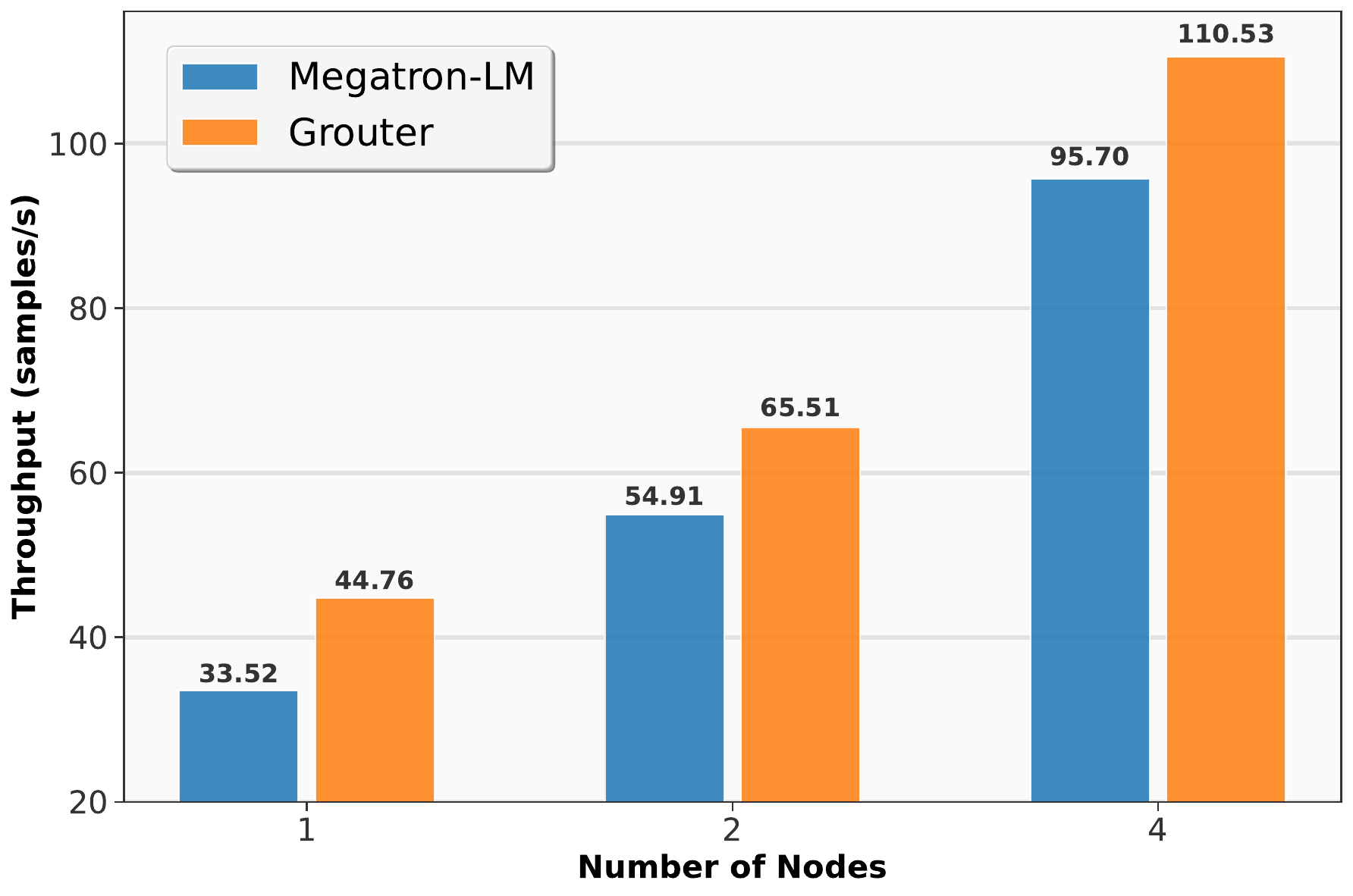}
        \caption{}
        \label{fig:throught}
    \end{subfigure}
    \vspace{-4mm}
    \caption{\small (a) Downstream Task Results: \GR\ achieves an average improvement of 2.80 across six benchmarks, with gains of up to 10 points on specific tasks. This demonstrates that the reduced validation loss achieved by \GR\ translates into genuine enhancements in model capabilities, rather than merely overfitting the validation metric. (b) Throughput Scaling: We evaluate throughput on 1, 2, and 4 nodes with Expert Parallelism (EP) degrees set to 8, 16, and 32, respectively. For the multi-node setups (2 and 4 nodes), we apply node-granularity communication optimization, while for the single-node case, we utilize GPU-granularity optimization.}
    \label{fig:insight_curve}
    \vspace{-3mm}
\end{figure*}

\subsection{Performance}

% 我们基于我们550m模型进行预训练实验，模型的参数设置可见附录，我们选用C4作为我们的数据集，这是大模型训练较常用的一个数据集。我们对每个模型进行30B token训练，根据Scaling law 这足以使模型收敛。我们使用验证集交叉熵损失来度量模型的表现，结果见图2，其中Grouter Raw为直接使用未经专家微调得到的Grouter进行预训练。结果显示，我们的Grouter可以使用23.3%的数据达到相同的收敛结果，即4.28倍加速，亦或者在相同的数据量下实现0.85的损失降低。且随着训练的进行，差距不断加大。这是因为我们的的Grouter规避了SPI的影响，使得原本随着训练不断累积的误差逐渐消失。且相对于HashMoE与StableMoE，Grouter提取到的稳态结构更加合理，因此显著提升了表现。此外，Grouter Raw和Grouter几乎没有差距，这说明我们的微调很好的保留了模型结构。

% \begin{figure}[htbp]
%     \centering
%     \includegraphics[width=0.9\linewidth]{figures/pretrain_loss.pdf}
%     \caption{\small Pre-training Validation Loss Curves Across 30B Tokens. $\text{Grouter Raw}$ denotes the distilled $\text{Grouter}$ used without the subsequent Expert-Tuning. The shaded Gap region illustrates the loss difference between the $\text{Grouter}$ curve and the best baseline. Overall, $\text{Grouter}$ achieves a $4.28\times$ acceleration or a maximum loss reduction of $0.85$ at the same training data volume.}
%     \label{fig:pretrain_loss}
%     \vspace{-3mm}
% \end{figure}

%We conduct pre-training experiments based on our Tiny-Qwen3 model. The detailed model architecture, experimental hyperparameter settings, and the construction details of the \GR\ are provided in \autoref{ap:Experiment Setup}. We use the C4 dataset, a standard corpus widely employed in large language model training. Each model is trained for $30$ Billion tokens, which is enough to achieve convergence according to the compute-optimal dataset size predicted by \cite{krajewski2024scaling}. 

Model performance is measured using the validation set cross-entropy loss, with the results presented in \autoref{fig:pretrain_loss}. Each model is trained for $30$ Billion tokens, which is enough to achieve convergence according to the compute-optimal dataset size predicted by \citet{krajewski2024scaling}. We distinguish between two \GR\ variants: $\textbf{Grouter Raw}$, which denotes the direct use of the distilled \GR\ without the subsequent load-balancing expert tuning, and $\textbf{Grouter}$,which refers to the version subjected to this fine-tuning on only $50$M tokens of the C4 dataset. The implementation details and specific configurations of the other baselines are provided in \autoref{ap:imple of baseline}.

%Our results demonstrate significant efficiency gains: \GR\ achieves the same convergence loss by leveraging only \textbf{23.3$\%$} of the training data, which corresponds to a \textbf{4.28$\times$} acceleration. 这展示了Grouter在提升了模型的收敛速度。Furthermore, when compared at the equivalent training volume, \GR\ exhibits superior final performance, achieving a loss reduction of up to $0.85$. %这说明Grouter不仅不会限制模型，反而可以帮助模型实现更好的收敛精度。We also observe that the gap consistently widens as training progresses. %我们认为这是因为Grouter彻底消除了router和representation同步学习带来的干扰带来的收益。

Our results demonstrate significant efficiency gains: \GR\ achieves the same convergence loss by leveraging only \textbf{23.3$\%$} of the training data, which corresponds to a \textbf{4.28$\times$} acceleration in convergence speed. Furthermore, when compared at the equivalent training volume, \GR\ exhibits superior final performance, achieving a loss reduction of up to $0.85$. This indicates that, rather than constraining the model capacity, the fixed structure of \GR\ actually facilitates superior convergence accuracy. We also observe that the performance gap consistently widens as training progresses. We attribute this sustained advantage to the fact that \GR\ completely eliminates the optimization interference arising from the synchronous learning of routing policies and representations. Finally, the marginal difference between $\textbf{Grouter Raw}$ and the fine-tuned $\textbf{Grouter}$ validates that our lightweight fine-tuning strategy successfully retains the high-quality structural information initially extracted from the teacher model.

%This sustained improvement is primarily attributed to $\text{Grouter}$'s ability to mitigate the detrimental effects of $\text{SPI}$, effectively eliminating the optimization error that typically accumulates with conventional dynamic routing. Furthermore, the $\text{Grouter}$ extracts a more structurally sound and stable prior compared to $\text{HashMoE}$ and $\text{StableMoE}$, leading to a notable performance enhancement.

% 之前的研究表明，过度追求负载均衡有可能会损失模型表现，因此我们进一步展示我们的方法在performance和load balance之间的trade off。我们对所有经过30B token训练的MoE测量MaxVio指标，这一指标可以很好的度量负载均衡水平，仍然使用验证集合损失作为表现的度量。结果显示在所有对比的方法里面，我们的Grouter方法并未因为提升表现过多的牺牲负载均衡。同时可以看到Grouter Raw有严重的负载不均衡，这也验证了我们之前在\ref{Expert Deviation}部分提到的Source Model和Target Model之间部署数据集分布的差异，导致蒸馏到的Grouter可能出现严重的负载不均衡。

Prior research suggests that excessive pursuit of load balancing can compromise model performance\cite{wang2024auxiliary}. In this context, we show that $\text{Grouter}$ achieves an excellent balance, significantly enhancing performance without unduly sacrificing load stability.

To quantify this, we measure the $\text{MaxVio}_{\text{Global}}$ metric defined in \autoref{Maximal Violation} for all $\text{MoE}$ models after $30\text{B}$ training tokens.  The validation set loss is used as the complementary measure of performance. \textbf{The results, displayed in \autoref{fig:load_balance}, confirm that \GR\ maintains competitive load balance despite its superior performance gains over all compared methods.} Significantly, the $\textbf{Grouter Raw}$ variant simultaneously exhibits severe load imbalance. This finding directly corroborates our discussion in Section \ref{chapter:expert_tuning}, validating that the disparity in data distributions between the source model and the target model may cause significant load imbalance observed in the directly distilled \GR.

% \begin{figure}[htbp]
%     \centering
%     \includegraphics[width=0.9\linewidth]{figures/load_balance_scatter.pdf}
%     \vspace{-2mm}
%     \caption{\small Comparison of Performance and Load Balance Trade-off. The third quadrant is highlighted in green to emphasize that models located within this region achieve an optimal balance between low load violation and superior model performance
%     }
%     \vspace{-2mm}
%     \label{fig:load_balance}
% \end{figure}

% 为了展示Grouter可以迁徙到不同配置的模型上，我们在三种不同尺寸，不同架构，不同专家配置的模型上验证了我们的Grouter的扩展性。其中Mini-Qwen3全部由MoE层构成，具有128个专家和8个激活专家，Mini-DS-V2-Lite参照了DeepSeek V2 Lite的结构，其特征包含Dense层与MoE层的混合以及Multi latent attention的使用，Mini-GPT-OSS借鉴了GPT-OSS的结构。这三个模型拥有不同数量的专家总数和激活专家数，我们测试我们的\GR\和预训练实验的基线SOTA，Z Loss + Aux Loss在经过充分训练后的验证集损失。我们通过expert folding技术，将亲和的专家折叠为同一个专家，实现了Grouter在不同专家场景的适配，之后使用expert tuning技术，使得在不同激活专家数量下，Grouter达到足够负载均衡。之后与Z Loss + Aux Loss, 缩写为ZL,在不同模型配置下比较最终的验证集损失。实验的结果显示在\autoref{tab:model_comparition}。实验结果显示\textbf{Grouter}在不同尺寸都展现出来了有效的训练加速。这说明了我们的expert folding和expert tuning方案的有效性。

To demonstrate that \GR\ can be effectively transferred across different model configurations, we validated its scalability on four models varying in size, architecture, and expert setup. We first utilize expert folding or expert expanding to adapt the structural prior to target models with fewer or more experts than the source, respectively. Subsequently, the expert tuning technique is applied to ensure $\text{Grouter}$ achieves adequate load balance across varying numbers of active experts. We compare the final validation set loss of our $\text{Grouter}$ against the $\text{SOTA}$ baseline from our pre-training experiments , Z-Loss + Aux Loss (abbreviated as $\text{ZL}$), after sufficient training. The results, presented in \autoref{tab:model_comparison}, clearly show that \GR\ demonstrates effective training acceleration across all tested scales and architectures. \textbf{This compelling outcome strongly validates the efficacy of our expert folding, expert expanding and expert tuning adaptation schemes.} The complete training curves for these experiments are provided in \autoref{ap:Training curves under different configurations}.

\begin{table*}[htbp]
    \centering
    \caption{\small Performance Comparison Across Diverse Model Configurations}
    \small
    %\caption{\ky{Table caption}}
    \setlength{\tabcolsep}{15pt}
    \begin{tabular}{lcccc}
        \toprule[1.2pt]
        & \textbf{Mini-GPT-OSS} & \textbf{Mini-DS-V2-Lite} & \textbf{Mini-Qwen3} & \textbf{Small-Qwen3} \\
        \midrule
        Attn           & GQA   & MLA           & GQA  & GQA \\ 
        Structure      & MoE           & Dense+MoE     & MoE & MoE\\
        Tokens  & 30B           & 30B           & 50B & 50B \\
        Grouter    & \textbf{2.940}    & \textbf{2.469}    & \textbf{1.763} & \textbf{1.648}\\
        ZL     & 3.346    & 3.057    & 2.740  & 2.633\\ 
        \bottomrule[1.2pt]
    \end{tabular}
    \label{tab:model_comparison} 
\end{table*}

% 我们使用Grouter和ZL 经过50B训练的Mini-Qwen3 进行了下游任务的评估，我们选用了BoolQ,Rte,HellaSwag,PIQA,Race,Mnli数据集作为评估集，评估的结果见\autoref{}。结果显示，Grouter在评估任务中由于Z Loss + Aux Loss,这说明我们的方法也提升了模型对于问题的解决能力。

We evaluated the $\text{Mini-Qwen3}$ model, pre-trained for $50$B tokens using both \GR\ and the $\text{ZL}$ baseline, on a suite of downstream tasks.The detailed experimental setup is described in \autoref{apdix:downstream}, and the evaluation results are presented in \autoref{fig:downstream}.

The results show that $\text{Grouter}$ consistently outperforms the $\text{ZL}$ baseline across these evaluation tasks. \textbf{This finding strongly suggests that our method not only enhances training efficiency but also improves the model's overall problem-solving capability and generalization performance.}

% \begin{figure}[htbp]
%     \centering
%     \includegraphics[width=0.7\linewidth]{figures/downstream_task_radar.pdf}
%     \vspace{-2mm}
%     \caption{\small Downstream Task Evaluation Results.
%     }
%     \vspace{-7mm}
%     \label{fig:downstream}
% \end{figure}

\vspace{-2mm}

\subsection{Efficiency}

% 接下来，我们利用Grouter获取的先验知识，进行Router解耦和数据重组，以提升模型的训练效率。我们首先基于\GR\获取C4数据集的专家指派信息，之后我们先后进行Expert Grouping和Sample Placement Optimization，尽可能节省EP通信。我们基于Mini-Qwen3进行通信节省实验，所有的实验均基于EP通信节省系统DeepEP\cite{liu2024deepseek}的基础上,这显示我们的方法可以与当下最先进的EP通信技术之一结合加强Efficiency。我们测量了基于Megatron-LM的运行的Aux Loss作为基线，实验结果见\autoref{fig:throughtput}.通过观察可以看到在不同的EP数量下，Grouter均有一定的通信节省。

Next, we leverage the prior knowledge acquired by \GR\ to enhance model training efficiency through router decoupling and data reorganization. Specifically, we first utilize \GR\ to obtain expert assignment information for the $\text{C4}$ dataset. Subsequently, we sequentially apply Expert Grouping and Sample Placement Optimization to maximize the reduction in $\text{EP}$ communication overhead.

We conducted the communication saving experiments using the $\text{Mini-Qwen3}$ model. Crucially, all experiments were integrated with $\text{DeepEP}$ \cite{liu2024deepseek}, one of the most advanced $\text{EP}$ communication saving systems. This integration demonstrates that our methodology can effectively combine with state-of-the-art $\text{EP}$ communication techniques to further boost training efficiency.

% \begin{figure}[htbp]
%     \centering
%     \includegraphics[width=0.8\linewidth]{figures/throughput.pdf}
%     \vspace{-2mm}
%     \caption{\small Throughtput experiment for different nodes.
%     }
%     \vspace{-6mm}
%     \label{fig:throught}
% \end{figure}

We measure the throughput against a baseline using the standard $\text{Auxiliary Loss}$ run on the $\text{Megatron-LM}$ framework, specifically calculating the throughput averaged over the first $1000$ training iterations. The experimental results, presented in \autoref{fig:throught}, show that $\text{Grouter}$ consistently achieves notable communication savings across various $\text{EP}$ counts. Specifically, we observe throughput speedups of $33.5\%$, $19.3\%$, and $15.5\%$ on $1$, $2$, and $4$ nodes, respectively. These improvements demonstrate that \GR\ significantly contributes to achieving higher training efficiency for $\text{MoE}$ models. The detailed experimental settings is in \autoref{Efficiency Experiment Detail}.

\vspace{-4mm}

\subsection{Discussion}

% 在这一部分我们讨论为什么\GR\可以展现出如此良好的表现。我们度量了训练过程中的gradient norm的波动程度，结果见图1。我们观察到使用\GR\进行训练时，模型gradient norm的变异系数极其平滑，训练过程中没有任何spike，而对于aux loss和HashLayer这样的方法，均频繁出现gradient norm的巨幅波动。我们将其解释为这是在这些方法下，专家有时会接收到了不熟悉的token，这些token会导致变化异常大的梯度，进而破坏专家已经学习到的结构。而\GR\已经预先构建好了专家指派结构，因此专家持续在给定任务上特化，极大的提升了学习效率，实现了稳定的训练。我们在appendix 1中展开更详细的讨论。

 We investigate the factors contributing to the superior performance of \GR. We quantify the volatility of the training process by measuring the gradient norms, as illustrated in \autoref{fig:grad_norm_cv}. We observe that \GR\ exhibits an exceptionally stable coefficient of variation for the gradient norm, devoid of any spikes throughout the training phase. In contrast, methods such as Aux Loss, Aux Loss Free and HashLayer suffer from frequent, high-magnitude fluctuations. We attribute this phenomenon to the routing instability in baseline methods, where experts occasionally encounter unfamiliar tokens. These tokens generate anomalous gradients that disrupt the expert's learned representations. Conversely, by establishing a structural prior for expert assignment, \GR\ facilitates continuous expert specialization on designated tasks, thereby significantly enhancing learning efficiency and ensuring training stability. A more detailed discussion is provided in \autoref{ap:Discussion on the Efficient Convergence Achieved by Grouter}.
 
\vspace{-2mm}

\section{Conclusion}

We have presented \GR, a preemptive routing framework that decouples routing structure optimization from representation learning in MoE training. By distilling high-quality routing decisions from a converged source model into a lightweight standalone network, \GR\ provides a fixed structural prior that eliminates the instability arising from joint router-expert optimization. Combined with expert folding, expert expanding, and expert tuning, a single distilled \GR\ instance adapts flexibly to diverse model scales and expert configurations. Extensive experiments demonstrate that \GR\ achieves a $4.28\times$ improvement in data utilization efficiency and up to $33.5\%$ throughput acceleration, while maintaining superior convergence quality across varying architectures. These results establish preemptive routing as a practical and effective paradigm for scalable MoE pre-training.

\vspace{-2mm}

\section{Future Works}

\GR's ability to pre-acquire $\text{Dispatch}$ information enables its integration with various communication saving and training acceleration methods. The deterministic routing provided by the frozen \GR\ also offers natural resilience to hardware failures, presenting a promising new avenue for algorithmic fault tolerance as explored by \citet{hu2026mecefo}. Furthermore, this stability is crucial for Reinforcement Learning, where the expert-activation volatility of MoE models can prevent RL training from converging properly~\cite{zheng2025group, ma2025stabilizing}. Since \GR\ is frozen during training, it bypasses this critical issue. We leave the detailed exploration of both these synergistic efficiency gains and post-training applications for future work.

\section*{Acknowledgments}

This work is funded by the National Key Research and Development Program of China (No. 2024YFA1012902) and the National Natural Science Foundation of China (No. 92370121, 12301392, 12288101, W2441021). This research is also  supported by Zhejiang Lab and the AI for Science Institute, Beijing, China.

\section*{Impact Statement}

This research focuses on optimizing the training dynamics of sparse models. By decoupling routing from representation, our approach lowers the data and computational barriers for training high-performance MoE models, potentially democratizing access to large-scale AI capabilities. The proposed method is a general optimization technique and does not introduce specific potential for societal harm.


\bibliography{bibtex}

@article{lepikhin2020gshard,
  title={Gshard: Scaling giant models with conditional computation and automatic sharding},
  author={Lepikhin, Dmitry and Lee, HyoukJoong and Xu, Yuanzhong and Chen, Dehao and Firat, Orhan and Huang, Yanping and Krikun, Maxim and Shazeer, Noam and Chen, Zhifeng},
  journal={arXiv preprint arXiv:2006.16668},
  year={2020}
}

@article{wang2024auxiliary,
  title={Auxiliary-loss-free load balancing strategy for mixture-of-experts},
  author={Wang, Lean and Gao, Huazuo and Zhao, Chenggang and Sun, Xu and Dai, Damai},
  journal={arXiv preprint arXiv:2408.15664},
  year={2024}
}

@article{liu2024deepseek,
  title={Deepseek-v2: A strong, economical, and efficient mixture-of-experts language model},
  author={Liu, Aixin and Feng, Bei and Wang, Bin and Wang, Bingxuan and Liu, Bo and Zhao, Chenggang and Dengr, Chengqi and Ruan, Chong and Dai, Damai and Guo, Daya and others},
  journal={arXiv preprint arXiv:2405.04434},
  year={2024}
}

@article{vaswani2017attention,
  title={Attention is all you need},
  author={Vaswani, Ashish and Shazeer, Noam and Parmar, Niki and Uszkoreit, Jakob and Jones, Llion and Gomez, Aidan N and Kaiser, {\L}ukasz and Polosukhin, Illia},
  journal={Advances in neural information processing systems},
  volume={30},
  year={2017}
}

@article{liu2019roberta,
  title={Roberta: A robustly optimized bert pretraining approach},
  author={Liu, Yinhan and Ott, Myle and Goyal, Naman and Du, Jingfei and Joshi, Mandar and Chen, Danqi and Levy, Omer and Lewis, Mike and Zettlemoyer, Luke and Stoyanov, Veselin},
  journal={arXiv preprint arXiv:1907.11692},
  year={2019}
}

@article{bahdanau2014neural,
  title={Neural machine translation by jointly learning to align and translate},
  author={Bahdanau, Dzmitry and Cho, Kyunghyun and Bengio, Yoshua},
  journal={arXiv preprint arXiv:1409.0473},
  year={2014}
}

@article{brown2020language,
  title={Language models are few-shot learners},
  author={Brown, Tom and Mann, Benjamin and Ryder, Nick and Subbiah, Melanie and Kaplan, Jared D and Dhariwal, Prafulla and Neelakantan, Arvind and Shyam, Pranav and Sastry, Girish and Askell, Amanda and others},
  journal={Advances in neural information processing systems},
  volume={33},
  pages={1877--1901},
  year={2020}
}

@article{kaplan2020scaling,
  title={Scaling laws for neural language models},
  author={Kaplan, Jared and McCandlish, Sam and Henighan, Tom and Brown, Tom B and Chess, Benjamin and Child, Rewon and Gray, Scott and Radford, Alec and Wu, Jeffrey and Amodei, Dario},
  journal={arXiv preprint arXiv:2001.08361},
  year={2020}
}

@article{shazeer2017outrageously,
  title={Outrageously large neural networks: The sparsely-gated mixture-of-experts layer},
  author={Shazeer, Noam and Mirhoseini, Azalia and Maziarz, Krzysztof and Davis, Andy and Le, Quoc and Hinton, Geoffrey and Dean, Jeff},
  journal={arXiv preprint arXiv:1701.06538},
  year={2017}
}

@inproceedings{du2022glam,
  title={Glam: Efficient scaling of language models with mixture-of-experts},
  author={Du, Nan and Huang, Yanping and Dai, Andrew M and Tong, Simon and Lepikhin, Dmitry and Xu, Yuanzhong and Krikun, Maxim and Zhou, Yanqi and Yu, Adams Wei and Firat, Orhan and others},
  booktitle={International conference on machine learning},
  pages={5547--5569},
  year={2022},
  organization={PMLR}
}

@article{wang2024remoe,
  title={Remoe: Fully differentiable mixture-of-experts with relu routing},
  author={Wang, Ziteng and Zhu, Jun and Chen, Jianfei},
  journal={arXiv preprint arXiv:2412.14711},
  year={2024}
}

@article{zhong2024lory,
  title={Lory: Fully differentiable mixture-of-experts for autoregressive language model pre-training},
  author={Zhong, Zexuan and Xia, Mengzhou and Chen, Danqi and Lewis, Mike},
  journal={arXiv preprint arXiv:2405.03133},
  year={2024}
}

@inproceedings{yan2025tc,
  title={TC-MoE: Augmenting Mixture of Experts with Ternary Expert Choice},
  author={Yan, Shen and Bin, Xingyan and Zhang, Sijun and Wang, Yisen and Lin, Zhouchen},
  booktitle={The Thirteenth International Conference on Learning Representations},
  year={2025}
}

@article{jin2024moe++,
  title={Moe++: Accelerating mixture-of-experts methods with zero-computation experts},
  author={Jin, Peng and Zhu, Bo and Yuan, Li and Yan, Shuicheng},
  journal={arXiv preprint arXiv:2410.07348},
  year={2024}
}

@article{roller2021hash,
  title={Hash layers for large sparse models},
  author={Roller, Stephen and Sukhbaatar, Sainbayar and Weston, Jason and others},
  journal={advances in neural information processing systems},
  volume={34},
  pages={17555--17566},
  year={2021}
}

@article{dai2022stablemoe,
  title={Stablemoe: Stable routing strategy for mixture of experts},
  author={Dai, Damai and Dong, Li and Ma, Shuming and Zheng, Bo and Sui, Zhifang and Chang, Baobao and Wei, Furu},
  journal={arXiv preprint arXiv:2204.08396},
  year={2022}
}

@article{su2024roformer,
  title={Roformer: Enhanced transformer with rotary position embedding},
  author={Su, Jianlin and Ahmed, Murtadha and Lu, Yu and Pan, Shengfeng and Bo, Wen and Liu, Yunfeng},
  journal={Neurocomputing},
  volume={568},
  pages={127063},
  year={2024},
  publisher={Elsevier}
}

@inproceedings{liu2025netmoe,
  title={Netmoe: Accelerating moe training through dynamic sample placement},
  author={Liu, Xinyi and Wang, Yujie and Fu, Fangcheng and Miao, Xupeng and Zhu, Shenhan and Nie, Xiaonan and Cui, Bin},
  booktitle={The Thirteenth International Conference on Learning Representations},
  year={2025}
}

@article{nie2023flexmoe,
  title={Flexmoe: Scaling large-scale sparse pre-trained model training via dynamic device placement},
  author={Nie, Xiaonan and Miao, Xupeng and Wang, Zilong and Yang, Zichao and Xue, Jilong and Ma, Lingxiao and Cao, Gang and Cui, Bin},
  journal={Proceedings of the ACM on Management of Data},
  volume={1},
  number={1},
  pages={1--19},
  year={2023},
  publisher={ACM New York, NY, USA}
}

@article{cai2024textit,
  title={$\textit{Read-ME}$: Refactorizing LLMs as Router-Decoupled Mixture of Experts with System Co-Design},
  author={Cai, Ruisi and Ro, Yeonju and Kim, Geon-Woo and Wang, Peihao and Ehteshami Bejnordi, Babak and Akella, Aditya and Wang, Zhangyang and others},
  journal={Advances in Neural Information Processing Systems},
  volume={37},
  pages={116126--116148},
  year={2024}
}

@article{zhang2025comet,
  title={Comet: Fine-grained computation-communication overlapping for mixture-of-experts},
  author={Zhang, Shulai and Zheng, Ningxin and Lin, Haibin and Jiang, Ziheng and Bao, Wenlei and Jiang, Chengquan and Hou, Qi and Cui, Weihao and Zheng, Size and Chang, Li-Wen and others},
  journal={arXiv preprint arXiv:2502.19811},
  year={2025}
}

@inproceedings{he2022fastermoe,
  title={Fastermoe: modeling and optimizing training of large-scale dynamic pre-trained models},
  author={He, Jiaao and Zhai, Jidong and Antunes, Tiago and Wang, Haojie and Luo, Fuwen and Shi, Shangfeng and Li, Qin},
  booktitle={Proceedings of the 27th ACM SIGPLAN Symposium on Principles and Practice of Parallel Programming},
  pages={120--134},
  year={2022}
}

@misc{qwen3technicalreport,
      title={Qwen3 Technical Report}, 
      author={Qwen Team},
      year={2025},
      eprint={2505.09388},
      archivePrefix={arXiv},
      primaryClass={cs.CL},
      url={https://arxiv.org/abs/2505.09388}, 
}

@misc{deepseekv2,
      title={DeepSeek-V2: A Strong, Economical, and Efficient Mixture-of-Experts Language Model}, 
      author={DeepSeek-AI},
      year={2024},
      eprint={2405.04434},
      archivePrefix={arXiv},
      primaryClass={cs.CL}
}

@misc{openai2025gptoss120bgptoss20bmodel,
      title={gpt-oss-120b \& gpt-oss-20b Model Card}, 
      author={OpenAI},
      year={2025},
      eprint={2508.10925},
      archivePrefix={arXiv},
      primaryClass={cs.CL},
      url={https://arxiv.org/abs/2508.10925}, 
}

@article{raffel2020exploring,
  title={Exploring the limits of transfer learning with a unified text-to-text transformer},
  author={Raffel, Colin and Shazeer, Noam and Roberts, Adam and Lee, Katherine and Narang, Sharan and Matena, Michael and Zhou, Yanqi and Li, Wei and Liu, Peter J},
  journal={Journal of machine learning research},
  volume={21},
  number={140},
  pages={1--67},
  year={2020}
}

@article{krajewski2024scaling,
  title={Scaling laws for fine-grained mixture of experts},
  author={Krajewski, Jakub and Ludziejewski, Jan and Adamczewski, Kamil and Pi{\'o}ro, Maciej and Krutul, Micha{\l} and Antoniak, Szymon and Ciebiera, Kamil and Kr{\'o}l, Krystian and Odrzyg{\'o}{\'z}d{\'z}, Tomasz and Sankowski, Piotr and others},
  journal={arXiv preprint arXiv:2402.07871},
  year={2024}
}

@article{zoph2022st,
  title={St-moe: Designing stable and transferable sparse expert models},
  author={Zoph, Barret and Bello, Irwan and Kumar, Sameer and Du, Nan and Huang, Yanping and Dean, Jeff and Shazeer, Noam and Fedus, William},
  journal={arXiv preprint arXiv:2202.08906},
  year={2022}
}

@article{clark2019boolq,
  title={Boolq: Exploring the surprising difficulty of natural yes/no questions},
  author={Clark, Christopher and Lee, Kenton and Chang, Ming-Wei and Kwiatkowski, Tom and Collins, Michael and Toutanova, Kristina},
  journal={arXiv preprint arXiv:1905.10044},
  year={2019}
}

@article{zellers2019hellaswag,
  title={Hellaswag: Can a machine really finish your sentence?},
  author={Zellers, Rowan and Holtzman, Ari and Bisk, Yonatan and Farhadi, Ali and Choi, Yejin},
  journal={arXiv preprint arXiv:1905.07830},
  year={2019}
}

@inproceedings{bisk2020piqa,
  title={Piqa: Reasoning about physical commonsense in natural language},
  author={Bisk, Yonatan and Zellers, Rowan and Gao, Jianfeng and Choi, Yejin and others},
  booktitle={Proceedings of the AAAI conference on artificial intelligence},
  volume={34},
  number={05},
  pages={7432--7439},
  year={2020}
}

@article{williams2017broad,
  title={A broad-coverage challenge corpus for sentence understanding through inference},
  author={Williams, Adina and Nangia, Nikita and Bowman, Samuel R},
  journal={arXiv preprint arXiv:1704.05426},
  year={2017}
}

@article{lai2017race,
  title={Race: Large-scale reading comprehension dataset from examinations},
  author={Lai, Guokun and Xie, Qizhe and Liu, Hanxiao and Yang, Yiming and Hovy, Eduard},
  journal={arXiv preprint arXiv:1704.04683},
  year={2017}
}

@article{wang2018glue,
  title={GLUE: A multi-task benchmark and analysis platform for natural language understanding},
  author={Wang, Alex and Singh, Amanpreet and Michael, Julian and Hill, Felix and Levy, Omer and Bowman, Samuel R},
  journal={arXiv preprint arXiv:1804.07461},
  year={2018}
}

@misc{caohuanqi2025lplb,
  author       = {Huanqi Cao},
  title        = {{Linear-Programming-Based Load Balancer (LPLB)}: A linear programming-based load balancing tool for Mixture-of-Experts (MoE) expert parallelism},
  year         = {2025},
  howpublished = {Unpublished software repository, DeepSeek AI},
  version      = {0.1.0},
  note         = {Accessed: 2025-XX-XX [Replace with your access date]},
  url          = {https://github.com/deepseek-ai/LPLB} 
}

@article{shoeybi2019megatron,
  title={Megatron-lm: Training multi-billion parameter language models using model parallelism},
  author={Shoeybi, Mohammad and Patwary, Mostofa and Puri, Raul and LeGresley, Patrick and Casper, Jared and Catanzaro, Bryan},
  journal={arXiv preprint arXiv:1909.08053},
  year={2019}
}

@article{zheng2025group,
  title={Group sequence policy optimization},
  author={Zheng, Chujie and Liu, Shixuan and Li, Mingze and Chen, Xiong-Hui and Yu, Bowen and Gao, Chang and Dang, Kai and Liu, Yuqiong and Men, Rui and Yang, An and others},
  journal={arXiv preprint arXiv:2507.18071},
  year={2025}
}

@article{ma2025stabilizing,
  title={Stabilizing moe reinforcement learning by aligning training and inference routers},
  author={Ma, Wenhan and Zhang, Hailin and Zhao, Liang and Song, Yifan and Wang, Yudong and Sui, Zhifang and Luo, Fuli},
  journal={arXiv preprint arXiv:2510.11370},
  year={2025}
}

@techreport{arthur2006k,
  title={k-means++: The advantages of careful seeding},
  author={Arthur, David and Vassilvitskii, Sergei},
  year={2006},
  institution={Stanford}
}

@misc{hastie2009elements,
  title={The elements of statistical learning: data mining, inference, and prediction},
  author={Hastie, Trevor},
  year={2009},
  publisher={Springer}
}

@book{golub2013matrix,
  title={Matrix computations},
  author={Golub, Gene H and Van Loan, Charles F},
  year={2013},
  publisher={JHU press}
}

@article{wei2026team,
  title={TEAM: Temporal-Spatial Consistency Guided Expert Activation for MoE Diffusion Language Model Acceleration},
  author={Wei, Linye and Luo, Zixiang and Tang, Pingzhi and Li, Meng},
  journal={arXiv preprint arXiv:2602.08404},
  year={2026}
}

@article{liu2026leveraging,
  title={Leveraging Error Diversity in Group Rollouts for Reinforcement Learning},
  author={Liu, Wenpu and Xu, Yuqi and Xie, Weichu and Zhu, Yongfu and Dong, Shuai and Wang, Ziyue and Shao, Wenqi and Zhang, Xiaoying and Yang, Tong and Duan, Nan and others},
  journal={arXiv preprint arXiv:2605.17333},
  year={2026}
}

@article{dong2025interleaved,
  title={Interleaved latent visual reasoning with selective perceptual modeling},
  author={Dong, Shuai and Wang, Siyuan and Liu, Xingyu and Li, Chenglin and Hou, Haowen and Wei, Zhongyu},
  journal={arXiv preprint arXiv:2512.05665},
  year={2025}
}

@misc{gu2026elasticmoeunlockinginferencetime,
      title={Elastic MoE: Unlocking the Inference-Time Scalability of Mixture-of-Experts}, 
      author={Naibin Gu and Zhenyu Zhang and Yuchen Feng and Yilong Chen and Peng Fu and Zheng Lin and Shuohuan Wang and Yu Sun and Hua Wu and Weiping Wang and Haifeng Wang},
      year={2026},
      eprint={2509.21892},
      archivePrefix={arXiv},
      primaryClass={cs.CL},
      url={https://arxiv.org/abs/2509.21892}, 
}

@misc{yang2026selfdistilledrlvr,
      title={Self-Distilled RLVR}, 
      author={Chenxu Yang and Chuanyu Qin and Qingyi Si and Minghui Chen and Naibin Gu and Dingyu Yao and Zheng Lin and Weiping Wang and Jiaqi Wang and Nan Duan},
      year={2026},
      eprint={2604.03128},
      archivePrefix={arXiv},
      primaryClass={cs.LG},
      url={https://arxiv.org/abs/2604.03128}, 
}

@misc{qin2026nearfuturepolicyoptimization,
      title={Near-Future Policy Optimization}, 
      author={Chuanyu Qin and Chenxu Yang and Qingyi Si and Naibin Gu and Dingyu Yao and Zheng Lin and Peng Fu and Nan Duan and Jiaqi Wang},
      year={2026},
      eprint={2604.20733},
      archivePrefix={arXiv},
      primaryClass={cs.LG},
      url={https://arxiv.org/abs/2604.20733}, 
}

@misc{gu2026coevolvingpolicydistillation,
      title={Co-Evolving Policy Distillation}, 
      author={Naibin Gu and Chenxu Yang and Qingyi Si and Chuanyu Qin and Dingyu Yao and Peng Fu and Zheng Lin and Weiping Wang and Nan Duan and Jiaqi Wang},
      year={2026},
      eprint={2604.27083},
      archivePrefix={arXiv},
      primaryClass={cs.LG},
      url={https://arxiv.org/abs/2604.27083}, 
}

@misc{hu2026synergisticintracrosslayerregularization,
      title={Synergistic Intra- and Cross-Layer Regularization Losses for MoE Expert Specialization}, 
      author={Rizhen Hu and Yuan Cao and Boao Kong and Mou Sun and Kun Yuan},
      year={2026},
      eprint={2602.14159},
      archivePrefix={arXiv},
      primaryClass={cs.LG},
      url={https://arxiv.org/abs/2602.14159}, 
}

@article{hinton2015distilling,
  title={Distilling the knowledge in a neural network},
  author={Hinton, Geoffrey and Vinyals, Oriol and Dean, Jeff},
  journal={arXiv preprint arXiv:1503.02531},
  year={2015}
}

@article{hu2026mecefo,
  title={MeCeFO: Enhancing LLM Training Robustness via Fault-Tolerant Optimization},
  author={Hu, Rizhen and He, Yutong and Yan, Ran and Sun, Mou and Yuan, Binhang and Yuan, Kun},
  journal={Advances in Neural Information Processing Systems},
  volume={38},
  pages={173324--173350},
  year={2026}
}

@article{xie2026step,
  title={Step-wise Rubric Rewards for LLM Reasoning},
  author={Xie, Weichu and Zhao, Haozhe and Liu, Wenpu and Zhu, Yongfu and Chen, Liang and Ye, Minghao and Chen, Zirong and Xu, Yuqi and Dong, Shuai and Wang, Ziyue and others},
  journal={arXiv preprint arXiv:2605.17291},
  year={2026}
}

@inproceedings{zhan2026mathsmith,
  title={Mathsmith: Towards extremely hard mathematical reasoning by forging synthetic problems with a reinforced policy},
  author={Zhan, Shaoxiong and Lai, Yanlin and Lu, Ziyu and Lin, Dahua and Yang, Ziqing and Tan, Fei},
  booktitle={Proceedings of the AAAI Conference on Artificial Intelligence},
  volume={40},
  number={41},
  pages={34602--34610},
  year={2026}
}

@article{zhan20263viewsense,
  title={3viewsense: Spatial and mental perspective reasoning from orthographic views in vision-language models},
  author={Zhan, Shaoxiong and Lai, Yanlin and Liu, Zheng and Lin, Hai and Li, Shen and Cai, Xiaodong and Lin, Zijian and Huang, Wen and Zheng, Hai-Tao},
  journal={arXiv preprint arXiv:2603.07751},
  year={2026}
}
\bibliographystyle{icml2026}

%%%%%%%%%%%%%%%%%%%%%%%%%%%%%%%%%%%%%%%%%%%%%%%%%%%%%%%%%%%%%%%%%%%%%%%%%%%%%%%
%%%%%%%%%%%%%%%%%%%%%%%%%%%%%%%%%%%%%%%%%%%%%%%%%%%%%%%%%%%%%%%%%%%%%%%%%%%%%%%
% APPENDIX
%%%%%%%%%%%%%%%%%%%%%%%%%%%%%%%%%%%%%%%%%%%%%%%%%%%%%%%%%%%%%%%%%%%%%%%%%%%%%%%
%%%%%%%%%%%%%%%%%%%%%%%%%%%%%%%%%%%%%%%%%%%%%%%%%%%%%%%%%%%%%%%%%%%%%%%%%%%%%%%
\newpage
\appendix
%\onecolumn

\begin{figure*}[tp]
    \begin{subfigure}{1\linewidth}
    \centering
    \includegraphics[width=1\linewidth]{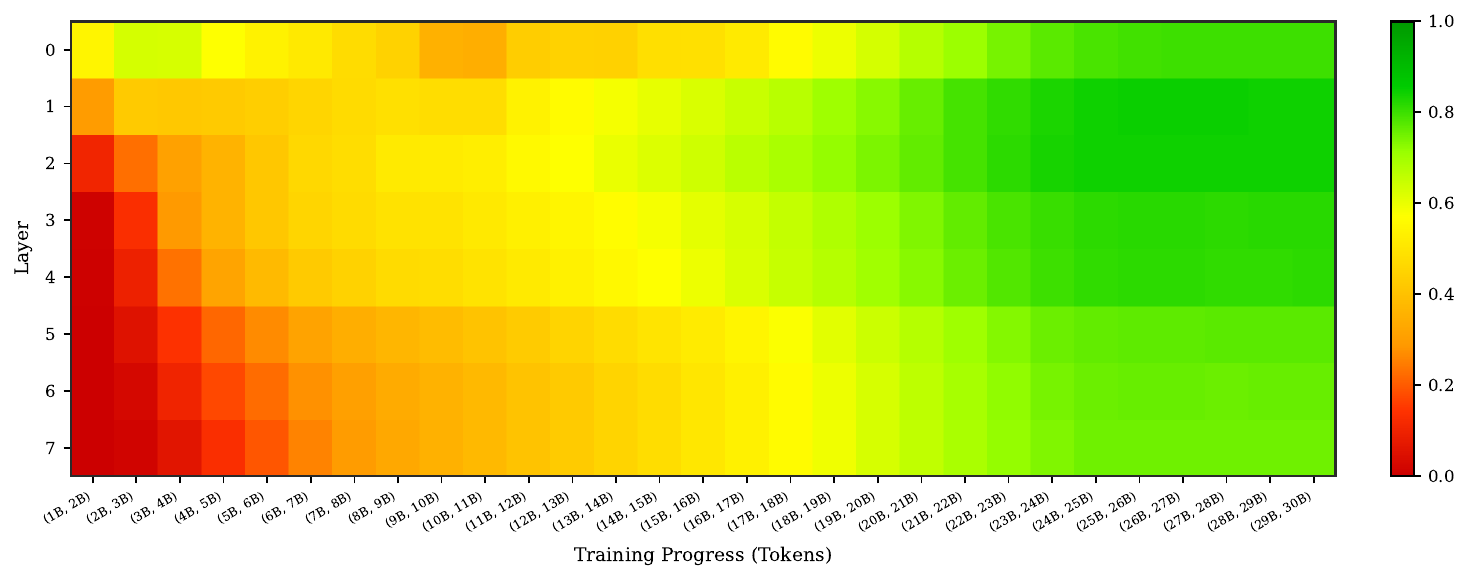}
    \caption{Exact Match Rate of Routing Decision}
    \label{fig:routing_match_heatmap_all}
    \end{subfigure}
    \begin{subfigure}{1\linewidth}
        \centering
        \includegraphics[width=0.9\linewidth]{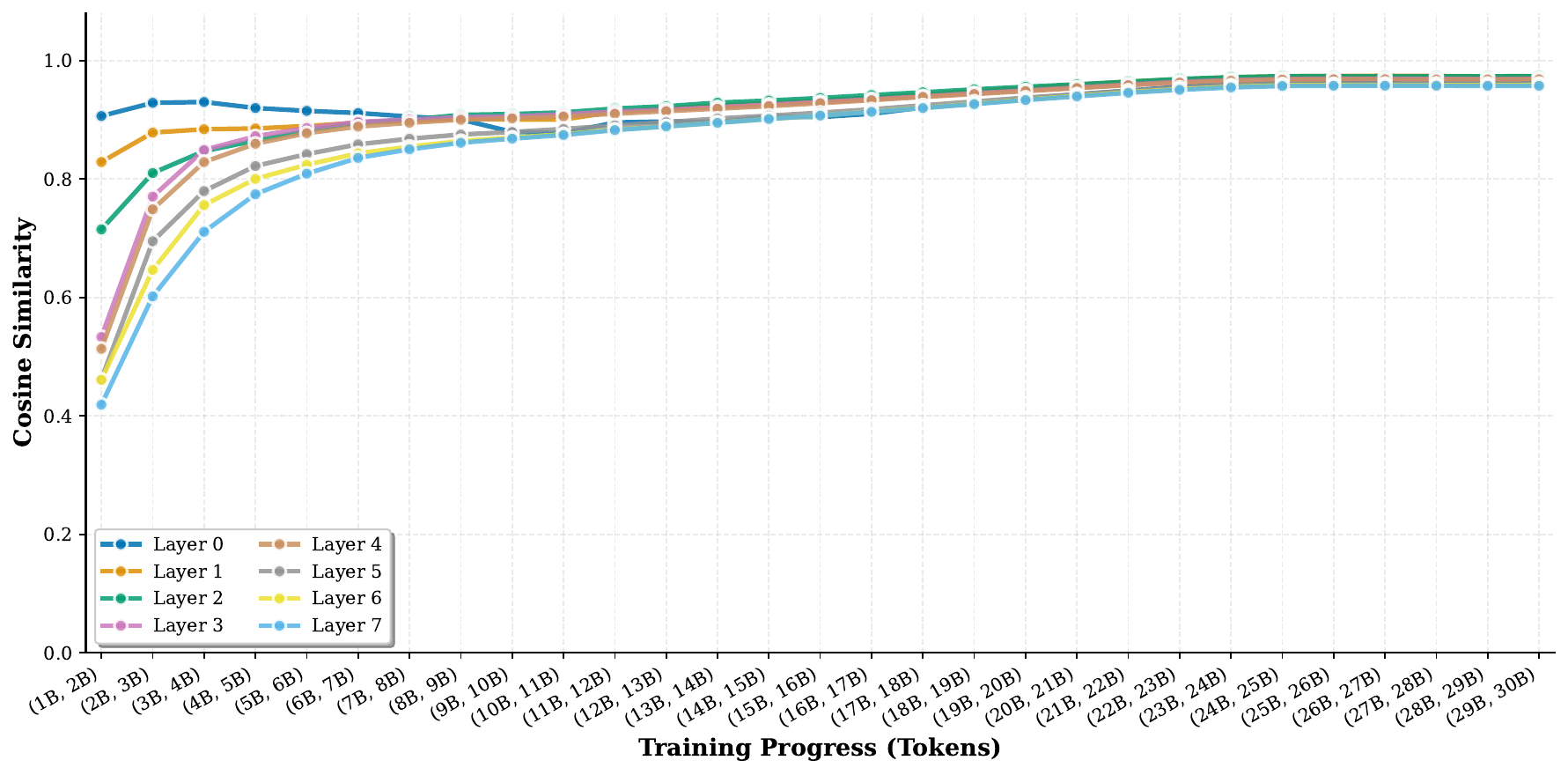}
        \caption{Cosine Similiarity of Routing Score}
        \label{fig:routing_similiarity_all}
    \end{subfigure}
    \caption{Routing fluctuations of a 550M model trained on 30B tokens. Given the high token-to-parameter ratio of 60, the model has largely converged. It is observed that the router exhibits intense volatility during the early stages of training; however, as training progresses, the routing patterns gradually converge and maintain stability.}
    \label{fig:routing_stability_analysis}
\end{figure*}

\section{Routing-Representation Interference}\label{ap: Structure-Performance Interference}

Suppose the MoE layer defined in \eqref{eq:moe_output}--\eqref{eq:routing_func} is parameterized by router parameters $\theta_r$ (within $\mathbf{s}(\mathbf{x})$) and expert parameters $\theta_{e_i}$ (within $\mathbf{f}_i(\mathbf{x})$). During training, these parameters are updated via gradient descent:
\begin{align}
    \label{simul-update}
    \theta_r^{t+1} = \theta_r^t - \alpha\nabla^t_{\theta_r} \mathcal{L}, \quad 
    \theta_e^{t+1} = \theta_e^t - \alpha\nabla^t_{\theta_e} \mathcal{L}
\end{align}
where $\mathcal{L}$ is the task loss and $e$ denotes all experts parameters. Due to the composite structure of the MoE output in \eqref{eq:moe_output}, the gradients decompose into two components by chain rule:
\begin{align}
\nabla^t_{\theta_r} \mathcal{L}=
\frac{\partial \mathcal{L}}{\partial \mathbf{y}}
\cdot \frac{\partial \mathbf{y}}{\partial \theta_r}\Big|_{\theta_r^t}, \quad 
\nabla^t_{\theta_e} \mathcal{L}=\frac{\partial \mathcal{L}}{\partial \mathbf{y}}
\cdot \frac{\partial \mathbf{y}}{\partial \theta_e}\Big|_{\theta_e^t}.
\end{align}
Consequently, at each training step, the MoE model simultaneously performs two updates: one optimizes the routing structure $\mathbf{r}(\mathbf{x})$ (via $\theta_r$) to determine which experts are activated and their blending weights, while the other enhances the experts' representation performance (via $\theta_e$).

However, this simultaneous optimization of structure and performance creates inherent interference. At training step $t$, the \textbf{observed} expert gradient is
% \vspace{-2mm}
\begin{equation}
    \nabla_{\theta_e}^t \mathcal{L} = \frac{\partial \mathcal{L}}{\partial \mathbf{y}}\cdot\sum_{i=1}^E \mathbf{r}(\mathbf{x};\theta_r^t)_i \cdot \nabla_{\theta_{e_i}} \mathbf{f}_i(\mathbf{x};\theta_{e_i}^t), 
\end{equation}
% \begin{equation}
%     \nabla_{\theta_e}^t \mathcal{L} = \frac{\partial \mathcal{L}}{\partial \mathbf{y}}\cdot\sum_{i=1}^E \mathbf{r}(\mathbf{x};\theta_\tau^t)_i \cdot \nabla_{\theta_{e_i}} \mathbf{f}_i(\mathbf{x};\theta_{e_i}^t) 
% \end{equation}
% \vspace{-1mm}
where $\mathbf{r}(\mathbf{x};\theta_r^t) \in \mathbb{R}^E$ is the router output vector under router parameters $\theta_r^t$, and $\mathbf{f}_i(\mathbf{x}; \theta_{e_i}^t)$ denotes the output of the $i$-th expert parameterized by $\theta_{e_i}^t$. In contrast, if training occurred with optimal, stable router parameters $\theta_r^*$, the \textbf{ideal} expert gradient would be:
\begin{equation}
\label{optimal-grad}
    \nabla_{\theta_e}^* \mathcal{L} = \frac{\partial \mathcal{L}}{\partial \mathbf{y}} \cdot \sum_{i=1}^E \mathbf{r}(\mathbf{x};\theta_r^*)_i \cdot \nabla_{\theta_{e_i}} \mathbf{f}_i(\mathbf{x};\theta_{e_i}^t). %\quad \text{(Ideal Stable Gradient)}
\end{equation}
Since $\theta_r^t$ continuously evolves during training ($\theta_r^t \neq \theta_r^*$), the observed expert parameter updates constantly deviate from the ideal stable trajectory. This deviation accumulates over time, degrading optimization efficiency. We quantify this accumulated optimization error as the sum of Euclidean distances between the observed and ideal expert gradients:
\begin{equation}
\label{249ansd}
    \mathbf{E}_{\text{opt}} = \sum_{t} \| \nabla_{\theta_e}^t \mathcal{L} - \nabla_{\theta_e}^* \mathcal{L} \|,
    \vspace{-2mm}
\end{equation}
which serves as a quantitative measure of interference severity.

To eliminate this error, we propose a decoupled optimization strategy. Rather than simultaneously updating both $\theta_e$ and $\theta_r$ as in \eqref{simul-update}, we adopt a two-stage approach. First, we obtain near-optimal router parameters $\theta_r^*$ by distilling from a fully trained MoE model (e.g., DeepSeek-V2-Lite, Qwen3-30B-A3B, Qwen3-235B-A22B). Second, we fix $\theta_r^*$ and optimize only the expert parameters using the gradient in \eqref{optimal-grad}. This ensures that expert optimization proceeds along the ideal trajectory throughout training.

\section{Detailed Desctiption of Routing Fluctuation Heatmap}\label{Detailed Desctiption of Routing Fluctuation Heatmap}

% 我们将更全面的训练过程中routing fluctuation heatmap展示于图1。这个heatmap的绘制方式如下，我们保存训练过程中的ckpt，之后对相邻的ckpt，例如经过1B token训练和经过2B token训练的ckpt，输入完全相同的输入，观察其在每个token上分配的专家是否完全相同，如果完全相同，我们认为在这个token上实现了稳定，最后所有MoE层中，我们统计整个输入的所有token中，保持稳定的token的比率，并将其结果放置在对应的((ckpt1,ckpt2), Layer)位置。

\autoref{fig:routing_stability_analysis} provides a comprehensive visualization of the routing dynamics throughout the training process. To quantify this evolution, we periodically save model checkpoints and evaluate the routing consistency between adjacent pairs (e.g., at 1B and 2B tokens) by feeding them an identical input batch. Specifically, \autoref{fig:routing_match_heatmap_all} illustrates the proportion of tokens assigned to the same experts across checkpoints, while Section \ref{fig:routing_similiarity_all} measures the cosine similarity of routing scores to reflect the magnitude of shift in the router's soft preferences.

As shown in the figures, both expert assignment and routing scores exhibit significant volatility during the early stages of training. However, as training progresses, these two metrics converge toward a steady state, demonstrating increasing routing stability. This observation empirically validates our hypothesis regarding the \textit{Inefficiency of Joint Router-Expert Optimization} discussed in \autoref{Inefficiency of Joint Router-Expert Optimization}, confirming that the router eventually reaches an asymptotic equilibrium despite the initial turbulence.

\section{Limitations in Existing Approaches}\label{Limitations in Existing Approaches}

Although several methods have attempted to decouple the router from the MoE framework, critical challenges remain unresolved. We summarize the limitations of these approaches as follows:

\textbf{Limitations in Static Structure Assignment.} Some prior works have attempted to construct stable structures by fixing token-to-expert assignment tables~\cite{roller2021hash}. However, this approach suffers from several critical drawbacks. First, a fixed token-to-expert mapping contradicts the core design principles of the Transformer. Due to attention mechanisms~\cite{vaswani2017attention} and positional encodings~\cite{su2024roformer}, the same token should activate different experts depending on its specific context. A static lookup table eliminates this crucial contextual sensitivity, effectively collapsing the routing space from exponentially many context-dependent assignments to a simple token-identity lookup. This rigidity prevents the model from capturing the nuanced semantic roles a token may play in different sequences. Second, lookup tables are inherently incapable of specifying the continuous blending weights for expert aggregation. While a table can determine which experts are activated, it cannot determine the specific values of $\mathbf{r}(\mathbf{x})$. These weights represent the relative importance of each expert for a given input, which are continuous values that must be computed dynamically by a router. Since these coefficients cannot be reasonably constructed through manual annotation or fixed assignments, a lookup table remains an incomplete solution for the decoupled routing paradigm.

\textbf{Limitations in Dynamic Structure Learning.} Compared to static assignment, which can only evaluate structure quality from a load-balancing perspective, learning structure through training is more principled. StableMoE~\cite{dai2022stablemoe} exemplifies this approach by constructing a linear network $\mathbf{D}$ that attempts to capture routing structure by distilling the MoE router's output. Specifically, StableMoE divides pretraining into two phases. The first phase employs a triple-loss function:
\begin{equation*}
    \mathcal{L} = \mathcal{L}_{\text{CE}} + \mathcal{L}_{\text{Bal}} + \mathcal{L}_{\text{Distill}}
\end{equation*}
where $\mathcal{L}_{\text{CE}}$ is the cross-entropy loss, $\mathcal{L}_{\text{Bal}}$ is the load balancing loss, and $\mathcal{L}_{\text{Distill}}$ represents the Kullback-Leibler divergence between $\mathbf{D}$'s predictions and the router's output:
\begin{equation*}
    \mathcal{L}_{\text{Distill}} = \text{KL} \left( \mathbf{D}(\mathbf{H}_{\text{in}}) \parallel \mathbf{s}(\mathbf{H}) \right)
\end{equation*}

During the first phase, the model simultaneously optimizes for task performance, load balance, and distillation. In the second phase, $\mathbf{D}$ is frozen and used to select experts. However, this method faces several critical issues.  First, the synchronous optimization of the triple-loss function intensifies the coupling between task performance and structural stability. Forcing the model to simultaneously learn representations, load balancing, and routing distillation creates a volatile optimization landscape that hinders efficient convergence. Second, because model training and distillation occur synchronously, the $\mathbf{D}$ network attempts to learn a continuously evolving target structure, resulting in poor distillation quality. Third, if training collapses or distillation fails, the entire procedure must be restarted from scratch, imposing significant computational costs.

\section{Benefits of Preemptive Structures}\label{ap:Benefits of Preemptive Structures}

Many existing MoE optimization algorithms are fundamentally constrained by their reliance on real-time router outputs during training. These methods must operate after dynamic routing decisions $\mathbf{r}(\mathbf{x})$ are made, limiting optimization to highly restricted conditions. For example, NetMoE~\cite{liu2025netmoe} reduces Expert Parallel~\cite{shazeer2017outrageously} communication costs by computing optimal token placements based on token-to-expert affinity during the dispatch stage. Similarly, FlexMoE~\cite{nie2023flexmoe} addresses dynamic load imbalance by adapting hardware resources to tokens' instantaneous routing demands, considering current expert loads and expert-to-GPU mappings. Furthermore, MoE++~\cite{jin2024moe++} proposes Zero-Experts, allowing tokens to activate fewer than TopK experts to accelerate inference based on immediate router outputs. Linear-Programming-Based Load Balancer~\cite {caohuanqi2025lplb} is another method that addresses dynamic load imbalance, yet it still depends on real-time router-derived token indices and per-batch workload statistics to optimize load distribution in expert-parallel groups, remaining constrained by runtime requirements.

%However, the necessity of acquiring router outputs in real time during training imposes significant constraints, often restricting these algorithms to localized optimizations. Consequently, establishing a preemptive routing structure would unlock substantially greater optimization potential for these methods, potentially allowing their objectives to be decoupled entirely from the main training loop.

However, the necessity of acquiring router outputs in real-time imposes a "computational bottleneck" on the optimization process itself. This dependency locks the MoE framework into a state of reactive adaptation rather than proactive orchestration. Consequently, establishing a preemptive routing structure—one that is determined before the training loop begins—would unlock a substantially broader optimization space.

% \section{Experiment Setup}\label{ap:Experiment Setup}
% \textbf{Infrastructure} Our experiments were conducted on a cluster comprising eight $\text{NVIDIA A}100$ GPUs and eight $\text{NVIDIA H}100$ GPUs, interconnected via $\text{NVLink}$. The software environment utilized $\text{PyTorch 2.8.0}$ with $\text{CUDA 12.9}$ and the $\text{Megatron-LM}$ framework (commit hash $\text{e7c55de9}$).

% \textbf{Grouter Distillation Setup} We distill our \GR\ from the Qwen3-30B-A3B model~\cite{qwen3technicalreport}. The distillation process is performed on the C4 dataset\cite{raffel2020exploring} for $2.6$ Billion tokens. We construct the \GR\ using a three-layer Transformer Encoder, resulting in a total parameter count of $60$M. Notably, $50$M of these parameters reside in the computationally inexpensive Embedding layer, which makes the overall \GR\ architecture highly lightweight. The consistently decreasing loss curve, as shown in \autoref{apd:distill_loss}, demonstrates the successful extraction of the structural prior. Unless otherwise specified, all subsequent experiments are based on this single distilled \GR\ instance.

% \textbf{Model Configurations} We define four distinct MoE model variants for our experiments: Tiny-Qwen3, Mini-Qwen3, $\text{Mini-DS-V2-Lite}$, and Mini-GPT-OSS. These models are structurally based on the architectures of Qwen3-30B-A3B~\cite{qwen3technicalreport}, DeepSeek-V2-Lite~\cite{deepseekv2}, and GPT-OSS-20B~\cite{openai2025gptoss120bgptoss20bmodel}, respectively. Their detailed configurations are presented in the \autoref{ap:configuration}.

\section{Related Works}\label{ap:extended_related_work}

We discuss additional related work that extends beyond the scope of the main text.

\textbf{Large Language Models.} Large Language Models have driven rapid progress across a broad range of tasks, from textual reasoning~\cite{xie2026step,zhan2026mathsmith} to multimodal understanding and generation~\cite{dong2025interleaved,zhan20263viewsense}. Modern LLM development typically follows a two-stage paradigm: large-scale pre-training to acquire general knowledge, followed by post-training~\cite{liu2026leveraging} alignment to elicit desired behaviors. Our work addresses the pre-training stage, where MoE architectures have become the dominant approach for scaling model capacity efficiently. While the above advances improve what models learn, \GR\ improves how efficiently MoE models train by providing a structural prior that decouples routing from representation learning.

\textbf{Knowledge Distillation.} Knowledge distillation~\cite{hinton2015distilling} transfers learned knowledge across models of different capacities and has been widely adopted throughout the LLM pipeline, from model compression to policy transfer in RL settings~\cite{yang2026selfdistilledrlvr, gu2026coevolvingpolicydistillation, qin2026nearfuturepolicyoptimization}. \GR\ employs distillation for a distinct purpose: rather than compressing a model's task knowledge, it extracts the routing structure from a converged MoE model into a lightweight standalone network, which then serves as a frozen structural prior for target model training.

\section{The Implementation Details of Baselines}\label{ap:imple of baseline}
% 我们基于Megatron-LM~(Git Bash e7c55de9)版本构建了我们的Baseline。对于包含额外超参的Baseline，我们选取超参的原则是如果有官方代码实现则和官方代码统一，如果没有官方则按照论文中提到的实验使用的超参数一致。对于ReMoE，我们采用了系数为0.01的负载均衡损失和Pre-softmax方法来确定专家权重。对于ToPP，我们采用了0.4的threshold，0.01的负载均衡辅助损失系数和0.0001为系数的dynamic loss。对于MoE++，我们在正常的专家之外，额外添加了6个constant expert，1个zero expert, 1个copy expert，使用了0.01系数的辅助损失函数和1.1专家容量，开启gating residuals。对于Z-Loss，我们参考\cite{zoph2022st}的设置，采用了0.01为系数的负载均衡损失，0.001系数的Z loss. 对于Aux Loss Free,我们采用了sigmoid作为router的归一化函数，0.001的更新步长，不采用辅助损失函数。对于HashLayer,我们采用\cite{roller2021hash}中表现最好的Balanced HashLayer,具体来说，我们首先统计数据中每个token的词频，之后按照负载均衡标准构建Token-专家映射表，在原论文中HashLayer没有router网络，无法给不同专家分配权重，为了确保对比的公平性，我们为HashLayer适配的router网络，在前向传播中，由router网络生成logits，token-expert映射表决定每个token选择的专家后，进行softmax归一化，之后将权重分配给选择的专家继续前传。对于StableMoE,我们和\cite{dai2022stablemoe}保持一致，设置第一阶段的步数为总训练步数的10%。ReMoE和ToPP采取了不固定激活专家数量的设置，我们发现在当前设置下他们的激活专家数量不会低于公平对比的标准，我们可视化他们的激活专家数量于图x。

 We compare \GR\ against a spectrum of prior works focused on router improvements. All baselines were implemented under the same operating environment and within the same version of the $\text{Megatron-LM}$ framework.In addition to \textbf{StableMoE} and \textbf{HashLayer}, representative methods that address the decoupling problem that we also target, we selected the following representative $\text{Router}$ enhancement methods as our baselines: (a) \textbf{ReMoE} replaces the $\text{Top-K}$ operation with a $\text{ReLU}$ gate to mitigate the issue of truncated router gradients. (b) \textbf{ToPP} and 
\textbf{MoE++} introduce mechanisms for dynamically selecting the number of active experts. (c) \textbf{Z-Loss} enhances router stability via an additional regularization loss, leading to improved model performance. (d) \textbf{Aux-Loss-Free} employs expert bias to remove the auxiliary balancing loss from the optimization objective, thereby avoiding performance degradation caused by loss interference.

All baselines were built upon the $\text{Megatron-LM}$ framework (commit hash $\text{e7c55de9}$). For baselines involving specific hyperparameters, we adhere to the following principles: if official code is available, we adopt the original configuration; otherwise, we strictly follow the parameters specified in the respective paper's experimental section.

The specific configurations for each representative baseline are as follows:
\begin{itemize}
    \item \textbf{ReMoE} We use a load balancing loss coefficient of $0.01$ and employ the Pre-Softmax method for determining expert weights, as is standard for this architecture.
    \item \textbf{ToPP} The configuration utilizes a $\text{threshold}$ of $0.4$, a load balancing auxiliary loss coefficient of $0.01$, and a dynamic loss coefficient of $0.0001$.
    \item \textbf{MoE++} In addition to the normal experts, we included six $\text{constant}$ experts, one $\text{zero}$ expert, and one $\text{copy}$ expert. We set the auxiliary loss coefficient to $0.01$, the expert capacity factor to $1.1$, and enabled gating residuals.
    \item \textbf{Z-Loss} Following the setup in ~\cite{zoph2022st}, we set the load balancing loss coefficient to $0.01$ and the $\text{Z-loss}$ coefficient to $0.001$.
    \item \textbf{Aux-Loss-Free} We adopt the $\text{sigmoid}$ function as the router's normalization function, use a learning rate of $0.001$, and disable the auxiliary loss.
    \item \textbf{HashLayer} We implement the Balanced $\text{HashLayer}$, which showed the best performance in \cite{roller2021hash}. Specifically, we first calculate the token frequency distribution in the dataset and then construct the $\text{Token-Expert}$ mapping table based on load-balancing criteria. To ensure a fair comparison, since $\text{HashLayer}$ inherently lacks a $\text{router}$ network for weight assignment, we adapted a $\text{router}$ network: during the forward pass, the $\text{router}$ network generates logits, the $\text{Token-Expert}$ mapping determines the selected expert for each token, and the weights are then calculated via $\text{softmax}$ and assigned to the selected experts before continuing the forward propagation.
    \item \textbf{StableMoE} We maintained consistency with \cite{dai2022stablemoe}, setting the first pre-training stage to $10\%$ of the total training steps.
\end{itemize}

\begin{figure}[htbp]
    \centering
    \includegraphics[width=0.9\linewidth]{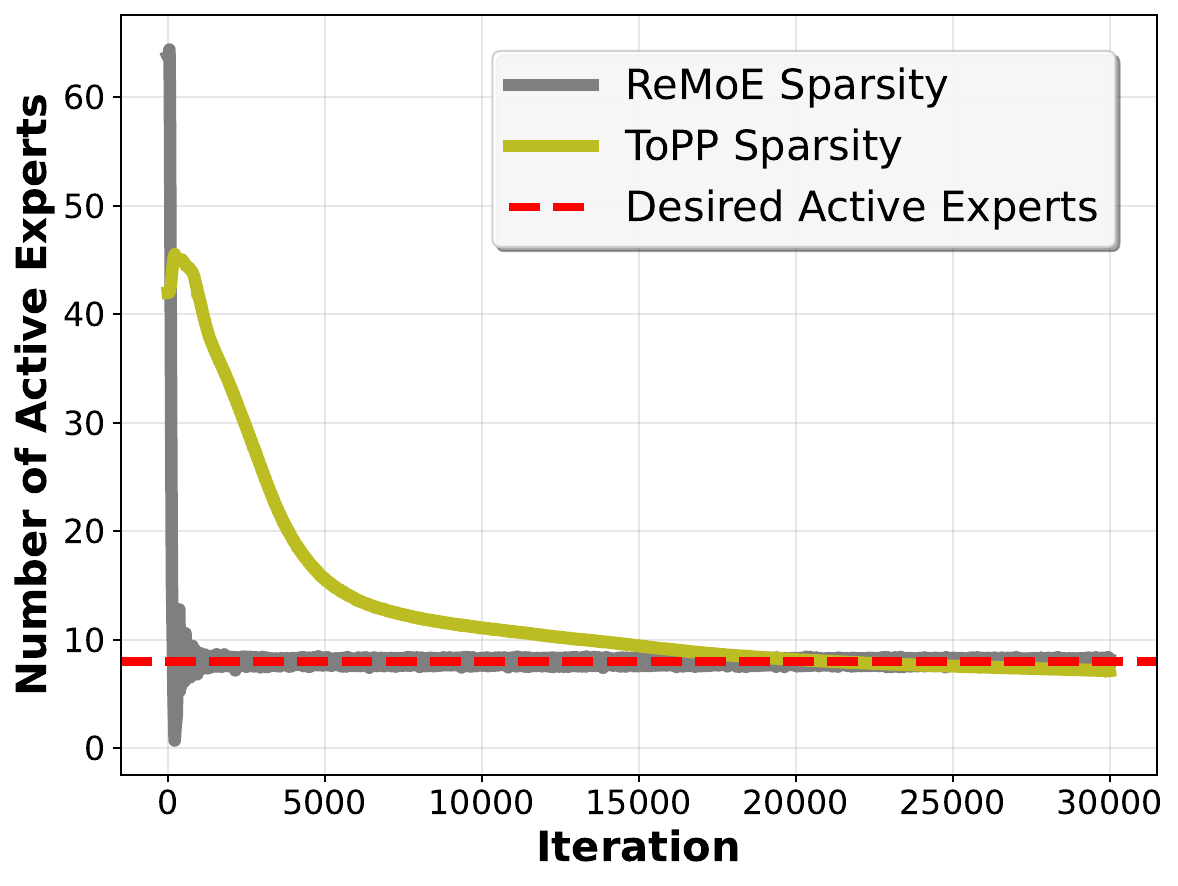}
    \caption{\small Number of Active Experts During Training for Dynamic Activation Baselines. The number of active experts remains consistently higher than the expected value throughout the training process, confirming a fair comparison.}
    \label{fig:sparsity_pre}
    \vspace{-2mm}
\end{figure}

Note that $\text{ReMoE}$ and $\text{ToPP}$ utilize a setup with a non-fixed number of active experts. We found that under our experimental settings, their number of active experts does not fall below the standard required for a fair comparison. The active expert counts for these methods are visualized in \autoref{fig:sparsity_pre}.

\begin{figure*}[tp]
    \centering 
    \begin{subfigure}{0.245\textwidth}
        \centering
        \includegraphics[width=\linewidth]{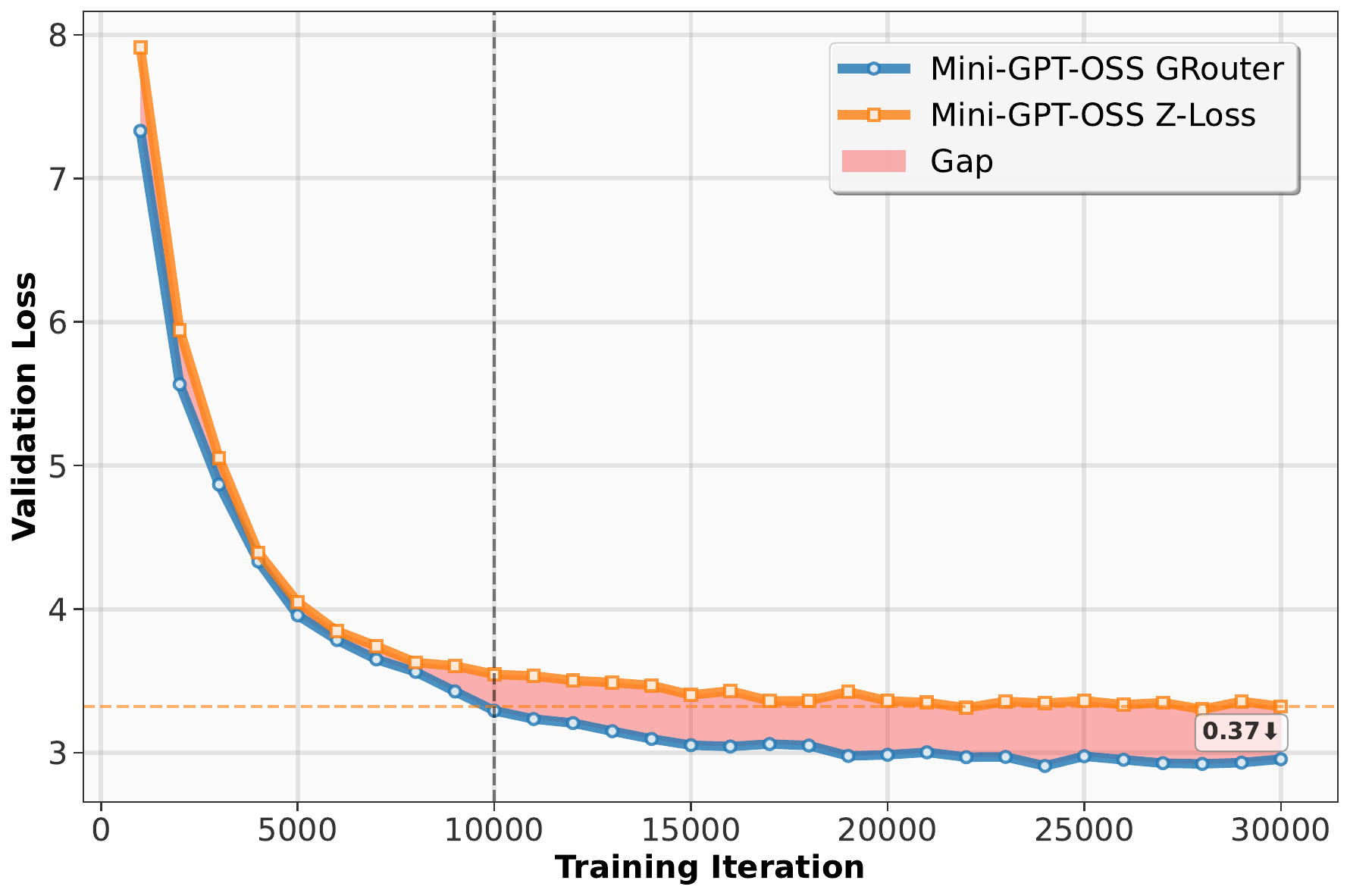}
        \caption{Mini-GPT-OSS} 
    \end{subfigure}
    \hfill  
    \begin{subfigure}{0.245\textwidth}
        \centering
        \includegraphics[width=\linewidth]{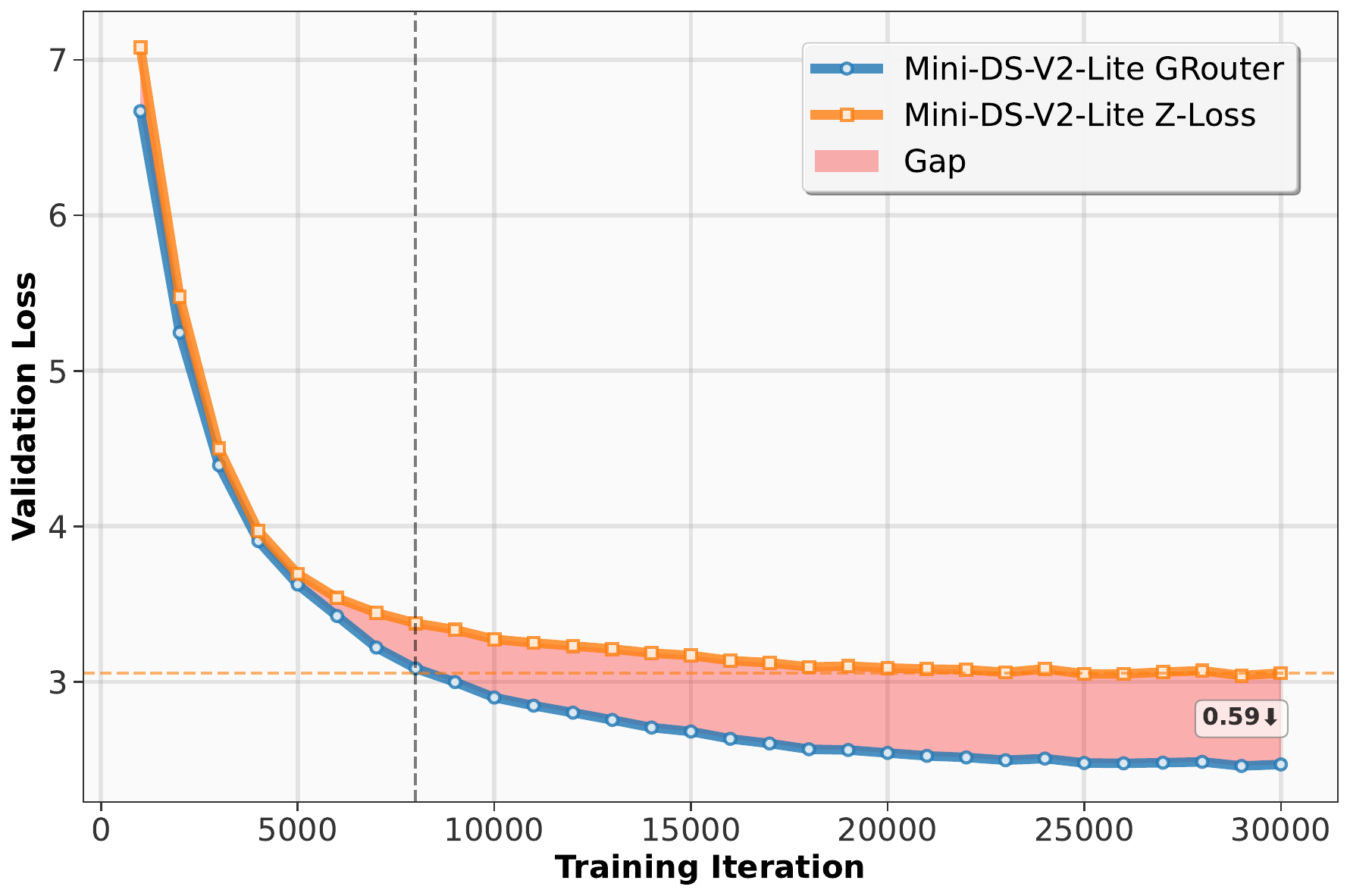}
        \caption{Mini-DS-V2-Lite}
    \end{subfigure}
    \hfill  
    \begin{subfigure}{0.245\textwidth}
        \centering
        \includegraphics[width=\linewidth]{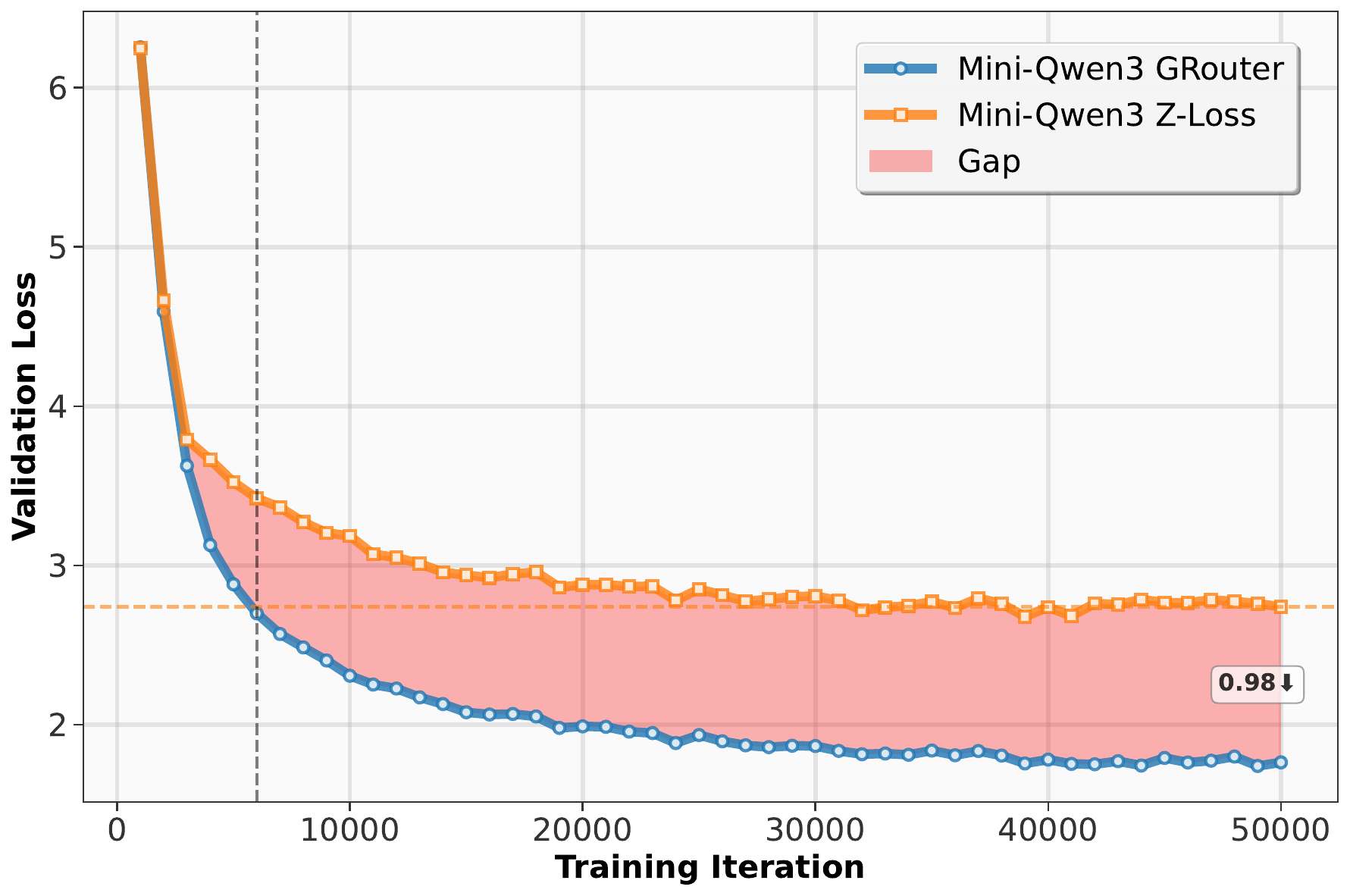}
        \caption{Mini-Qwen3}  
    \end{subfigure}
    \hfill  
    \begin{subfigure}{0.245\textwidth}
        \centering
        \includegraphics[width=\linewidth]{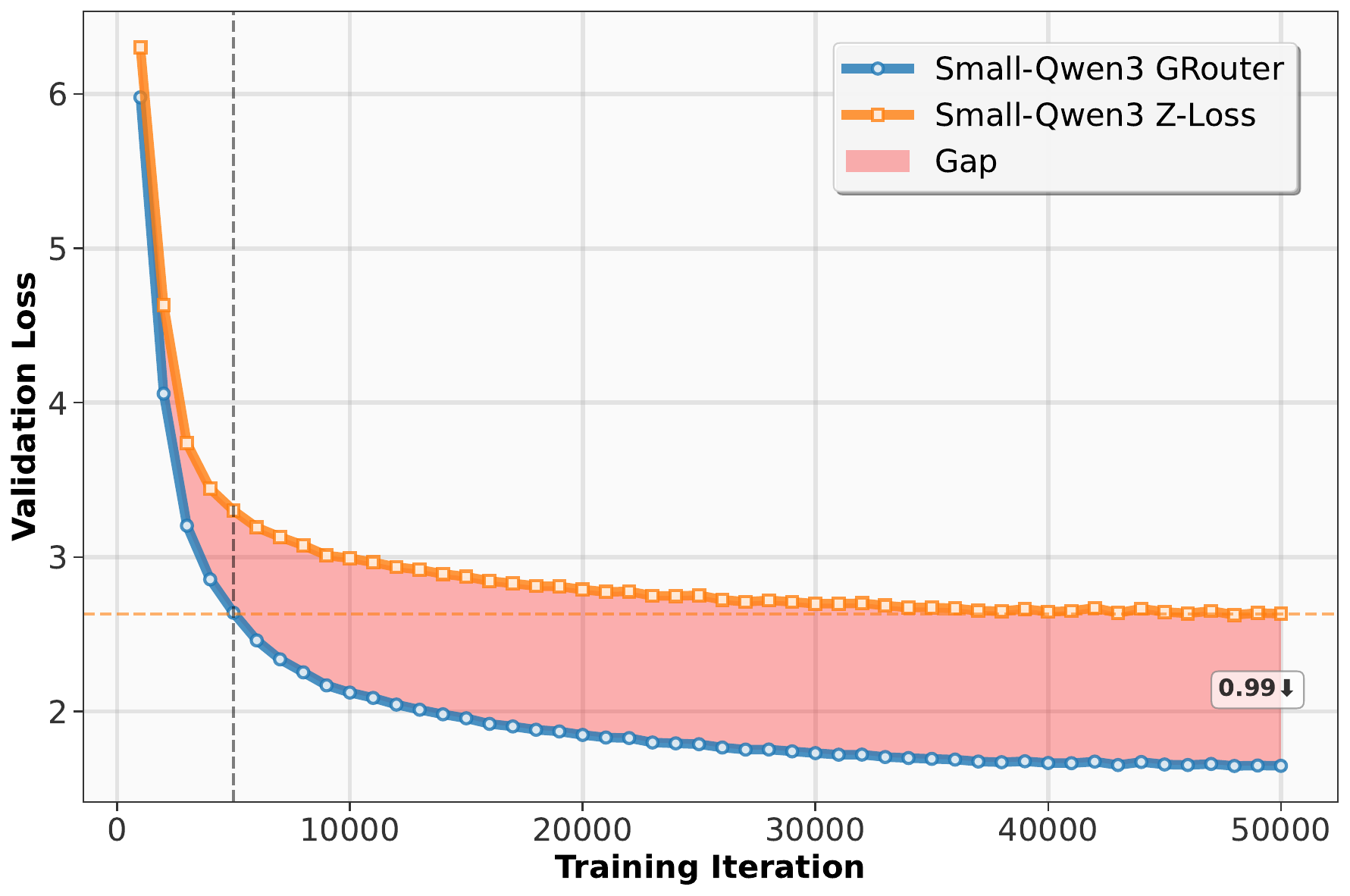}
        \caption{Small-Qwen3}  
    \end{subfigure}
    \caption{\small Training curves of Grouter and Z-loss for models with varying sizes, architectures, and expert counts. The results demonstrate that a single Grouter can be migrated to different configurations via expert folding and tuning, while retaining superior convergence speed and precision.}
    \label{fig:diff_config}
\end{figure*}

\section{Ablation}\label{ap:Ablation}

\begin{table*}[htbp]
    \centering

    \caption{Ablation Study on \GR\ Architectures\label{abl:str}}
    \begin{tabular}{@{} *{10}{c} @{}}
        \toprule
        \multicolumn{4}{c}{Multi Head Attn} & 
        \multicolumn{4}{c}{Multi Latent Attn} & 
        \multicolumn{2}{c@{}}{Residual Linear} \\ 
        \cmidrule(lr){1-4} \cmidrule(lr){5-8} \cmidrule(l){9-10} 
        \multicolumn{2}{c}{Decoder} & 
        \multicolumn{2}{c}{Encoder} & 
        \multicolumn{2}{c}{Decoder} & 
        \multicolumn{2}{c}{Encoder} & 
        \multicolumn{2}{c@{}}{Linear} \\ 
        \cmidrule(lr){1-2} \cmidrule(lr){3-4} \cmidrule(lr){5-6} \cmidrule(lr){7-8} \cmidrule(l){9-10} 
        Use\_Pos & No\_Pos & Use\_Pos & No\_Pos & 
        Use\_Pos & No\_Pos & Use\_Pos & No\_Pos & 
        Use\_Pos & No\_Pos \\
        \midrule
        3.342 & 3.365 & \textbf{3.318} & 3.367 & 3.335 & 3.387 & 3.356 & 3.360 & 3.375 & 3.359 \\
        \bottomrule
    \end{tabular} 

\end{table*}

% 这一部分我们对\GR\的一些关键选择进行消融，我们首先针对Grouter的结构进行消融实验，我们首先探究哪种结构可以最好的帮助Grouter提取模型结构。根据之前尝试将Router解耦的工作\cite{dai2022stablemoe, cai2024textit},我们将选择范围限制在三种可能的模型结构上，分别是基于Multi Head Attention的Transformer，基于Multi Latent Attention de Transformer，基于有残差连接结构的线性层。他们均会接入Embedding和输出线性层之间，负责提取Source Model的特征结构。除去模型结构本身，是否需要传递位置信息也是一个需要考虑的因素，对于路由决策来说，位置信息并不是一个必需的信息，甚至可能会造成干扰，但是传入位置信息也可以让Grouter更好的理解Source Model的路由决策，因此我们对于这两方面做了消融实验，具体来说，我们对每种类型的Grouter进行了2.5B token的蒸馏，之后将其应用于Mini-DS-V2-Lite训练10B数据，结果见\autoref{abl:str}。可以看到，使用Encoder Multi Head Attention的transformer模型具有显著的优势。

In this section, we conduct ablation studies on several key design choices for the \GR. We first focus on the \GR's internal structure to determine which architecture is most effective in extracting the model's structural prior. Following prior work on decoupled routers \cite{dai2022stablemoe, cai2024textit}, we restrict our scope to three plausible structures: (1) a Multi-Head Attention-based Transformer, (2) a Multi-Latent Attention-based Transformer, and (3) a Linear Layer with Residual Connections. Each of these structures is positioned between the $\text{Embedding}$ and the output $\text{Linear}$ layers, where it is tasked with capturing the feature structure required for routing in the Source Model.

Beyond the model architecture itself, whether to incorporate positional information is another critical factor. While positional information is not strictly necessary for routing decisions and could potentially introduce noise, its inclusion might enable the \GR\ to better comprehend the Source Model's routing decisions. Consequently, we perform ablations on both the architecture type and the inclusion of positional information.

%Beyond the model architecture itself, whether to incorporate positional information is another critical factor. While positional information is not strictly necessary for routing decisions and could potentially introduce noise, its inclusion might enable the \GR\ to better comprehend the source model's routing decisions. Consequently, we perform ablations on both the architecture type and the inclusion of positional information. Specifically, we distill each \GR\ variant for $2.5$ Billion tokens and subsequently apply it to the $\text{Mini-DS-V2-Lite}$ model for $10$ Billion tokens of pre-training. The results validation loss, presented in \autoref{abl:str}, demonstrate that the Multi-Head Attention-based Transformer Encoder exhibits a significant advantage.%我们认为这是因为Encoder Transformer具有注意力机制，更易理解Source Model的决策，且具备全局感受野，对于相同的输入有更多的信息来决定如何分配。

Specifically, we distill each \GR\ variant for $2.5$ Billion tokens and subsequently apply it to the $\text{Mini-DS-V2-Lite}$ model for $10$ Billion tokens of pre-training. The validation loss results, presented in \autoref{abl:str}, demonstrate that the Multi-Head Attention-based Transformer Encoder exhibits a significant advantage. We attribute this superiority to the $\text{Encoder}$ $\text{Transformer}$'s inherent attention mechanism, which is better suited to interpret the Source Model's global routing decisions. And its built-in global receptive field allows the \GR\ to leverage more context for token assignment.

% 我们探索不同的专家合并方法会对Grouter迁徙到不同配置的模型后有什么影响，我们对比了以下方案，分别是随机折叠，基于负载均衡的折叠，即将负载高的专家和负载低的专家折叠为一个专家，以此尽可能使的在新配置下的Grouter负载均衡，以及我们的基于亲和度的Grouter。我们分别使用三种方法得到三个Grouter，之后在Mini-GPT-OSS模型上进行5B数据的预训练，结果可见\autoref{abl:fold_bar}.通过对比最终验证集损失可以发现，基于专家亲和度的专家折叠效果最好。这是因为较高的专家亲和度意味着更多的共同职能，因此将这两个专家折叠时，损失的结构信息最少。

\begin{figure}[htbp]
    \centering
    \includegraphics[width=0.8\linewidth]{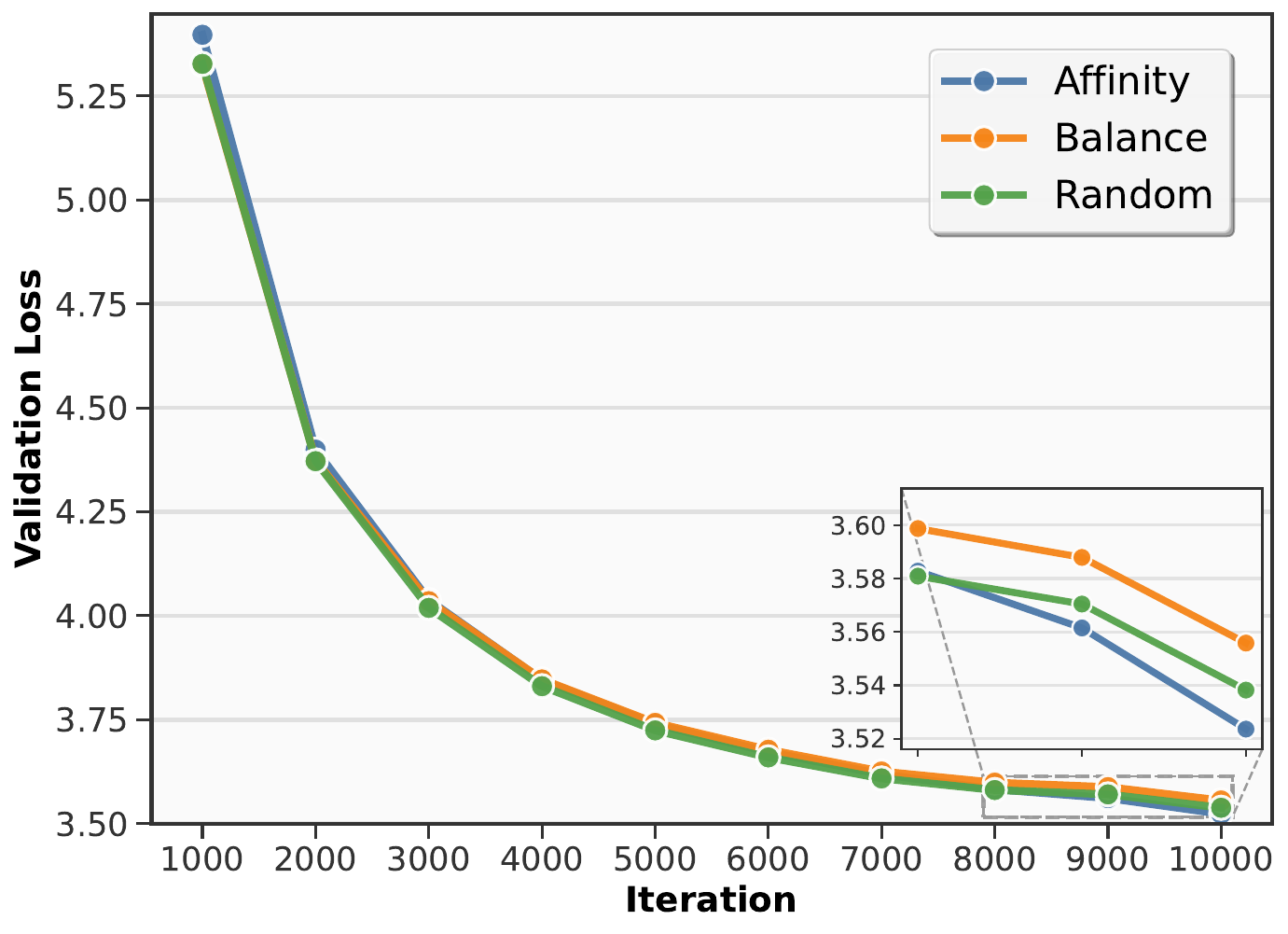}
    \caption{\small Ablation Study on Expert Folding Strategies.
    }
    \vspace{-3mm}
    \label{abl:fold_bar}
\end{figure}

We explore how different expert folding methods impact the performance of \GR\ when transferred to models with distinct configurations. We contrast three distinct strategies: Random Folding, where experts are merged arbitrarily; Load-Balance Folding, which merges high-load experts with low-load experts to intrinsically balance the \GR's load under the new configuration; and our proposed Affinity-Based Folding, where experts are merged based on their affinity.

We prepared three $\text{Grouter}$ variants, one for each folding method. Subsequently, these variants were applied to the $\text{Mini-GPT-OSS}$ model and pre-trained on $5$ Billion tokens. The results, measured by the final validation set loss, are presented in \autoref{abl:fold_bar}. The comparison clearly shows that the affinity-based expert folding yields the best performance. This is because high expert affinity indicates a greater shared functionality; therefore, merging these experts minimizes the loss of crucial structural information, leading to better model fidelity and overall performance.

\section{Maximal Violation}\label{Maximal Violation}

$\text{MaxVio}_{\text{Global}}$ serves as a robust and widely used indicator of the global load balance and is formally defined as:
\vspace{-2mm}
\begin{equation}
    \text{MaxVio}_{\text{Global}} = \frac{\max_i L_{e_i} - \overline{L}}{\overline{L}}
\end{equation}
\vspace{-1mm}
where $L_{e_i}$ is the total number of tokens afforded to expert $e_i$ across the entire training process, and $\overline{L}$ represents the uniform average load distributed among all experts.

% Maximal Violation(MaxVio)是一个被广泛使用的度量负载均衡的指标。

\section{Downstream Experiments}\label{apdix:downstream}
% 由于我们训练得到的Mini-Qwen3是未经指令跟随微调的Base模型版本，因此我们采取PPL形式来进行下游任务测试。具体来说，我们会将任务的不同选项构建为上下文加选项的prompt，这些prompt将会通过模型的前传得到模型针对选项的PPL,之后我们会选择模型PPL最少的选项作为模型的选择，与正确答案进行对比。下面是各个任务的Prompt模板。BoolQ:"Context is {passage}\nQuestion is {question}\nDoes the context answer the question with 'Yes'? Answer with 'Yes' or 'No':"。RTE:"Premise is {premise}\nHypothesis is {hypothesis}\nDoes the premise entail the hypothesis? Answer with 'Yes' or 'No':"。HellaSwag，我们将每个选项都拆分为一个判断正负的任务，这可以帮助模型更好的决定这一选项的置信度，我们的两套提示词为"Context is: {context}\nEnding is: {ending}\nIs this ending a reasonable continuation of the context? Answer with 'Yes' or 'No':" +" Yes"/"No"。PIQA，我们同样将每个选项均分为二分类任务，“Goal is: {goal}\nSolution is: {sol}\nIs this solution effective for the goal? Answer with 'Yes' or 'No':” + “Yes"/”No"。RACR,"Article is {article}\nQuestion is {question}\nIs the option '{option}' the correct answer to the question? Answer with 'Yes' or 'No':" + "Yes"/"No"。Mnli,"Premise: {premise}\nHypothesis: {hypothesis}\nWhat is the relationship between the premise and hypothesis? Choose from 'entailment', 'neutral', 'contradiction':"

 We selected the standard evaluation datasets: $\text{BoolQ}$~\cite{clark2019boolq}, $\text{RTE}$~\cite{wang2018glue}, $\text{HellaSwag}$~
\cite{zellers2019hellaswag}, $\text{PIQA}$~\cite{bisk2020piqa}, $\text{RACE}$~\cite{lai2017race}, and $\text{MNLI}$~\cite{williams2017broad}.

As the $\text{Mini-Qwen3}$ model we trained is a $\text{Base}$ version without instruction-following fine-tuning, we adopted a Perplexity ($\text{PPL}$) ranking strategy for downstream task evaluation. Specifically, for each task, we construct prompts by concatenating the context and different options. These prompts are fed through the model's forward pass to obtain the $\text{PPL}$ for each option. The option that yields the lowest $\text{PPL}$ is selected as the model's prediction and compared against the ground truth.

The specific prompt templates used for each task are detailed below:

\paragraph{Prompt Templates}
The specific prompt templates used for the PPL ranking strategy are detailed below. We use the \texttt{Context/Premise} and \texttt{Question/Hypothesis} fields from each dataset to construct the final prompt. For tasks marked with an asterisk (*), the final answers ("Yes" / "No") are appended for PPL calculation and ranking.

\begin{description}
    \item[\textbf{BoolQ}] \texttt{Context is \{passage\}\textbackslash nQuestion is \{question\}\textbackslash nDoes the context answer the question with 'Yes'? Answer with 'Yes' or 'No':}
    \item[\textbf{RTE}] \texttt{Premise is \{premise\}\textbackslash nHypothesis is \{hypothesis\}\textbackslash nDoes the premise entail the hypothesis? Answer with 'Yes' or 'No':}
    \item[\textbf{HellaSwag*}] \texttt{Context is: \{context\}\textbackslash nEnding is: \{ending\}\textbackslash nIs this ending a reasonable continuation of the context? Answer with 'Yes' or 'No':}
    \item[\textbf{PIQA*}] \texttt{Goal is: \{goal\}\textbackslash nSolution is: \{sol\}\textbackslash nIs this solution effective for the goal? Answer with 'Yes' or 'No':}
    \item[\textbf{RACE*}] \texttt{Article is \{article\}\textbackslash nQuestion is \{question\}\textbackslash nIs the option '\{option\}' the correct answer to the question? Answer with 'Yes' or 'No':}
    \item[\textbf{MNLI}] \texttt{Premise: \{premise\}\textbackslash nHypothesis: \{hypothesis\}\textbackslash nWhat is the relationship between the premise and hypothesis? Choose from 'entailment', 'neutral', 'contradiction':}
\end{description}
Note on Binary Tasks: For $\text{HellaSwag}$, $\text{PIQA}$, and $\text{RACE}$, the full prompt includes appending the target answers, such as " Yes" or " No," to the base template for PPL calculation and ranking.

%%%%%%%%%%%%%%%%%%%%%%%%%%%%%%%%%%%%%%%%%%%%%%%%%%%%%%%%%%%%%%%%%%%%%%%%%%%%%%%
%%%%%%%%%%%%%%%%%%%%%%%%%%%%%%%%%%%%%%%%%%%%%%%%%%%%%%%%%%%%%%%%%%%%%%%%%%%%%%%

\section{Efficiency Experiment Detail}\label{Efficiency Experiment Detail}
In this section, we present the detailed experimental settings for the efficiency evaluations.

%\textbf{Pre-Dispatch.} 我们首先使用Megatron-LM的数据处理程序将数据全部转换为token并进行packing，从而将预训练数据集转换为训练时的序列。我们同步使用\GR\处理得到的序列，将专家分配结果以uint8格式预先存储，对于routing score以bfloat16存储，这极大的节省了存储空间。

\textbf{Pre-Dispatch.} We initially employ the data processing pipeline of Megatron-LM to tokenize and pack the raw data, thereby transforming the pre-training dataset into fixed-length training sequences. Concurrently, we process these generated sequences using \GR. To maximize storage efficiency, we pre-compute and store the expert assignment indices in \texttt{uint8} format and the routing scores in \texttt{bfloat16} format, which significantly reduces the storage footprint.

%\textbf{Expert Grouping.} 对于每条序列，我们计算routing affinity vectors，并基于其进行聚类。我们采用层次化聚类方法，即先使用K-means++方法将所有序列分为100簇，然后使用Agglomerative Hierarchical Clustering将其再聚为4簇。我们去除routing affinity vectors分布过于平均的序列后再执行聚类，这是因为这些分布过于平均的序列往往意味着并不具备某个专家组的特别偏好，因此可以去除他们以减少聚类计算量。具体到实验，我们会去除routing affinity vectors熵高于6.85的序列。完成聚类后，我们设置簇和专家为节点，簇的平均routing affinity vectors为簇到专家的权值，构建了一个二分图匹配问题。我们使用匈牙利算法求解这一问题，确定N个专家组，这里N受EP数和优化粒度的影响。我们设置了两种优化粒度，分别为GPU粒度和节点粒度。如果是节点粒度，我们有
% \begin{equation}
%     N = \frac{N_{EP}}{N_{Node}}
% \end{equation}
% 对于GPU粒度，我们有
% \begin{equation}
%     N = \frac{N_{EP}}{N_{GPU}}
% \end{equation}
% 当使用节点粒度时，节点内的专家随机分配。

\textbf{Expert Grouping.} For each sequence, we compute routing affinity vectors to serve as the basis for clustering. We employ a hybrid clustering strategy: initially partitioning all sequences into 100 clusters using K-means++~\cite{arthur2006k}, followed by Agglomerative Hierarchical Clustering~\cite{hastie2009elements} to further merge these into targeted final clusters. Prior to clustering, we filter out sequences with uniformly distributed routing affinity vectors. Such sequences exhibit no distinct preference for specific expert groups; removing them reduces computational overhead without compromising the grouping structure. In our experiments, this is implemented by discarding sequences where the routing affinity vector's entropy exceeds 6.85.Upon completing the clustering, we formulate a bipartite matching problem where the two sets of nodes represent the sequence clusters and the experts, respectively. The edge weights are defined by the average routing affinity vector of each cluster. We solve this problem using the Hungarian algorithm to establish $N$ fixed expert groups, where $N$ depends on the Expert Parallel size and the optimization granularity. We consider two granularity levels: GPU-level and Node-level. Specifically, for a EP group, $N$ equals the total count of individual GPUs for GPU-level optimization, or the total count of server nodes for Node-level optimization. When operating at the node level, experts mapped to a node are subsequently distributed randomly across its intra-node GPUs.

% \textbf{Sample Placement Optimization.}我们基于DeepEP的特殊通信操作执行序列分配，具体来说，我们对每个序列计算其放置到不同设备上时的通信量。我们利用Numpy库实现了以下高效算法，可以快速完成序列分配。

\textbf{Sample Placement Optimization.} We perform sequence assignment based on the specialized communication pattern of DeepEP~\cite{liu2024deepseek}. Specifically, for each sequence, we calculate the potential communication volume associated with its placement on different candidate devices. We implemented the Algorithm \ref{alg:sample_assignment_optimize} using the NumPy library to facilitate rapid sequence assignment.

\begin{algorithm}[t]
    \caption{Optimize Sample-to-Node Assignment}
    \label{alg:sample_assignment_optimize}
    \begin{algorithmic}[1] % [1] 显示行号
        \REQUIRE 
            $\mathcal{S}$: Set of samples,
            $\mathcal{E}: \mathcal{N} \mapsto 2^{\mathbb{N}}$: Expert placement map,
            $k$: Number of experts per token,
            $N$: Total number of nodes

        \STATE Initialize $\mathcal{A} = \{n \mapsto \emptyset \mid n \in \{0,1,\dots,N-1\}\}$ 
        \FORALL{$s \in \mathcal{S}$}
            \STATE $\text{best\_node} \leftarrow 0$, $\max_{\text{comm}} \leftarrow -\infty$ 
            \STATE $\mathbf{D}_s \leftarrow \text{reshape}(\text{dispatch}(s), (-1, k))$ 
            \FORALL{$n \in \mathcal{N}$}
                \STATE $\mathcal{T}_s(n) \leftarrow \left\{ t \in \mathcal{T}_s \mid \mathbf{D}_s[t, :] \cap \mathcal{E}(n) \neq \emptyset \right\}$ 
                \STATE $\text{comm}(n, s) \leftarrow |\mathcal{T}_s(n)|$ 
                \IF{$\text{comm}(n, s) > \max_{\text{comm}}$}
                    \STATE $\max_{\text{comm}} \leftarrow \text{comm}(n, s)$
                    \STATE $\text{best\_node} \leftarrow n$
                \ENDIF
            \ENDFOR
            \STATE $\mathcal{A}(\text{best\_node}) \leftarrow \mathcal{A}(\text{best\_node}) \cup \{s\}$ 
        \ENDFOR
        \RETURN $\mathcal{A}$
    \end{algorithmic}
\end{algorithm}

\section{Training curves under different configurations}\label{ap:Training curves under different configurations}

We present the detailed experimental curves for the models under the four different setups shown in \autoref{ap:configuration} in \autoref{fig:diff_config}. The results demonstrate that the advantages of Grouter persist across different configurations.

% 从结果中可以看到，随着参数量的增加，\GR\的优势也在扩大。我们统计了三种模型的总参数量和专家模块的参数量，结果展示在table1，结果显示三种模型的专家层参数占比随着总参数量上升逐步升高。由于\GR\主要作用是促进专家特化来加速收敛，因此专家层参数占比增加使得\GR\能发挥的作用更明显，从而使得效果也更好。对于这一现象的更深入的研究我们留作未来工作。

As illustrated by the results, the performance advantage of \GR\ expands with increasing model scale. \autoref{tab:expert_tatio} details the breakdown of total versus expert-specific parameters across the three architectures. A key observation is that the ratio of expert parameters to total parameters scales positively with model size. Given that the primary mechanism of \GR\ is to accelerate convergence by fostering expert specialization, an increased proportion of expert parameters amplifies its efficacy, resulting in more substantial performance gains. We reserve a more granular analysis of this scaling behavior for future research.

\begin{table}[htbp]
    \centering
    \caption{Parameter statistics for the evaluated models}
    \resizebox{\linewidth}{!}{
    \begin{tabular}{lccc}
    \toprule
    \textbf{Model} & \textbf{Mini-GPT-OSS} & \textbf{Mini-DS-V2-Lite} & \textbf{Mini-Qwen3} \\
    \midrule
    Parameter & 315M & 958B & 2.81B \\
    Expert Parameter & 199M & 785M & 2.41B \\
    Expert Ratio & 63.19\% & 81.99\% & 85.96\% \\
    \bottomrule
    \end{tabular}
    }
    \label{tab:expert_tatio}
\end{table}

\section{Configuration}\label{ap:configuration}

The detailed configurations for all MoE models employed in our experiments are presented in \autoref{tab:model_configuration}, covering key specifications (e.g., parameter count, hidden dimension, expert setup) for models with diverse scales and architectural designs to support reproducibility and cross-reference in this paper. \textbf{Note that Small-Qwen3 is omitted from this table, as it is derived by simply doubling the expert count of Mini-Qwen3 to 256 to form a 7B model.}

% \begin{table}[t]
%     \centering
%     \caption{Model Configurations}
%     %\vspace{-5pt}
%     \resizebox{\linewidth}{!}{
%     \begin{tabular}{lcccc}
%     \toprule
%     \textbf{Parameter} & \textbf{Mini-GPT-OSS} & \textbf{Tiny-Qwen3} &  \textbf{Mini-DS-V2-Lite} & \textbf{Mini-Qwen3} \\
%     \midrule
%     Parameter & 350M & 550M & 1B & 3B \\
%     hidden\_size & 720 & 384 & 512 & 1024 \\
%     num\_layer & 8 & 8 & 12 & 16 \\
%     num\_experts & 32& 128 & 64 & 128 \\
%     top\_k & 4 & 8 & 6 & 8 \\
%     \bottomrule
%     \end{tabular}
%     }
%     \vspace{-2pt}
%     \label{tab:model_configuration}
% \end{table}

\begin{table}[t]
    \centering
    \caption{Model Configurations}
    %\vspace{-5pt}
    \resizebox{\linewidth}{!}{
    \begin{tabular}{lcccc}
    \toprule
    \textbf{Model} & \textbf{Mini-GPT-OSS} & \textbf{Tiny-Qwen3} &  \textbf{Mini-DS-V2-Lite} & \textbf{Mini-Qwen3} \\
    \midrule
    Parameter & 350M & 550M & 1B & 3B \\
    Hidden\_Size & 720 & 384 & 512 & 1024 \\
    Num\_Layer & 8 & 8 & 12 & 16 \\
    Num\_Experts & 32 & 128 & 64 & 128 \\
    Top\_k & 4 & 8 & 6 & 8 \\
    FFN\_Hidden\_size & 720 & 688 & 2736 & 3072 \\
    Num\_Attention\_Heads & 32 & 8 & 8 & 16 \\
    MoE\_FFN\_Hidden\_size & 720 & 344 & 704 & 384 \\
    Max\_Position\_Embeddings & 131072 & 32768 & 4096 & 40960 \\
    \bottomrule
    \end{tabular}
    }
    \vspace{-2pt}
    \label{tab:model_configuration}
\end{table}

\section{The distillation Loss of \GR\ }\label{apd:distill_loss}

\begin{figure}[htbp]
    \centering
    \includegraphics[width=\linewidth]{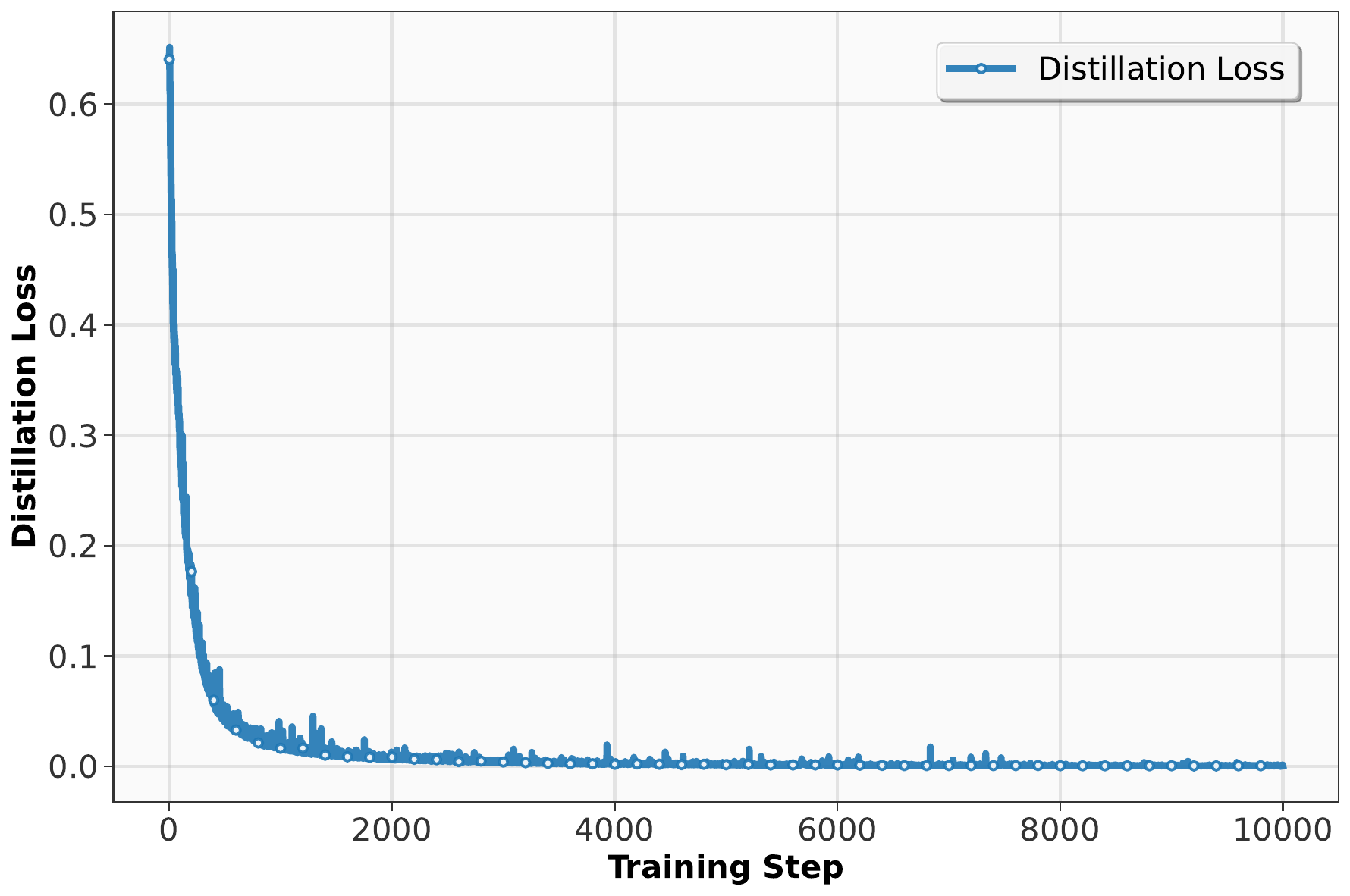}
    \vspace{-3mm}
    \caption{\small Learning Curve of the Distillation Process}
    \label{fig:distill}
    \vspace{-3mm}
\end{figure}

The convergence curve, presented in \autoref{fig:distill}, confirms that the \GR\ effectively learns the structural prior from the Source Model, reaching a stable, low loss value.

\section{Discussion on the Efficient Convergence Achieved by Grouter}\label{ap:Discussion on the Efficient Convergence Achieved by Grouter}

\begin{figure*}[tp]
    \centering 
    \begin{subfigure}{0.33\textwidth}
        \centering
        \includegraphics[width=\linewidth]{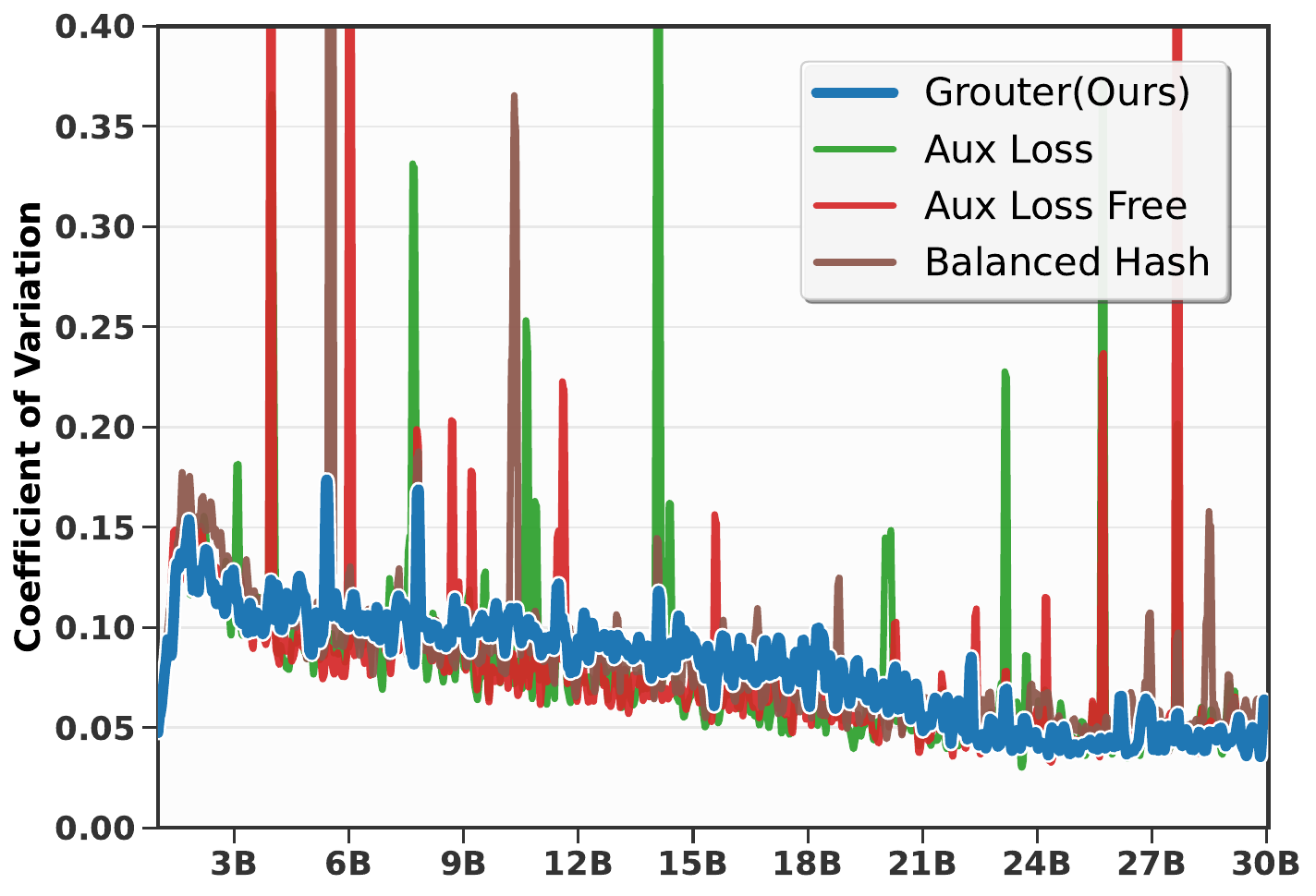}
        \caption{} 
        \label{fig:grad_norm_cv_650m_50}
    \end{subfigure}
    \hfill  
    \begin{subfigure}{0.33\textwidth}
        \centering
        \includegraphics[width=\linewidth]{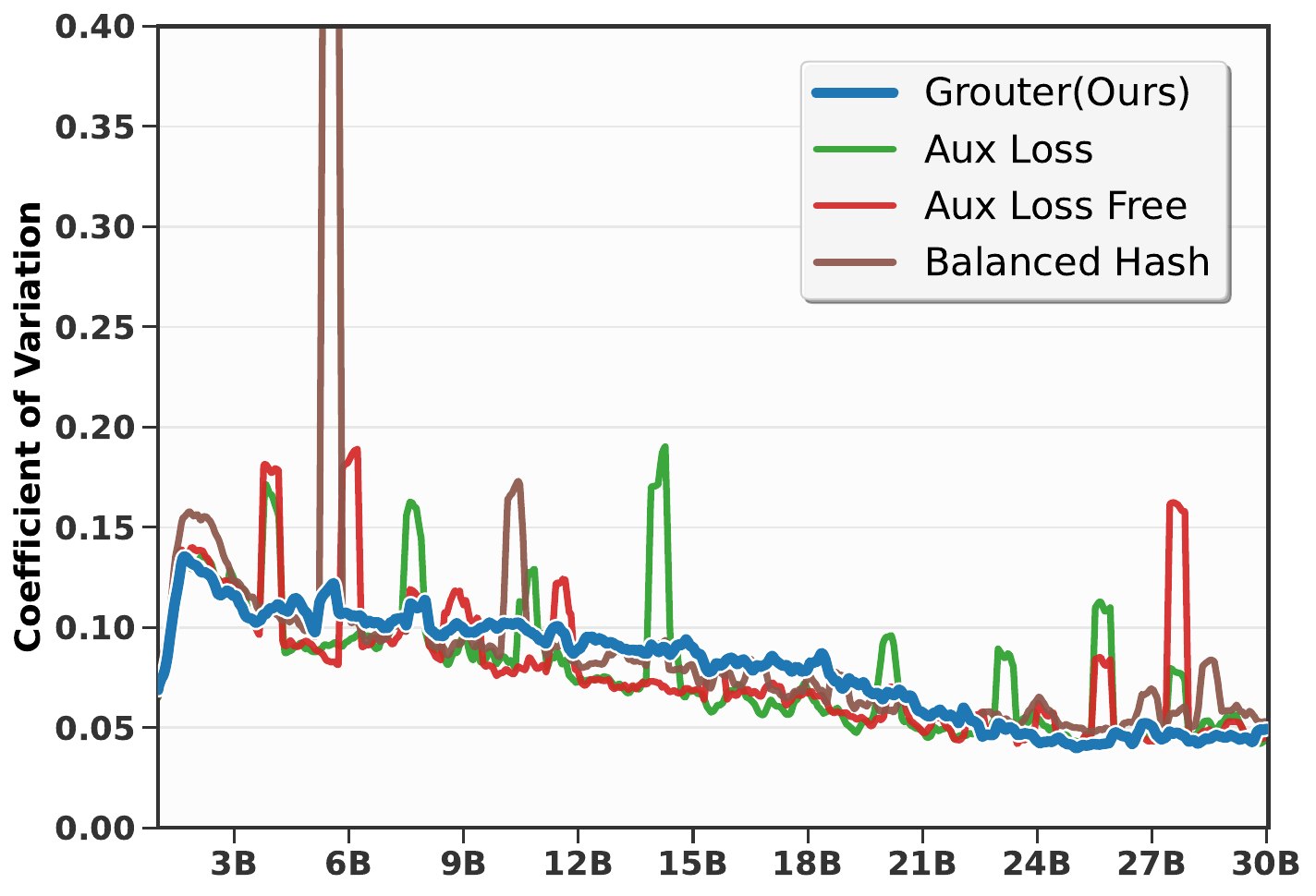}
        \caption{}  
        \label{fig:grad_norm_cv_650m_500}
    \end{subfigure}
    \hfill  
    \begin{subfigure}{0.33\textwidth}
        \centering
        \includegraphics[width=\linewidth]{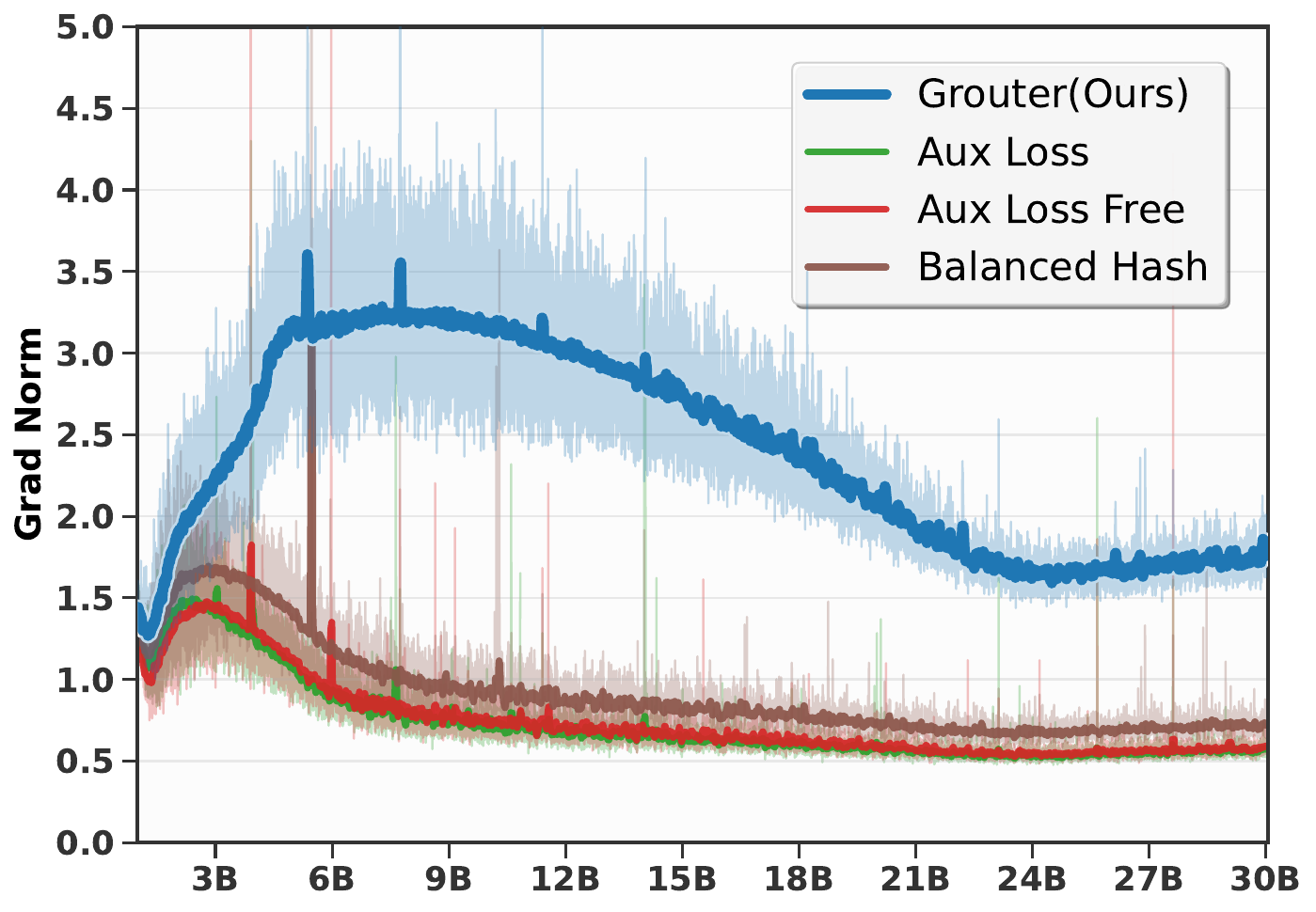}
        \caption{}  
        \label{fig:grad_norm_mean_650m}
    \end{subfigure}
    \vspace{-4mm}
    \caption{\small Analysis of training stability and gradient magnitudes. (a) The CV of the gradient norm, computed using a sliding window of size 50. This illustrates Grouter's superior stability at short timescales, whereas baseline methods exhibit significant spikes. (b) The gradient norm CV with a larger sliding window size of 500, demonstrating Grouter's sustained stability over longer timescales. (c) The trajectory of the gradient norm throughout training. An Moving Average curve (window size 50) is highlighted for visual clarity. The results indicate that Grouter consistently maintains a higher gradient norm compared to baselines.}
    \label{fig:grad_norm}
\end{figure*}

% 我们展示不同滑动窗口下的梯度变异系数于xxx于图1,这样可以展示出在长时间尺度和短时间尺度上不同方法的稳定程度。可以看到无论长期还是短期，Grouter均展现出极佳的稳定性，grad norm始终保持平稳，没有出现突然的波动。这一方面表明\GR\在训练中保持极佳的稳定性，另一方面也说明\GR\能够有效的确保每个专家被分配熟悉的内容，使得整体训练过程持续特化。

We illustrate the coefficient of variation of the gradient norm under different sliding window sizes in \autoref{fig:grad_norm_cv_650m_50}, \autoref{fig:grad_norm_cv_650m_500}, effectively demonstrating stability across both short and long timescales. Regardless of the temporal scale, \GR\ exhibits exceptional stability; the gradient norm remains consistently smooth without abrupt fluctuations or spikes. This not only confirms the robust training stability of \GR\ but also indicates its effectiveness in ensuring that experts consistently receive familiar tokens, thereby facilitating continuous specialization throughout the training process.

% 我们展示训练过程的grad norm于图2。从图中可以看到，\GR\在训练过程中拥有更高的梯度。我们将其解释为由于Grouter已经预先学习到了极好的结构，因此会将相近的embeddin分配给不同的专家，这些embedding会促进专家参数向特定方向优化。而传统router由于不具备这样的先验信息，需要在训练的过程中不断的与专家参数磨合来学习分配能力，其往往会将不相近的embedding分配到同一专家，这些embedding有着不同方向的优化需求，从而造成了梯度相互抵消。

%Figure 2 depicts the training gradient norms, showing that \GR\ consistently yields larger gradient magnitudes. This phenomenon implies more effective optimization: \GR's pre-learned structure routes semantically similar embeddings to distinct experts, thereby aligning the gradient directions for expert parameters. Conversely, traditional routers lack this prior knowledge and must adaptively learn assignments. This learning lag frequently results in the routing of heterogeneous embeddings to the same expert, causing their optimization directions to diverge—a phenomenon known as gradient interference, which mutually cancels out the gradient updates.

\autoref{fig:grad_norm_mean_650m} depicts the training gradient norms, showing that \GR\ consistently yields larger gradient magnitudes. We analyze this by considering the aggregated gradient $\mathbf{G} = \sum_{i \in \mathcal{B}} \mathbf{g}_i$ for an expert over a batch of assigned tokens $\mathcal{B}$. The squared norm of this gradient is given by:

\begin{equation}
\|\mathbf{G}\|^2 = \sum_{i \in \mathcal{B}} |\mathbf{g}_i|^2 + \sum_{i \neq j} \mathbf{g}_i^\top \mathbf{g}_j
\end{equation}

where $\mathbf{g}_i$ represent the gradient of token i.

In \GR, the pre-learned structure routes semantically similar embeddings to the same expert, resulting in positively aligned gradients where $\mathbf{g}_i^\top \mathbf{g}_j > 0$. This constructive interference significantly amplifies the total gradient magnitude $\|\mathbf{G}\|$. Conversely, traditional routers often assign heterogeneous embeddings to the same expert due to the lack of priors. This leads to conflicting optimization directions where $\mathbf{g}_i^\top \mathbf{g}_j < 0$ for many pairs, causing gradient cancellation that diminishes the effective update magnitude.

As training progresses, the update of expert parameters may stagnate when the following condition is met:

\begin{equation}\label{ep:approx}
\sum_{i \neq j} \mathbf{g}_i^\top \mathbf{g}_j \approx -\sum_{i \in \mathcal{B}} |\mathbf{g}_i|^2 
\end{equation}

%专家参数的更新停滞，从而router也倾向于保持当前的状态，以维持当前似乎收敛的表象。然而实际上，此时模型仍然具有很大的收敛空间，只是因为不同embedding表征需求的冲突导致无法进一步收敛。通过优化路由结构，可以向专家指派相近的embedding使得\ref{ep:approx}更晚到来，这极大的提升了模型的收敛潜力。这也是为什么在其他方法的grad norm几乎接近0的时候，\GR\ 仍然保持较大的gradient norm。这也与我们在收敛性实验中\GR\ 有更大的收敛潜力这一现象相符。

When this occurs, the aggregated gradient vanishes due to cancellation. Consequently, the router tends to freeze its current state, creating an illusion of convergence. In reality, however, the model retains significant capacity for optimization; further progress is merely impeded by the conflicting representational requirements of diverse embeddings. By optimizing the routing structure to assign semantically similar embeddings to specific experts, \GR\ effectively mitigates these conflicts and delays the onset of the condition defined in \eqref{ep:approx}. This mechanism significantly extends the model's optimization trajectory. It also explains why \GR\ maintains a substantial gradient norm even when baseline methods approach zero—a finding that aligns with the superior convergence potential observed in our experiments.

\section{Cross-Scale Transferability of \GR}\label{ap:cross_scale}

A key question for the practical applicability of \GR\ is whether its structural prior transfers across model scales—specifically, whether a \GR\ distilled from a smaller source model can effectively guide the training of a larger target model ($P_T > P_S$), or vice versa ($P_S > P_T$). We investigate the former scenario, as it is particularly relevant in practice: practitioners may only have access to a small, trained MoE model yet wish to leverage its routing structure to accelerate the training of a larger target model.

\textbf{Experimental Setup.} Due to the limited availability of small open-source MoE models, we first pre-train a 550M-parameter Tiny-Qwen3 on 30B tokens. We then distill \GR\ from its first MoE layer and apply this \GR\ to guide the training of Mini-DS-V2-Lite. This setup constitutes a challenging cross-architecture, cross-scale transfer: the source and target models differ in both parameter count and expert configuration.

\begin{figure}[htbp]
    \centering
    \includegraphics[width=\linewidth]{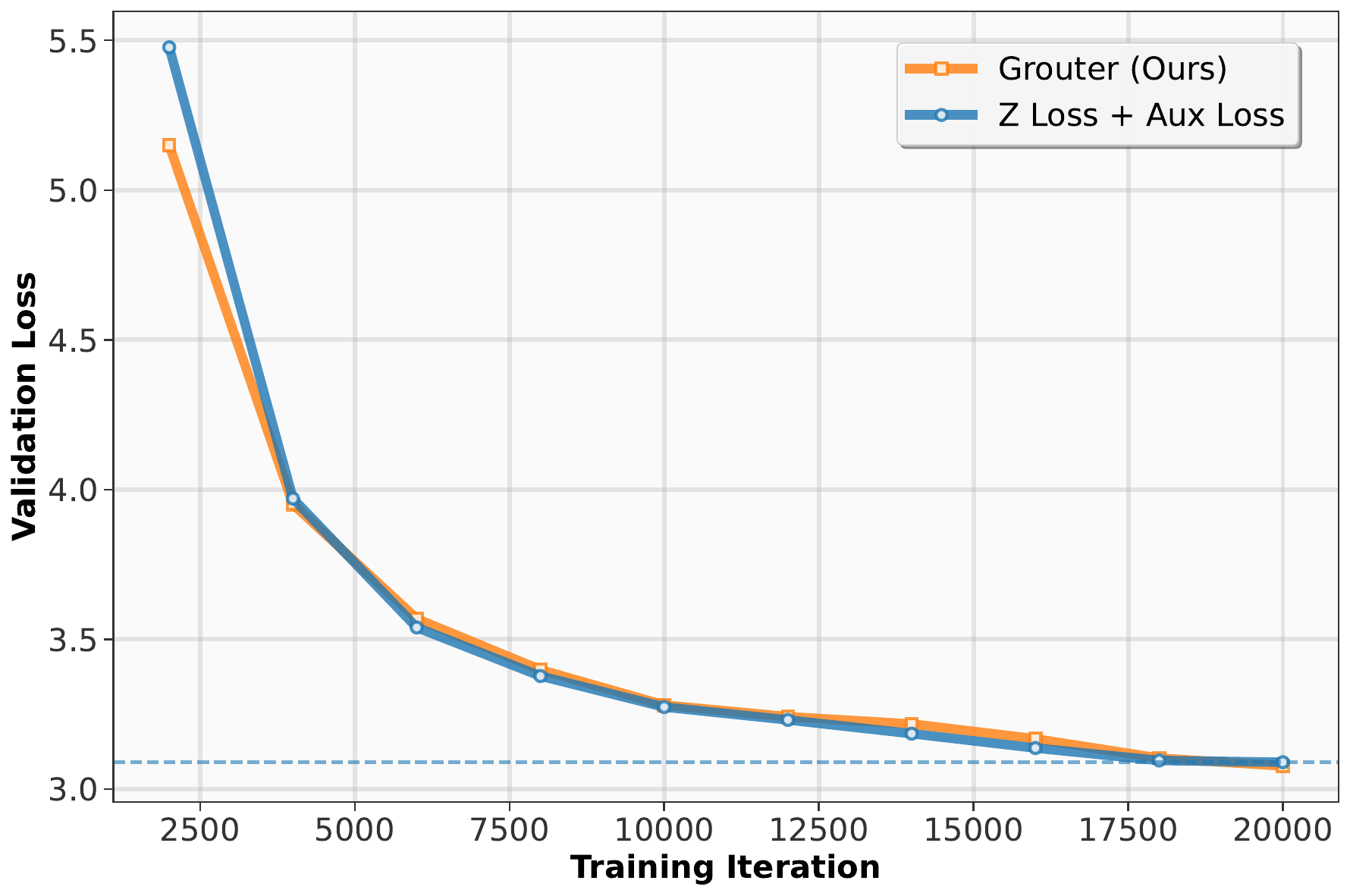}
    \caption{Validation loss comparison for cross-scale transfer. \GR\ matches or even outperforms the baseline throughout, demonstrating effective cross-scale structural transfer.}
    \label{fig:cross_scale_loss}
\end{figure}

\textbf{Results.} \autoref{fig:cross_scale_loss} presents the validation loss curves over 20B tokens pre-training. Despite being distilled from a substantially smaller model, \GR\ achieves performance comparable to the baseline throughout training. 

These results demonstrate that \GR's structural prior generalizes across model scales. Combined with the throughput gains from preemptive routing, this cross-scale transferability translates directly into significant end-to-end training acceleration—even when the \GR\ is distilled from a model substantially smaller than the target.

\section{Overhead Analysis of Preemptive Routing}\label{ap:overhead}

A natural concern regarding preemptive routing is its additional overhead in storage and computation. We provide a detailed cost breakdown below, demonstrating that the total overhead of the \GR\ pipeline is modest relative to the substantial training savings it enables.

\paragraph{Storage Requirements.}
During preemptive routing, the routing decisions for each token are cached to disk. We store each activated expert index as a \texttt{uint8} value (1 byte) and its corresponding routing score in \texttt{bfloat16} (2 bytes), yielding a per-token storage cost of $k \times 3$ bytes, where $k$ denotes the number of activated experts. For Qwen3 with $k{=}8$, this amounts to 24 bytes per token. For a pretraining corpus on the order of trillions of tokens, the total storage requirement is approximately tens of terabytes. This is well within the petabyte-scale capacity of modern cluster storage systems. Moreover, this storage need not be allocated simultaneously: routing decisions can be computed and consumed incrementally, with completed batches deleted before processing subsequent data.

\paragraph{Computational Requirements.}
We further analyze the computational cost of Mini-Qwen3. We estimate FLOPs following the standard approximation from \citet{brown2020language}: for a Transformer with $l$ layers and hidden dimension $h$, the parameter count is approximated as $P \approx 12lh^2$, and the computational costs for $N$ tokens are $C_{\text{fwd}} = 2NP$ (forward) and $C_{\text{bwd}} = 4NP$ (backward). For MoE models, we apply a sparsity correction factor $\alpha = E_a / E$, where $E_a$ and $E$ denote the number of activated and total experts, respectively. The relevant model configurations are summarized as follows:
\begin{itemize}
    \item \textbf{\GR:} $l_{\mathbf{G}} = 3$, $h_{\mathbf{G}} = 512$.
    \item \textbf{Source model (Qwen3-MoE):} $h_{\mathbf{S}} = 2048$, $\alpha_{\mathbf{S}} = 1/16$.
    \item \textbf{Target model (Mini-Qwen3):} $l_{\mathbf{T}} = 16$, $h_{\mathbf{T}} = 1024$, $\alpha_{\mathbf{T}} = 1/16$.
\end{itemize}

We now detail the cost of each pipeline stage.

\textbf{Distillation.} During distillation, each training sequence requires one forward pass through both \GR\ and the source model, plus one backward pass through \GR. Due to the sequential nature of Transformers, only the first $n{=}1$ layer of the source model needs to be executed. Using $N_d = 2.4\text{B}$ distillation tokens:
\begin{align}
    C_{\text{distill}} &= N_d \big( 6 \cdot 12 l_{\mathbf{G}} h_{\mathbf{G}}^2 + 2 \cdot 12 \cdot h_{\mathbf{S}}^2 \cdot \alpha_{\mathbf{S}} \big) \nonumber \\
    &\approx 150{,}995 \text{ TFLOPs}.
\end{align}

\textbf{Expert Folding} (Optional)\textbf{.} Computing the expert co-activation affinity matrix requires a single \GR\ forward pass over $N_f = 52\text{M}$ tokens:
\begin{equation}
    C_{\text{fold}} = N_f \cdot 2 \cdot 12 l_{\mathbf{G}} h_{\mathbf{G}}^2 \approx 981 \text{ TFLOPs}.
\end{equation}

\textbf{Expert Tuning.} Fine-tuning the final projection layer for load balance requires forward and backward passes through \GR\ on $N_t = 52\text{M}$ tokens:
\begin{equation}
    C_{\text{tune}} = N_t \cdot 6 \cdot 12 l_{\mathbf{G}} h_{\mathbf{G}}^2 \approx 2{,}944 \text{ TFLOPs}.
\end{equation}

\textbf{Preemptive Routing.} Pre-computing routing decisions for $N_r = 6\text{B}$ tokens requires a single \GR\ forward pass:
\begin{equation}
    C_{\text{route}} = N_r \cdot 2 \cdot 12 l_{\mathbf{G}} h_{\mathbf{G}}^2 \approx 113{,}246 \text{ TFLOPs}.
\end{equation}

\textbf{Pretraining with \GR.} With the precomputed structural prior, Mini-Qwen3 achieves equivalent loss with only $N_g = 6\text{B}$ tokens:
\begin{align}
    C_{\text{train}}^{\GR} &= N_g \cdot 6 \cdot 12 l_{\mathbf{T}} h_{\mathbf{T}}^2 \cdot \alpha_{\mathbf{T}} \nonumber \\
    &\approx 452{,}983 \text{ TFLOPs}.
\end{align}

\textbf{Baseline (Pretraining Directly).} Pretraining Mini-Qwen3 from scratch to the same loss requires $N_0 = 50\text{B}$ tokens:
\begin{equation}
    C_{\text{train}}^{\text{base}} = N_0 \cdot 6 \cdot 12 l_{\mathbf{T}} h_{\mathbf{T}}^2 \cdot \alpha_{\mathbf{T}} \approx 3{,}774{,}873 \text{ TFLOPs}.
\end{equation}

The complete breakdown is presented in \autoref{tab:overhead}.

\begin{table}[h]
\centering
\caption{Computational cost breakdown of the \GR\ pipeline compared with standard pretraining. Percentages are relative to the baseline pretraining cost.}
\label{tab:overhead}
\vspace{1mm}
\begin{tabular}{lrr}
\toprule
\textbf{Stage} & \textbf{Cost} & \textbf{Relative Cost} \\
\midrule
Distillation & 150,995 & 4.00\% \\
Expert Folding & 981 & 0.03\% \\
Expert Tuning & 2,944 & 0.08\% \\
Preemptive Routing & 113,246 & 3.00\% \\
Pretraining w/ \GR & 452,983 & 12.00\% \\
\midrule
\textbf{Total w/ \GR} & \textbf{721,149} & \textbf{19.11\%} \\
\midrule
Pretraining Directly & 3,774,873 & 100.00\% \\
\bottomrule
\end{tabular}
\end{table}

The total \GR\ pipeline cost amounts to approximately $721{,}149$ TFLOPs, constituting only \textbf{19.1\%} of the baseline pretraining cost. As shown in \autoref{tab:overhead}, the additional overhead introduced by the \GR\ pipeline—encompassing distillation, expert folding, expert tuning, and preemptive routing—accounts for only approximately $7.1\%$ of the baseline pretraining cost. The dominant cost reduction stems from the improved sample efficiency: the target model achieves equivalent loss with $8.3\times$ fewer training tokens.

We further note two practical considerations that reduce the effective overhead. First, \GR\ can be obtained directly from open-source releases, entirely eliminating the distillation cost for practitioners who do not train their own. Second, preemptive routing can be scheduled during GPU idle periods prior to the main training run, effectively overlapping with other cluster operations at zero marginal wall-clock cost. Additionally, the expert grouping and sample placement optimization stages are performed entirely on CPUs using standard scientific computing libraries. Given that CPU resources in modern training clusters are typically underutilized during GPU-intensive workloads, these stages introduce negligible wall-clock overhead when executed concurrently with GPU training preparation.

%%%%%%%%%%%%%%%%%%%%%%%%%%%%%%%%%%%%%%%%%%%%%%%%%%%%%%%%%%%%%%%%%%%%%%%%%%%%%%%
%%%%%%%%%%%%%%%%%%%%%%%%%%%%%%%%%%%%%%%%%%%%%%%%%%%%%%%%%%%%%%%%%%%%%%%%%%%%%%%

\end{document}